\documentclass{article}
\usepackage{spconf}
\usepackage{scalefnt}
\usepackage{relsize}

\usepackage{graphicx}
\usepackage{sidecap}
\usepackage{caption}
\usepackage{calc} 

\usepackage{graphicx}
\usepackage{booktabs}
\usepackage{setspace}
\usepackage{arydshln}

\usepackage{natbib}
\usepackage{xcolor}         
\usepackage[utf8]{inputenc} 
\usepackage[T1]{fontenc}    

   \usepackage[colorlinks=true,
               linkcolor=red,       
               citecolor=green,      
               urlcolor=blue,         
               filecolor=magenta     
               ]{hyperref}

\usepackage{url}            
\usepackage{booktabs}       
\usepackage{tablefootnote}
\usepackage{amsfonts}       
\usepackage{nicefrac}       
\usepackage{microtype}      

\definecolor{myblue}{rgb}{0.255, 0.412, 0.882} 

\definecolor{green1}{RGB}{10,158,10}
\definecolor{blue1}{RGB}{17,85,204}
\definecolor{red1}{RGB}{204,0,0}
\definecolor{mygray}{gray}{0.85}

\definecolor{myblue2}{RGB}{187, 213, 232} 

\usepackage{hyperref}       
\usepackage{url}            
\usepackage{booktabs}       
\usepackage{amsfonts}       
\usepackage{nicefrac}       
\usepackage{microtype}      
\usepackage{lipsum}
\usepackage{wrapfig}
\usepackage{siunitx}
\usepackage{sidecap}

\usepackage{amsmath,amsfonts,amssymb,amsthm}
\usepackage{mathtools}
\usepackage{multirow}

\usepackage{algorithm}
\usepackage{algorithmicx}
\usepackage{algpseudocode}
\usepackage{amsmath}
\usepackage{amssymb}
\usepackage{bm} 

\usepackage{colortbl}
\usepackage{booktabs}
\usepackage{multirow}
\usepackage{siunitx}
\usepackage{array}

\usepackage{subcaption}

\usepackage{bbm}

\usepackage{enumitem}

\newcolumntype{P}[1]{>{\hspace{1ex}}p{#1}<{\hspace{1ex}}}

\title{\vspace{-30pt}A Stable Neural Statistical Dependence Estimator for Autoencoder Feature Analysis\vspace{-10pt}}

\author{%
  Bo Hu, \; Yuheng Bu,\; Jos{\'e}~C.~Pr{\'\i}ncipe\\
  Department of Electrical and Computer Engineering\\
  University of Florida\\
  \texttt{\{hubo, buyuheng\}@ufl.edu\; principe@cnel.ufl.edu }\\}

\name{Bo Hu, \; Jos{\'e}~C.~Pr{\'\i}ncipe\vspace{-10pt}}
\address{Department of Electrical and Computer Engineering\\
University of Florida\\
  \texttt{hubo@ufl.edu\; principe@cnel.ufl.edu }\\\vspace{-25pt}}

\begin{document}

\maketitle

\vspace{-6pt}

\begin{abstract}
Statistical dependence measures like mutual information is ideal for analyzing autoencoders, but it can be ill-posed for deterministic, static, noise-free networks. We adopt the variational (Gaussian) formulation that makes dependence among inputs, latents, and reconstructions measurable, and we propose a stable neural dependence estimator based on an orthonormal density-ratio decomposition. Unlike MINE, our method avoids input concatenation and product-of-marginals re-pairing, reducing computational cost and improving stability. We introduce an efficient NMF-like scalar cost and demonstrate empirically that assuming Gaussian noise to form an auxiliary variable enables meaningful dependence measurements and supports quantitative feature analysis, with a sequential convergence of singular values.
\end{abstract}\vspace{-5pt}

\section{Introduction}

The application of statistical dependence measures to analyze autoencoders is important but technically challenging. Statistical dependence measures, most notably mutual information, quantify uncertainty and the statistical relationships between two random variables. Applying 
such measures to analyze a cascaded encoder-decoder structure has been common in communication systems and information theory~\cite{xu2024neural, shwartz2023information}.

Thus, it is natural to apply statistical dependence measures to analyze an autoencoder structure. Another well-established result is that, for a cascaded structure like an autoencoder, minimizing the Mean-Squared Error (MSE) between the input and the output variables amounts to maximizing the statistical dependence, particularly the mutual information, between the input and the intermediate variables~\cite{chen2016infogan, xu2022information}. 

However, we find the central obstacle to be that, for a static end-to-end neural network, the statistical dependence between its input and output variables is undefined and not measurable when no noise is assumed or present. Directly applying an estimator to a deterministic end-to-end neural network is therefore generally ill-posed. The contradiction is that a neural network’s projection function is defined over an actual domain far larger than the finite set of provided dataset samples. Otherwise, the network will not have generalization capability. A common remedy is the variational Bayesian treatment of autoencoders~\cite{kingma2013auto, lopez2018information}, in which the encoder and decoder are assumed to parameterize two conditional distributions under Gaussian assumptions. We find this assumption generally correct and practically helpful. We will show that, under this Gaussianity assumption, statistical dependence between inputs, features, and reconstructions becomes well-defined and can be estimated meaningfully.

The obstacle also lies in the stability of the statistical dependence estimator. The conventional procedure of building a statistical dependence estimator often starts with picking a convex function to define an $f$-divergence, build a variational bound for the $f$-divergence with the convex conjugate~\cite{jordan1, jordan2} or the Donsker–Varadhan formula~\cite{belghazi2018mine}, and derive a variational cost from this bound such that minimizing this cost will give us estimation of the mutual information. When parameterized by a neural network, this approach is typically referred to as the Mutual Information Neural Estimator (MINE)~\cite{belghazi2018mine}. In MINE, a neural network $f_\theta$ is to estimate the density ratio with $f_\theta(X,Y) = \frac{p(X,Y)}{p(X)p(Y)}$ , whose inputs are the concatenation of $X$ and $Y$. 

It is well known that MINE can be unstable in practice. This paper argues that this instability may come from the parameterization $f_\theta(X,Y)$. In MINE, the variational cost involves two expectations, $\mathbb{E}_{X,Y\sim p(X,Y)}[f_\theta(X,Y)]$ and $\mathbb{E}_{X,Y\sim p(X)p(Y)}[f_\theta(X,Y)]$. The second expectation requires sample pairs drawn from the product of the marginals, whereas typically we only have sample pairs from the joint. In practice, it is approximated by re-pairing samples within a batch. With $N$ joint pairs, naive re-pairing will yield $N^2$ marginal pairs, resulting in high computational complexity and instability.

To address MINE’s instability, we build on our prior work~\cite{ma2024learning, hu2024learning} by leveraging an orthonormal decomposition of the density  $\frac{p(X,Y)}{p(X)p(Y)}=\sum_{k=1}^K \sqrt{\lambda_k}\cdot\phi_k(X)\cdot\psi_k(Y)$. Instead of approximating the density ratio directly, we learn its left and right singular functions $\boldsymbol{\phi}(X)$ and $\boldsymbol{\psi}(Y)$ with neural networks, avoiding input concatenation and eliminating the need for re-pairing. Our previous papers proposed two matrix-based costs to learn these functions, based on either matrix inverses or log-determinants. Here we introduce a third alternative: a scalar cost similar to Nonnegative Matrix Factorization (NMF) that avoids both operations, further improving efficiency. We find this to be more stable and natural than MINE. 

Combining the variational assumptions for autoencoders with our stabilized dependence estimator, we can give an accurate statistical dependence measurement in the autoencoder setting. Beyond feature analysis, we also provide an intuitive example of feature learning from the principle of maximizing statistical dependence, although this requires assuming additive noise on the data samples.\vspace{7pt}

\noindent \textbf{Major result.} Our main experimental result shows that, given $X$ and $Y$, we can construct an auxiliary variable $X'$ by assuming additive Gaussian noises on $X$. Although dependence for $\{X,Y\}$ may be ill-posed to measure in the static setting, dependence for $\{X',X\}$ and $\{X',Y\}$ is well-defined and measurable. Good features $Y$ should be such that the dependence from $\{X',X\}$ to $\{X',Y\}$ does not decay when $X$ is replaced with $Y$. While this is a compromise relative to the noise-free idealization, we find it effective for quantitative analysis and feature learning.\vspace{-10pt}

\section{Orthonormal Decomposition of the Density Ratio} \label{costs_ratio_sec}


As discussed, our prior work considers approximating the density ratio by its singular functions, rather than approximating it directly. This decomposition is formally described in Eq.~\eqref{orthonormal_decomposition}. We want to approximate the density ratio using the left and right singular functions $\widehat{\phi_1},\widehat{\phi_2},\ldots,\widehat{\phi_K}$ and $\widehat{\psi_1},\widehat{\psi_2},\ldots,\widehat{\psi_K}$. The singular values $\sqrt{\lambda_1},\sqrt{\lambda_2},\ldots,\sqrt{\lambda_K}$ lie in $[0,1]$ and in fact a multivariate statistical dependence measure; they are all zero iff the two variables are statistically independent.

Since we want to use two neural networks to approximate the top $K$ left and right singular functions, these networks must be multiple-output. Let $\mathbf{f} = [f_1, f_2,\ldots,f_K]^\intercal$ and $\mathbf{g} = [g_1, g_2,\ldots,g_K]^\intercal$ denote their outputs. We have previously identified two training costs, the log-determinant cost and the trace cost in Eq.~\eqref{costfunctions2}. Both require constructing the autocorrelation matrices $\mathbf{R}_F$ and $\mathbf{R}_G$, as well as the cross-correlation matrix $\mathbf{P}$, by taking expectations of their inner and cross products.

Proofs that optimizing these two costs yields the singular functions of the density ratio are provided in the appendix.\vspace{-5pt}

\begin{equation}
\resizebox{.63\linewidth}{!}{
$\begin{aligned}
\frac{p(X,Y)}{{p(X)}\,{p(Y)}} = \sum_{k=1}^K \sqrt{\lambda_k}\cdot \widehat{\phi_k}(X)\cdot \widehat{\psi_k}(Y).
\end{aligned}$}
\label{orthonormal_decomposition}
\end{equation}\vspace{-15pt}

\begin{equation}
\resizebox{1\linewidth}{!}{
$\begin{gathered}
 \text{(log-det cost)\;\;} \min_{\mathbf{f}, \mathbf{g}} \; \log\det \mathbf{R}_{FG} - \log\det \mathbf{R}_F - \log\det \mathbf{R}_G,\\
\text{(trace cost)\;\;} \max_{\mathbf{f}, \mathbf{g}} \; Trace(\mathbf{R}_F^{-1}\mathbf{P}\mathbf{R}_G^{-1}\mathbf{P}^\intercal),\\
\mathbf{R}_F = \mathbb{E}\left[ \mathbf{f}(X)\mathbf{f}^\intercal (X)\right],\; \mathbf{R}_G = \mathbb{E} \left[\mathbf{g}(Y) \mathbf{g}^\intercal (Y)\right],\\ \mathbf{P} = \mathbb{E} \left[ \mathbf{f}(X) \mathbf{g}^\intercal (Y)\right], \; \mathbf{R}_{FG} = \begin{bmatrix}\mathbf{R}_F & \mathbf{P} \\
\mathbf{P}^\intercal & \mathbf{R}_G
\end{bmatrix}.
\end{gathered}$}
\label{costfunctions2}
\end{equation}

In this paper, we propose a third alternative. The bottlenecks of the previous costs are the need to compute matrix inverses and log-determinants. Beyond the SVD, another useful algebraic decomposition is the Nonnegative Matrix Factorization (NMF), which we find well suited here as the density ratio is nonnegative. 

Given neural networks $\mathbf{f} = [f_1, f_2,\ldots,f_K]^\intercal$ and $\mathbf{g} = [g_1, g_2,\ldots,g_K]^\intercal$, we directly approximate the density ratio as $\sum_{k=1}^K f_k(X)\, g_k(Y)$. The functions are no longer required to be orthonormal. Instead, we only impose nonnegativity, which can be enforced by using a ReLU activation in the final layer. By the Schwarz inequality, the following inequality will hold: 
\begin{equation}
\resizebox{1\linewidth}{!}{
$\begin{aligned}
\frac{\left(\mathbb{E}\left[ \sum_{k=1}^K f_k(X)g_k(Y) \right]\right)^2} {\sum_{i, j = 1}^K \mathbb{E}\left[  f_i(X)f_j(X) \right] \cdot \mathbb{E}\left[ g_i(Y) g_j(Y) \right] } \leq \iint \frac{p^2(X,Y)}{p(X)p(Y)} dXdY.
\end{aligned}$}
\label{bound2}
\end{equation}
Deriving this inequality is simple: it follows from a direct application of the Schwarz inequality to the inner product $\langle \sum_{k=1}^K f_k(X)\, g_k(Y),\, p(X,Y)\rangle$. For completeness, we provide the full proof in the appendix. 

We take the left-hand side of inequality~\eqref{bound2} to be our new NMF-inspired cost. It is bounded by the mutual information when the convex function $f$ defining the $f$-divergence is chosen to be linear, corresponding to Rényi’s divergence.

We still compute the autocorrelation matrices $\mathbf{R}_F = \mathbb{E}\left[ \mathbf{f}(X)\mathbf{f}^\intercal (X)\right]$ and $\mathbf{R}_G = \mathbb{E} \left[\mathbf{g}(Y) \mathbf{g}^\intercal (Y)\right]$. The denominator of the new cost (the left side of the bound~\eqref{bound2}) is equivalent to $\sum_{i,j=1}^K (\mathbf{R}_F)_{i, j} (\mathbf{R_G})_{i,j}$, or further $\sum_{i,j=1}^K (\mathbf{R}_F \odot \mathbf{R}_G)_{i,j}$, where \(\odot\) denotes the Hadamard (elementwise) product. Here $\mathbf{R}_F$ and $\mathbf{R}_G$ are $K\times K$ matrices. The final cost has the form
\begin{equation}
\begin{aligned}
c = \frac{\left(\mathbb{E}\left[ \sum_{k=1}^K f_k(X)g_k(Y) \right]\right)^2}{\sum_{i,j=1}^K (\mathbf{R}_F \odot \mathbf{R}_G)_{i,j}}, \text{\;or\;} \log c.
\end{aligned}
\label{maximize_cost}
\end{equation}
We want to maximize $c$ or $\log c$. A small constant $\epsilon = 10^{-6}$ is added to the denominator for stability. This new cost no longer contains matrix inverses or log-determinants nor requires their gradients for the updates. The full algorithm is detailed below.\vspace{-13pt}
\begin{algorithm}
\small
\caption{Neural NMF for the density ratio.}
\begin{algorithmic}[1]
\Require Initialize two neural networks with multivariate outputs $\mathbf{f} = [f_1,f_2,\cdots,f_K]^\intercal$ and $\mathbf{g} = [g_1,g_2,\cdots,g_K]^\intercal$. The outputs must be nonnegative, so the final activation layer can be ReLU.\vspace{4pt}
\Statex \hspace*{-\algorithmicindent} \textbf{At each iteration do:}
\State Pass $X_1, X_2,\cdots,X_N$ through network $\mathbf{f}$ to obtain $\mathbf{f}(X_1), \mathbf{f}(X_2),\cdots, \mathbf{f}(X_N)$;
\State Pass $Y_1, Y_2,\cdots,Y_N$ through network $\mathbf{g}$ to obtain $\mathbf{g}(Y_1), \mathbf{g}(Y_2),\cdots, \mathbf{g}(Y_N)$;
\State Compute $\mathbf{R}_F = \mathbb{E}\left[ \mathbf{f}(X)\mathbf{f}^\intercal (X)\right]$ and $\mathbf{R}_G = \mathbb{E} \left[\mathbf{g}(Y) \mathbf{g}^\intercal (Y)\right]$ using the corresponding networks; compute the joint expectation $\mathbb{E}\left[ \sum_{k=1}^K f_k(X)g_k(Y) \right]$ using both networks;
\State Construct and maximize the cost in Eq.~\eqref{maximize_cost};\vspace{2pt}

\Statex \hspace*{-\algorithmicindent} \textbf{Output:} A nonnegative factorization of the density ratio. If needed, an SVD can be used after training to recover the singular functions and singular values.
\end{algorithmic}
\label{alg1}
\end{algorithm}\vspace{-5pt}

The same trick may apply to estimating Shannon's mutual information via the Donsker–Varadhan formula, but we focus on the $L_2$ case because of its link to the eigenexpansion.

A common critique of our old costs is also the choice of the number of singular functions, the neural network output dimensions, as if this number is too large, the model tends to diverge or be biased. But with this new cost, we found that the dimension can be picked to be very large while the training is still stable.

\section{Applying A Statistical Dependence Estimator to An Autoencoder}

We now describe how the proposed measure can be applied to an autoencoder. We use the notation $\mathcal{N}(X; v)$ to denote a regular Gaussian density function with a shared scalar variance $v$ across dimensions.

A conventional variational analysis of the autoencoder starts with the assumption $p(Y|X) = \mathcal{N}(Y-\textbf{E}(X);v_p)$ and $q(X|Y) = \mathcal{N}(X-\textbf{D}(Y);v_q)$. The objective of an autoencoder, under this assumption, follows
\begin{equation}
\resizebox{.78\linewidth}{!}{
$\begin{gathered}
\max_{\mathbf{P}_{Y|X}, \mathbf{Q}_{X|Y}} \;\; {Trace}\left({diag}(\mathbf{P}_X) \cdot \mathbf{P}_{Y|X} \cdot \log \mathbf{Q}_{X|Y}\right).\vspace{-5pt}
\end{gathered}$}
\label{discrete_case222}
\end{equation}\vspace{-15pt}
\begin{equation}
\resizebox{.78\linewidth}{!}{
$\begin{gathered}
\max_{p(Y|X), p(X|Y)}\; \iint p(X) \cdot p(Y|X) \cdot \log q(X|Y) dX dY.\vspace{6pt}
\end{gathered}$}
\label{continuous_case222}
\end{equation}
We find it always helpful for analytical purposes to write down the discrete equivalence of an autoencoder (Eq.~\eqref{discrete_case222}), where $\mathbf{P}_{Y|X}$, $\mathbf{Q}_{X|Y}$ are Markov transition matrices. Our main focus, however, is the continuous formulation in Eq.~\eqref{continuous_case222}: we sample from the data distribution $p(X)$, pass samples through an encoder $p(Y|X)=\mathcal{N}(Y;\mathbf{E}(X),v_p)$, and then a decoder $q(X|Y)=\mathcal{N}(X;\mathbf{D}(Y),v_q)$. The useful property of this objective is that taking $\log q(X|Y)$ cancels the exponential term in the Gaussian density, reducing the reconstruction term to just the MSE. 

A naive but important bound is to apply the nonnegativity of the conditional KL divergence:
\begin{equation}
\resizebox{.8\linewidth}{!}{
$\begin{aligned}
&\iint p(X) \cdot p(Y|X) \cdot \log q(X|Y) dX dY  \\
&\hspace{25pt}\leq \iint p(X) \cdot p(Y|X) \cdot \log p(X|Y) dXdY\\
&\hspace{25pt} = \mathbf{MI}(X;Y)-\mathbf{H}(X).
\end{aligned}$}
\label{inequality_pq}
\end{equation}
It shows that the optimality condition is met when the decoder $q$ is an inverse mapping of the encoder $p$, such that $q(X|Y) = p(Y|X)$. When this condition is satisfied, the cost can be written as the mutual information $\mathbf{MI}(X; Y)$ minus the data entropy $\mathbf{H}(X)$. Note that we seek the maximum value of the objective, which means that even after the decoder becomes an inverse mapping of the encoder, we must still find the $Y$ that maximizes the mutual information with $X$. This shows why the statistical dependence measure is needed. 

The variational treatment of autoencoders defines two joint densities: $p(X,Y)$ for the encoder and $q(X,Y)$ for the decoder. We know that $q$ is intended to be an inverse mapping of $p$, and that the statistical dependence of $p(X,Y)$ is encouraged to be maximized. Our goal is then to measure the statistical dependence of these joint densities. These variances are not directly computable, but can only be estimated. The hidden dilemma here is therefore in fact how to set the two noise variances, $v_p$ for the encoder and $v_q$ for the decoder, or what their true values are. Our arguments are as follows.

\vspace{12pt}

\noindent \textbf{Pick the implicit $v_q$.} For the decoder variance $v_q$ for the reconstructions, we argue that even when the static network is trained noise-free, in fact, the optimal  decoder variance $v_q$ is exactly the empirical value of the MSE. 

When $v_q$ is included, the autoencoder objective (Eq.~\eqref{continuous_case222}) is a Gaussian negative log-likelihood: it is the squared reconstruction error scaled by $v_q$, plus the normalization terms of the Gaussian:
\begin{equation}
\resizebox{1\linewidth}{!}{
$\begin{gathered}
ae\_obj =  \frac{L}{2} \cdot (\log 2\pi + \log v_q) + \frac{\iint ||X-\textbf{D}(Y)||_2^2 \,  p(X,Y) \, dXdY}{2v_q}.
\end{gathered}$}
\end{equation}
The autoencoder minimizes this. In practice, we often drop the scaling and normalizing constants, and optimize only the reconstruction MSE in the numerator. But if we do care about $v_q$, let us take the derivative of this objective $ae\_obj$ with respect to the variance $v_q$, and set it to zero:
\begin{equation}
\resizebox{.89\linewidth}{!}{
$\begin{gathered}
\frac{\partial ae\_obj}{\partial v_q} = \frac{L}{2v_q} - \frac{\iint ||X-\textbf{D}(Y)||_2^2 \, p(X,Y)\,dXdY }{2v_q^2} = 0, \\
v_q = \frac{1}{L} \iint ||X-\textbf{D}(Y)||_2^2\,  p(X,Y) \, dXdY.
\end{gathered}$}
\end{equation}
Therefore, the optimal variance $v_q$ for a certain MSE value is the MSE value itself. If we minimize the MSE while ignoring $v_q$, then after training we can simply set $v_q$ to the current (or final optimal) MSE value. Minimizing the reconstruction error can be viewed as shrinking the Gaussian balls: the smaller the error, the smaller the radius of these Gaussian balls.\vspace{12pt}

\noindent \textbf{Pick the implicit $v_p$.} Retrieving the encoder variance $v_p$ is much more difficult. The question is posed as follows. If we train a static noise-free autoencoder, empirically, the intermediate $Y$ should not have any external noise. But the assumption requires $p(Y|X) = \mathcal{N}(Y-\textbf{E}(X);v_p)$ to be parameterized as a Gaussian. So when we want to measure the statistical dependence, we need to handle this mismatch.

Our empirical conclusion in this paper is that the this noise variance $v_p$ is at a level of $10^{-4}$ to $10^{-5}$ for toy datasets and a sigmoid activation, and may exist when the setting is static without additive Gaussian noises. The evidence is empirical, and comes from the following few angles. 

First, if we train a standard static autoencoder, it can clearly be observed that convergence has stages: there is a noticeable transition in which the reconstructions and the learned feature boundaries appear coarse at first, and then become more and more fine-grained as training progresses.

Now we train an autoencoder with additive noise in the features $Y' = Y + \sqrt{v_p}\cdot {noise}$ like in VAE. We repeat training multiple times, each time using a different (and progressively smaller) $v_p$. Across these runs, the reconstructions again exhibit a transition from coarse to fine-grained as $v_p$ decreases. This suggests that even in the nominally static setting, an effective feature variance $v_p$ may be present and may implicitly decrease during training. Empirically, the smallest $v_p$ that does not affect reconstruction quality or training error is on the order of $10^{-5}$, which may reflect the effective $v_p$ in the static case. 

Second, if we look at the reconstructions when the feature dimension is low and visualize them, they look like a low-dimensional manifold existing in a high dimensional space. In order to create this manifold, there must be continuity in the feature variable. And the source of this continuity may come from the implicit Gaussian noises. 

We conducted a quantitative experiment in a Nyström-style fashion. If the data density for toy examples is given in closed form, reasonably, it should be possible to learn $p$ and $q$ in a non-empirical way, not from empirical data, but by operating directly on the probability density. Thus, we parameterize the encoder and decoder directly with two Markov transition matrices, without neural networks. We found that Gaussian assumptions are required for this non-empirical approach to match the results of an autoencoder. Sweeping over $v_p$, we found that the closest match occurs when $v_p$ is at a level of $10^{-5}$.

The third piece of evidence is that, in statistical dependence estimation, assuming the reconstruction variance $v_q$ without assuming the feature variance $v_p$ makes the estimation unstable. Without $v_p$, we found that the dependence estimation between data samples and reconstructions is stable, but the estimation between features and reconstructions is not. After adding the feature noise (i.e., assuming $v_p$), the estimation becomes more stable. The intrinsic $v_p$ for a regular autoencoder, while still requiring further research, could be around $10^{-8}$ or $10^{-9}$.

Details of these investigations are given in the appendix.\vspace{12pt}

\noindent \textbf{Our findings.} After selecting $v_p$ and $v_q$, we apply statistical dependence estimators to a trained autoencoder to measure the statistical dependence between the inputs, the intermediate features, and the reconstructions. The following notations are used:\vspace{-3pt}
\begin{itemize}[leftmargin=*]
\item $X$: data samples;\vspace{-4pt}
\item $Y$: the outputs from passing data samples through the static encoder (features);\vspace{-4pt}
\item $Y'$: $Y' = Y + \sqrt{v_p}\cdot {noise}$, features $Y$ with additive noise;\vspace{-4pt}
\item $\widehat{\,X\,}$: the outputs obtained from passing the noisy $Y'$ through the static decoder (reconstructions);\vspace{-4pt}
\item $\widehat{\,X\,'}$: $\widehat{\,X\,'} = \widehat{\,X\,} + \sqrt{v_q}\cdot {noise}$, reconstructions $\widehat{\,X\,}$ with additive noise.
\end{itemize}
With $X$, $Y$, $Y'$, $\widehat{\,X\,}$, and $\widehat{\,X\,'}$, we can measure the statistical dependence between any pair of them, and generate a meaningful measurement. Our findings can be summarized below:\vspace{-4pt}
\begin{enumerate}[leftmargin=*]
\item $\{X,Y\}$ and $\{Y',\widehat{\,X\,}\}$: measuring dependence between the input and the output of a static network is ill-posed. During training, the estimate increases without bound and diverge.\vspace{-3pt}

\item $\{X,Y'\}$ and $\{Y,Y'\}$: the dependence between $X$ and the noise-corrupted $Y'$ is well-defined and can be meaningfully estimated. Further, it coincides with the dependence between $Y$ and $Y'$, i.e., between the feature and its noise-corrupted version.

\item $\{Y',\widehat{\,X\,'}\}$ and $\{\widehat{\,X\,},\widehat{\,X\,'}\}$: similarly, we can meaningfully estimate the dependence between $Y'$ and noise-corrupted $\widehat{\,X\,}'$, but not between $Y'$ and noise-free $\widehat{\,X\,}$. The measurement value for these two pairs are equal.\vspace{-3pt}
\end{enumerate}

\noindent This raises the point that in a static neural network, a meaningful measurement of the statistical dependence requires the Gaussianity assumption. 

The pair $\{Y, Y'\}$ is simple, since $Y$ is the feature and $Y'$ is its noise-corrupted version. The pair $\{X, Y'\}$ is complex, since $X$ is high-dimensional data and $Y'$ is the projected feature with noises. Despite this difference in complexity, they have the same statistical dependence value. The same analysis applies to the decoder side for the pairs $\{Y',\widehat{\,X\,'}\}$ and $\{\widehat{\,X\,},\widehat{\,X\,'}\}$: when $X$ is substituted by $Y$, the dependence does not decay.\vspace{-15pt}

\section{Experiments}\vspace{-5pt}

\noindent \textbf{Datasets.} We conduct experiments on two datasets: a two-moons toy dataset (visualized in Fig.~\ref{interpolated_samples}) and the standard MNIST handwritten digit dataset. Four neural networks are required: an encoder, a decoder, and two dependence-estimator networks.\vspace{3pt}

\noindent \textbf{Baselines and parameters.} To demonstrate the effectiveness of our dependence estimator, we compare against three groups. (i) We contrast our new NMF-like scalar cost (\textcolor{red}{\textbf{NMF-DR}}) with our previous logdet and trace costs (\textcolor{blue}{\textbf{LOGDET}}, \textcolor{blue}{\textbf{TRACE}}). (ii) We include standard MINE estimating Shannon mutual information (\textcolor{orange}{\textbf{MINE}}). (iii) We also report kernel-based dependence measures, including \textbf{KDE}, \textbf{KICA}~\cite{bach2002kernel}, and \textbf{HSIC}~\cite{gretton2005measuring}. 

Our main emphasis is on (i) and (ii); kernel methods are reported for reference and are not intended as state-of-the-art or practical baselines. All details can be found in the appendix.

Key hyperparameters are as follows. Unless stated otherwise, the encoder projects data into \textbf{1D features}. We also report higher-dimensional features and sweep this dimension for MNIST (Table~\ref{tab:trajectory_pairs9999}).

For decomposition costs, we need to set the estimator output dimension (number of singular functions). It is fixed to be \textbf{2000} for the new NMF-like cost. For the trace and logdet costs, we use \textbf{50}, except for the four cases noted in Table~\ref{tab:objectives_pairs222}, where we use \textbf{500}; values above 500 are very costly.

If this decoder dimension is much larger than the number of positive singular values, trace and logdet show an unwanted bias, while the NMF-like cost does not.

We perturb the features with noise as $Y' = Y + \sqrt{v_p} \cdot \textit{noise}$ and set $v_p = 10^{-4}$. We also sweep $v_p$ in Table~\ref{tab:vp_sweep_mnist999} and~\ref{tab:vp_sweep_mnistttt}.\vspace{7pt}

\noindent \textbf{Result 1.} Direct comparisons are reported in Table~\ref{tab:objectives_pairs} (two-moon) and Table~\ref{tab:objectives_pairs222} (MNIST).

First, we guarantee that ours three costs yield the same unbiased estimate of R\'enyi mutual information across variable pairs, while NMF-DR is the most efficient and scalable.

Second, we observe consistent equivalences in dependence: the pairs $\{X,Y'\}$ and $\{Y,Y'\}$ are close, and so are $\{Y',\widehat{X'}\}$ and $\{\widehat{\,X\,},\widehat{X'}\}$, whereas $\{X,\widehat{X'}\}$ is the lowest. Overall,they suggest a clear substitution pattern: the original data $X$ can be replaced by the noise-free features $Y$ without changing the dependence, and the noise-corrupted features $Y'$ (input to the decoder) and the noise-free reconstruction $\widehat{\,X\,}$ (output of the decoder) are also interchangeable. The encoder-side dependence is consistent across datasets (approximately $28$ for 1D features at $v_p=10^{-4}$), whereas the decoder-side dependence varies with the data dimension.

A notable implication is that the dependence between the data $X$ and the reconstruction $\widehat{\,X\,}$ equals that between noise-free features $Y$ and noisy features $Y'$, because the observed equivalences allow us to substitute $X$ by $Y$ and $\widehat{\,X\,}$ by $Y'$.


MINE does not reveal such patterns.\vspace{-2pt}

\newcommand{\rpair}[1]{\rotatebox[origin=c]{0}{\scriptsize$#1$}}
\definecolor{darkgreen}{RGB}{0,100,0} 

\begin{table}[H]
\centering
\scriptsize
\setlength{\tabcolsep}{3pt} 
\renewcommand{\arraystretch}{1}
\begin{tabular}{l c c c c c c}
\toprule[1.2pt]
& \multicolumn{2}{c}{\textbf{Encoder pairs}}
& \multicolumn{2}{c}{\textbf{Decoder pairs}}
& \multicolumn{2}{c}{\textbf{End-to-end}} \\
\cmidrule(lr){2-3}\cmidrule(lr){4-5}\cmidrule(lr){6-7}
\textbf{Objective}
& \rpair{(X, Y')}
& \rpair{(Y, Y')}
& \rpair{(Y', \widehat{X'})}
& \rpair{(\widehat{\,X\,}, \widehat{X'})}
& \rpair{(X, \widehat{\,X\,})}
& \rpair{(X, \widehat{X'})} \\
\midrule
\textcolor{red}{\textbf{NMF-DR}} & 28.57 & 28.68 & 22.37 & 22.38 & 28.64 & 16.58 \\
\textcolor{blue}{\textbf{LOGDET}} & 28.35 & 28.60 & 22.35 & 22.36 & 28.34 &
       16.72 \\
\textcolor{blue}{\textbf{TRACE}}  & 28.46 & 28.66 & 22.40 & 22.40 & 28.44 & 
       19.59* \\  \midrule
\textcolor{orange}{\textbf{MINE}}   & 2.84 & 3.17 & 2.63 & 2.81 & 2.96 &
       2.36 \\  
\textbf{KDE}   & 20.48 & 20.57 & 17.94 & 20.29 & 23.22 & 15.78 \\
\textbf{KICA}   & 10.99 & 16.52 & 10.43 & 16.45 & 17.74 & 12.36 \\
\textbf{HSIC}   & 10.33 & 10.78 & 10.04 & 18.04 & 19.03 & 15.02\\
\bottomrule[1.2pt]
\end{tabular}\vspace{1pt}
\par\noindent\raggedright{\footnotesize *For the trace cost, if the estimator output dimension far exceeds the number of positive singular values, it may introduce an unwanted bias.}\vspace{-7pt}
\caption{Two-moon dataset: measurement comparisons.}
\label{tab:objectives_pairs}
\end{table} 

\vspace{-10pt}

\begin{table}[H]
\centering
\scriptsize
\setlength{\tabcolsep}{3pt}
\renewcommand{\arraystretch}{1}
\begin{tabular}{l c c c c c c}
\toprule[1.6pt]
& \multicolumn{2}{c}{\textbf{Encoder pairs}}
& \multicolumn{2}{c}{\textbf{Decoder pairs}}
& \multicolumn{2}{c}{\textbf{End-to-end}} \\
\cmidrule(lr){2-3}\cmidrule(lr){4-5}\cmidrule(lr){6-7}
\textbf{Objective}
& \rpair{(X, Y')}
& \rpair{(Y, Y')}
& \rpair{(Y', \widehat{X'})}
& \rpair{(\widehat{\,X\,}, \widehat{X'})}
& \rpair{(X, \widehat{\,X\,})}
& \rpair{(X, \widehat{X'})} \\
\midrule
\textcolor{red}{\textbf{NMF-DR}} & 28.81 &  28.85 & 200.74 & 207.90 & 28.81 & 28.51 \\
\textcolor{blue}{\textbf{LOGDET}} & 27.98 & 28.04 & 199.30* & 200.75* & 27.98 & 27.70 \\
\textcolor{blue}{\textbf{TRACE}}  & 28.28 & 28.34 &  201.10* & 206.97* & 28.29 & 28.02 \\ \midrule
\textcolor{orange}{\textbf{MINE}}   & 1.85 & 2.09 & 3.31 & 4.90 & 2.07 & 2.07 \\
\textbf{KDE}    & 2.50 & 3.18 & 3.18 & 4.93 & 3.00 &
       2.96
 \\
\textbf{KICA}   & 9.43 & 16.87 & 9.47 & 276.15 & 270.89 & 344.04 \\
\textbf{HSIC}   & 11.62 & 10.89 & 11.68 & 410.93 & 404.82 & 495.02 \\
\bottomrule[1.6pt]
\end{tabular}\vspace{1pt}
\par\noindent\raggedright{\footnotesize *We need to increase the estimator output dimension to 500 in these cases.}\vspace{-7pt}
\caption{MNIST: measurement comparisons.}
\label{tab:objectives_pairs222}
\end{table}

\vspace{-3pt}

\begin{figure}[H]
  \centering
\begin{subfigure}{.172\textwidth}\includegraphics[width=\linewidth]{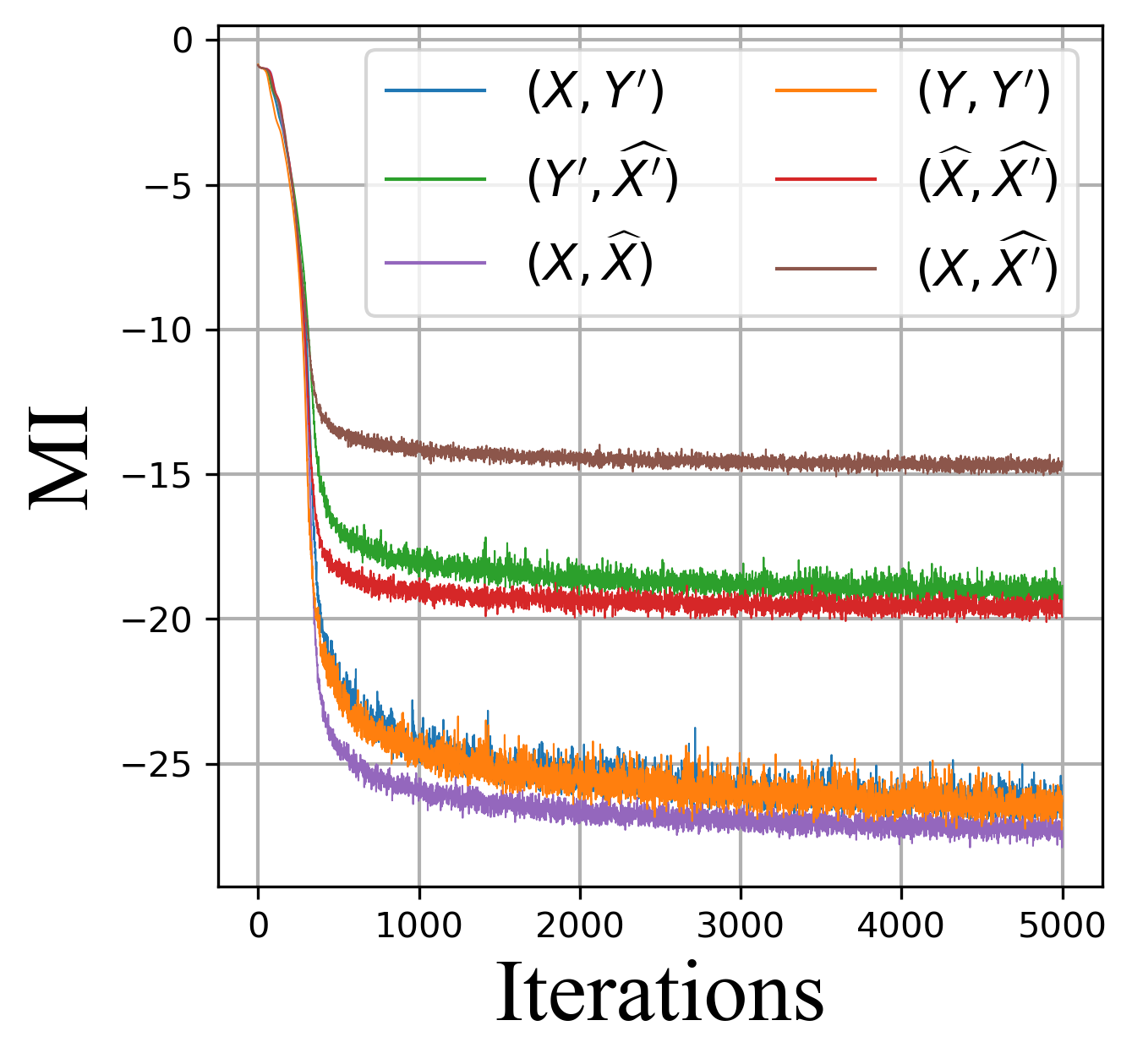}\vspace{-6pt}
\caption{\footnotesize Two-moon curves}
\end{subfigure}\hspace{-6pt}
\begin{subfigure}{.157\textwidth}\includegraphics[width=\linewidth]{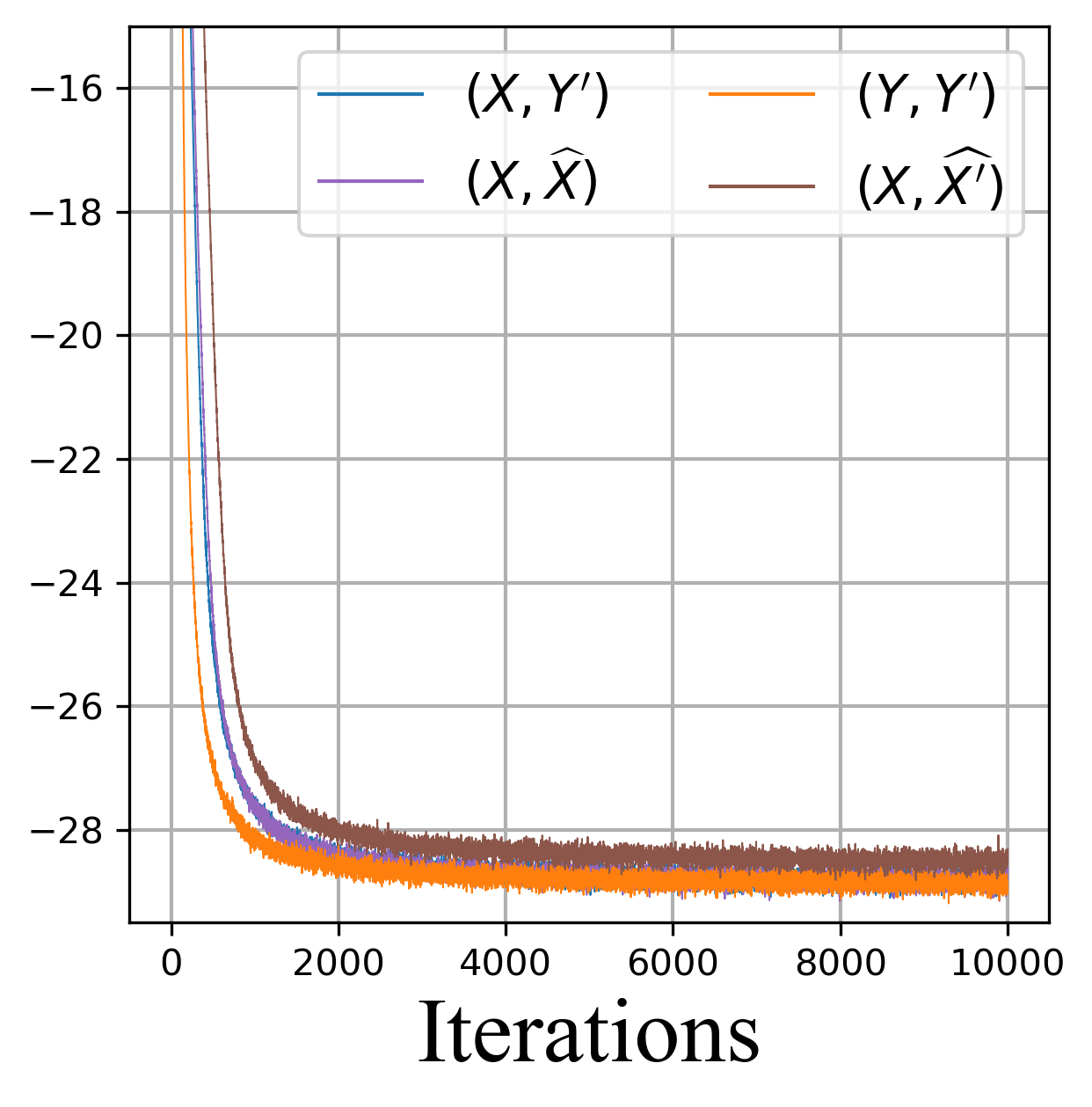}\vspace{-5pt}
\caption{\footnotesize MNIST (4 curves)}
\end{subfigure}\hspace{-4pt}
\begin{subfigure}{.157\textwidth}\includegraphics[width=\linewidth]{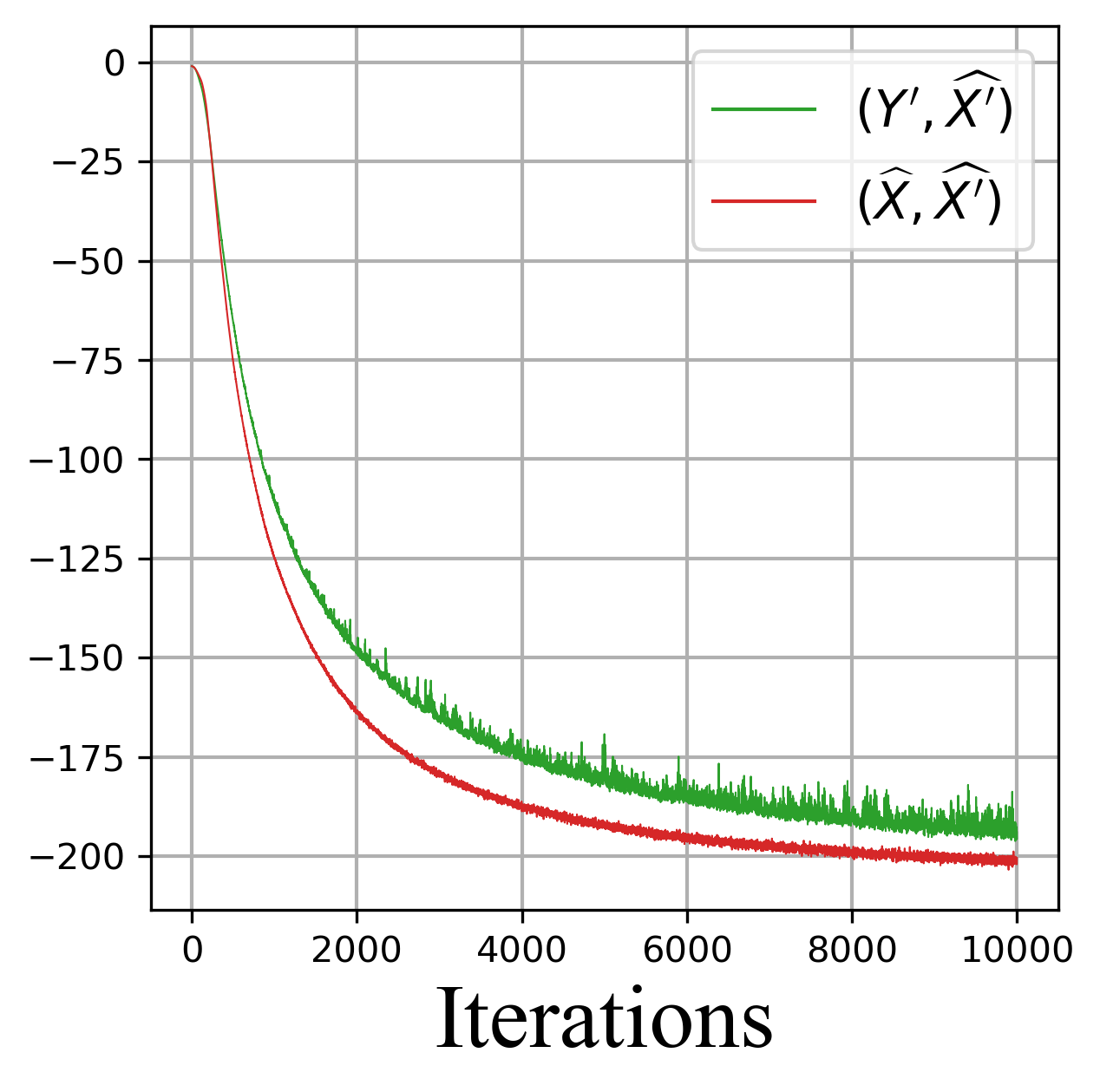}\vspace{-6pt}
\caption{\footnotesize MNIST (2 curves)}
\end{subfigure}\vspace{-5pt}
\caption{Learning curves for the NMF-like cost. The curves are smooth and stable because no re-pairing is required.}
\label{111111111}
\end{figure}

\begin{figure}[H]
\centering
\hspace{-10pt}\begin{minipage}{0.16\textwidth}
  \centering
  \includegraphics[width=\linewidth]{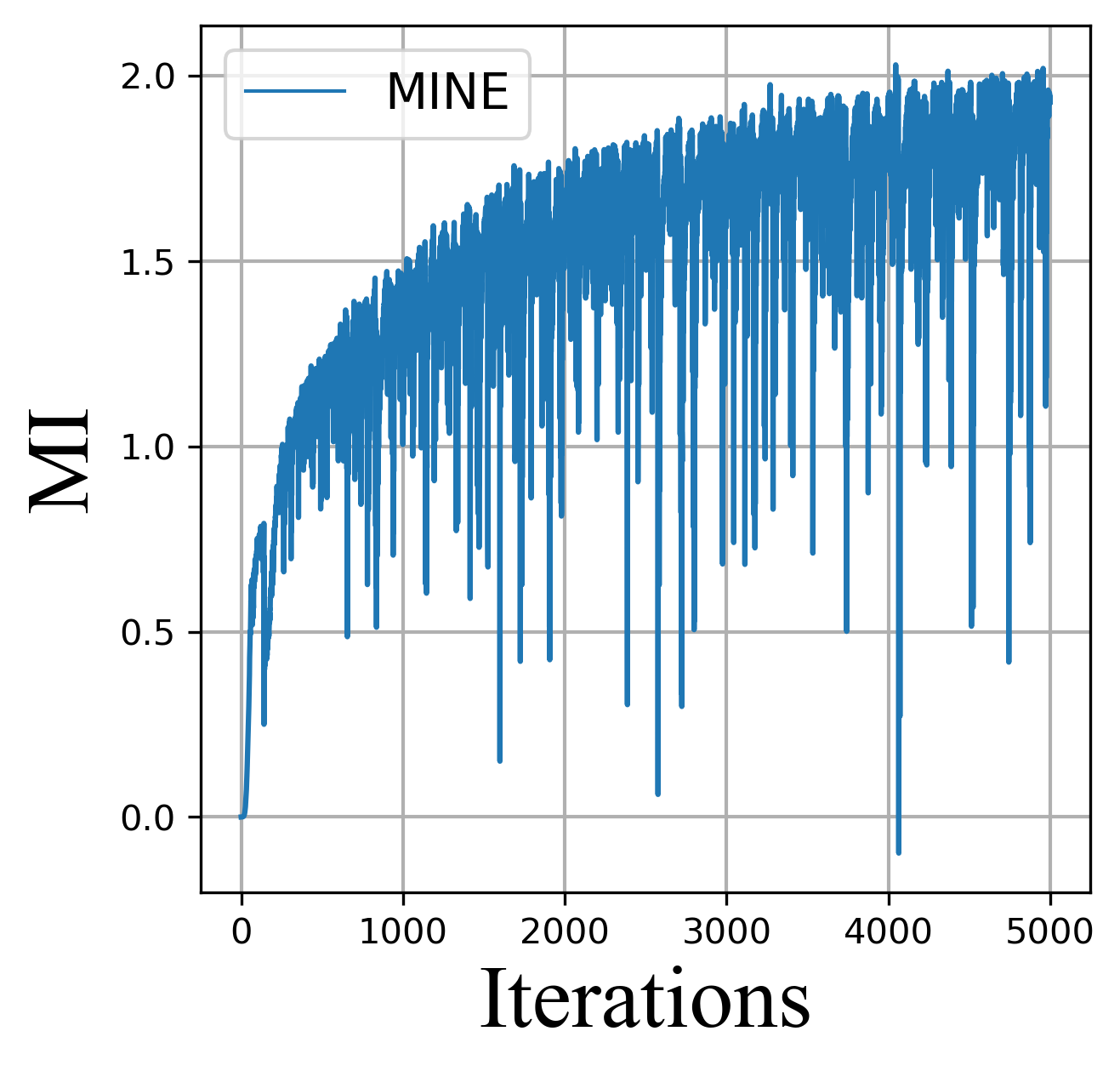}
\end{minipage}\hspace{5pt}
\begin{minipage}{0.32\textwidth}
  \caption{A learning curve of MINE on MNIST. The sudden ``dip'' in the curve is largely due to the re-pairing step for sampling from the product of marginals. The learning curve would be smoother if we lowered the learning rate, but convergence would take significantly longer.}
  \label{22222222}
\end{minipage}
\end{figure}

\noindent \textbf{Result 2.} Learning curves are shown in Fig.~\ref{111111111} for our costs and in Fig.~\ref{22222222} for MINE. Our costs learn more smoothly and stably, since we avoid MINE’s re-pairing step used for sampling from the product of marginals.

\vspace{-5pt}

\vspace{-10pt}

\noindent \textbf{Result 3.} The previous experiments fix the feature-noise variance $v_p = 10^{-4}$. For completeness, we sweep $v_p$ from $10^{-7}$ to $10^{-1}$: results for two-moon are reported in Table~\ref{tab:vp_sweep_mnist999}, and for MNIST in Appendix Table~\ref{tab:vp_sweep_mnistttt}. As the noise level increases, the MSE increases (performance degrades) and the estimated dependence decays. However, the substitution pattern discussed above still holds across the sweep. We can guarantee that in this example the estimator is unbiased.

\begin{table}[H]
\centering
\scriptsize
\setlength{\tabcolsep}{1pt}
\renewcommand{\arraystretch}{1}
\begin{tabular}{c c c c c c c c}
\toprule[1.2pt]
\multicolumn{1}{c}{\textbf{$v_p$}} &
\multicolumn{1}{c}{\textbf{MSE}} &
\multicolumn{2}{c}{\textbf{Encoder pairs}} &
\multicolumn{2}{c}{\textbf{Decoder pairs}} &
\multicolumn{2}{c}{\textbf{End-to-end}} \\
\cmidrule(lr){3-4}\cmidrule(lr){5-6}\cmidrule(lr){7-8}
& \multicolumn{1}{c}{$(\times 10^{-3})$} &
\multicolumn{1}{c}{\rpair{(X, Y')}} &
\multicolumn{1}{c}{\rpair{(Y, Y')}} &
\multicolumn{1}{c}{\rpair{(Y', \widehat{X'})}} &
\multicolumn{1}{c}{\rpair{(\widehat{\,X\,}, \widehat{X'})}} &
\multicolumn{1}{c}{\rpair{(X, \widehat{\,X\,})}} &
\multicolumn{1}{c}{\rpair{(X, \widehat{X'})}} \\
\midrule
$10^{-7}$ & 0.540 & 519.68 & 509.61 & 46.55 & 46.79 & 506.74 & 45.84 \\
$10^{-6}$ & 0.624 & 213.59 & 212.88 & 40.05 & 40.25 & 216.28 & 37.45 \\
$10^{-5}$ & 0.650 &  86.87 &  87.25 & 38.71 & 38.84 &  87.06 & 31.90 \\
$10^{-4}$ & 1.10 &  28.78 &  28.83 & 20.49 & 20.52 &  28.80 & 15.37 \\
$10^{-3}$ & 2.46 &   9.82 &   9.82 &  9.03 &  9.04 &   9.82 &  6.69 \\
$10^{-2}$ & 7.18 &   3.77 &   3.77 &  3.76 &  3.76 &   3.77 &  2.73 \\
$10^{-1}$ & 14.6  &   1.85 &   1.85 &  1.83 &  1.83 &   1.85 &  1.54 \\

\bottomrule[1.2pt]
\end{tabular}\vspace{-6pt}
\caption{Two-moon: dependence vs.\ noise levels (NMF-like).}
\label{tab:vp_sweep_mnist999}
\end{table}
\vspace{-3pt}

\noindent \textbf{Result 4.} The previous experiments fix the feature dimension to 1D (the encoder maps the data to a 1D features). We now sweep it. This is uninformative for two-moon due to its low data dimension, so we report MNIST results.

As the feature dimension increases, the MSE decreases and the estimated dependence increases for all variable pairs. The substitution pattern remains. We can only guarantee unbiased, accurate estimates up to dimension 3, since the total number of used singular functions is capped by the decoder output dimension (2000).

\begin{table}[H]
\centering
\scriptsize
\setlength{\tabcolsep}{2.0pt}
\renewcommand{\arraystretch}{1.10}
\begin{tabular}{c c c c c c c c}
\toprule[1.2pt]
\multicolumn{1}{c}{\textbf{Fea. Dim.}} &
\multicolumn{1}{c}{\textbf{MSE}} &
\multicolumn{2}{c}{\textbf{Encoder pairs}} &
\multicolumn{2}{c}{\textbf{Decoder pairs}} &
\multicolumn{2}{c}{\textbf{End-to-end}} \\
\cmidrule(lr){3-4}\cmidrule(lr){5-6}\cmidrule(lr){7-8}
\multicolumn{1}{c}{} &
\multicolumn{1}{c}{} &
\multicolumn{1}{c}{\rpair{(X, Y')}} &
\multicolumn{1}{c}{\rpair{(Y, Y')}} &
\multicolumn{1}{c}{\rpair{(Y', \widehat{X'})}} &
\multicolumn{1}{c}{\rpair{(\widehat{\,X\,}, \widehat{X'})}} &
\multicolumn{1}{c}{\rpair{(X, \widehat{\,X\,})}} &
\multicolumn{1}{c}{\rpair{(X, \widehat{X'})}} \\
\midrule
1 & 0.048 &  28.81 &  28.85 & 200.74 & 207.90 &  28.81 &  28.51 \\
2 & 0.030 & 648.40 & 675.31 & 1632.26 & 1661.05 & 647.28 & 634.45 \\
3 & 0.021 & 1604.27 & 1679.04 & 1859.84 & 1819.99 & 1576.40 & 1623.15 \\
4 & 0.016 & 1886.03 & 1885.60 & 1899.80 & 1839.93 & 1846.92 & 1857.05 \\
5 & 0.012 & 1929.32 & 1875.47 & 1914.92 & 1840.91 & 1892.85 & 1886.75 \\
\bottomrule[1.2pt]
\end{tabular}\vspace{-6pt}
\caption{MNIST: dependence vs.\ feature dims (NMF-like).}
\label{tab:trajectory_pairs9999}
\end{table}

\vspace{-20pt}

\section{Conclusion}


The experiments we show demonstrate that we can go beyond end-to-end mean-squared error, giving statistical dependence measures among many autoencoder components such as data, features, and reconstructions, under a Gaussian assumption.

The experiments here are only a fraction of the full set. The appendix will continue with additional experiments.

\bibliography{reference}

\appendix

\section{Extended Experimental Results}


\subsection{Extra tables}\label{section_extra_tables}

First, we present a few additional tables that continue from the results reported in the main text.\vspace{7pt}

\noindent \textbf{Result 5.} We have shown a consistent pattern: smaller Gaussian variance corresponds to smaller MSE and higher statistical dependence. This suggests a conjugated relationship among Gaussian variance, MSE, and statistical dependence.

During autoencoder training, the cost minimizes the MSE. At each iteration, the current MSE can be interpreted as inducing an effective (optimal) Gaussian radius for the reconstructions. As training progresses and the MSE decreases, these Gaussian balls shrink, and the statistical dependence increases. This motivates the question of whether a similar ``ball-shrinking'' process also occurs for the features, not just the reconstructions.

To investigate this, we stop training at 100, 200, 300, 400, 500, and 600 iterations. At each checkpoint, we measure the statistical dependence values for all pairs and examine whether any patterns emerge. Table~\ref{tab:trajectory_pairs4555555} corresponds to the two-moons dataset, and Table~\ref{tab:trajectory_pairs888888} corresponds to MNIST.




\begin{table}[H]
\centering
\scriptsize
\setlength{\tabcolsep}{2.0pt}
\renewcommand{\arraystretch}{1.10}
\begin{tabular}{r c c c c c c c}
\toprule[1.2pt]
\multicolumn{1}{c}{\small \textbf{Iters}} &
\multicolumn{1}{c}{\small \textbf{MSE}} &
\multicolumn{2}{c}{\textbf{Encoder pairs}} &
\multicolumn{2}{c}{\textbf{Decoder pairs}} &
\multicolumn{2}{c}{\textbf{End-to-end}} \\
\cmidrule(lr){3-4}\cmidrule(lr){5-6}\cmidrule(lr){7-8}
\multicolumn{1}{c}{$\times 10^{2}$} &
\multicolumn{1}{c}{$\times 10^{-3}$} &
\multicolumn{1}{c}{\rpair{(X, Y')}} &
\multicolumn{1}{c}{\rpair{(Y, Y')}} &
\multicolumn{1}{c}{\rpair{(Y', \widehat{X'})}} &
\multicolumn{1}{c}{\rpair{(\widehat{\,X\,}, \widehat{X'})}} &
\multicolumn{1}{c}{\rpair{(X, \widehat{\,X\,})}} &
\multicolumn{1}{c}{\rpair{(X, \widehat{X'})}} \\
\midrule
0 & 32.0 & 0.997& 0.998 & 0.997& 0.997& 0.997 & 0.997 \\
1 &  1.88 & 23.10 & 24.03 &  9.39 &  9.49 & 23.24 &  8.38 \\
2 &  1.64 & 24.20 & 24.94 & 13.26 & 13.44 & 24.98 & 11.27 \\
3 &  1.56 & 24.16 & 25.29 & 13.98 & 14.24 & 25.42 & 11.83 \\
4 &  1.47 & 24.68 & 25.50 & 14.66 & 14.97 & 25.46 & 12.23 \\
5 &  1.40 & 25.01 & 25.43 & 15.08 & 15.45 & 25.77 & 12.54 \\
6 &  1.37 & 24.98 & 25.68 & 15.07 & 15.39 & 25.79 & 12.46 \\
\bottomrule[1.2pt]
\end{tabular}\vspace{-6pt}
\caption{Two-moon dataset: dependence vs.\ iterations.}
\label{tab:trajectory_pairs4555555}
\end{table} \vspace{-20pt}
\begin{table}[H]
\centering
\scriptsize
\setlength{\tabcolsep}{2.0pt}
\renewcommand{\arraystretch}{1.10}
\begin{tabular}{c c c c c c c c}
\toprule[1.2pt]
\multicolumn{1}{c}{\small \textbf{Iters}} &
\multicolumn{1}{c}{\small \textbf{MSE}} &
\multicolumn{2}{c}{\textbf{Encoder pairs}} &
\multicolumn{2}{c}{\textbf{Decoder pairs}} &
\multicolumn{2}{c}{\textbf{End-to-end}} \\
\cmidrule(lr){3-4}\cmidrule(lr){5-6}\cmidrule(lr){7-8}
\multicolumn{1}{c}{$\times 10^{2}$} &
\multicolumn{1}{c}{} &
\multicolumn{1}{c}{\rpair{(X, Y')}} &
\multicolumn{1}{c}{\rpair{(Y, Y')}} &
\multicolumn{1}{c}{\rpair{(Y', \widehat{X'})}} &
\multicolumn{1}{c}{\rpair{(\widehat{\,X\,}, \widehat{X'})}} &
\multicolumn{1}{c}{\rpair{(X, \widehat{\,X\,})}} &
\multicolumn{1}{c}{\rpair{(X, \widehat{X'})}} \\
\midrule
0 & 0.239  & 0.998& 0.997 & 0.997& 0.998& 0.997 & 0.997 \\
1 & 0.119 & 25.00 & 24.96 & 28.84 & 28.96 & 25.00 & 17.74 \\
2 & 0.089 & 27.12 & 27.06 & 70.17 & 71.06 & 27.06 & 24.78 \\
3 & 0.077 & 28.29 & 28.25 & 97.60 & 99.55 & 28.26 & 26.92 \\
4 & 0.071 & 28.46 & 28.45 & 110.14 & 113.46 & 28.44 & 27.43 \\
5 & 0.067 & 28.42 & 28.34 & 117.90 & 122.06 & 28.38 & 27.44 \\
6 & 0.065 & 28.51 & 28.50 & 124.77 & 129.50 & 28.49 & 27.71 \\
\bottomrule[1.2pt]
\end{tabular}\vspace{-6pt}
\caption{MNIST: dependence vs.\ iterations.}
\label{tab:trajectory_pairs888888}
\end{table}

\noindent First, the substitution pattern, i.e., the equality of the dependence values for the relevant pairs, holds at every training iteration, starting from initialization. Second, the statistical dependence increases for all variable pairs as training proceeds, which supports our argument. In other words, although the cost minimizes the end-to-end MSE, the statistical dependence between the data and the learned features, considering the encoder alone, also increases during training. Moreover, training begins from a state of statistical independence: the lower bound of the cost is $1$, and attaining this lower bound implies statistical independence.



\vspace{10pt}

In the main text, we presented a sweep over the encoder feature variance $v_p$ for the toy dataset. We also performed the same experiment on MNIST. The results are reported in Table~\ref{tab:vp_sweep_mnistttt}.  

A notable observation is that the encoder-side statistical dependence, which is bounded by \(v_p\), is very similar across the toy dataset (Table~\ref{tab:vp_sweep_mnist999}) and MNIST (Table~\ref{tab:vp_sweep_mnistttt}). The corresponding values are close despite the differences in dataset characteristics and input dimensionality. This suggests that the encoder-side dependence pairs \(\{X, Y'\}\) and \(\{Y, Y'\}\), as well as the reconstruction pair \(\{X, \widehat{\,X\,}\}\), may be largely irrelevant to the dataset, depending primarily on the feature dimension and the variance parameter $v_p$.

\begin{table}[H]
\centering
\scriptsize
\setlength{\tabcolsep}{2.0pt}
\renewcommand{\arraystretch}{1.10}
\begin{tabular}{c c c c c c c c}
\toprule[1.2pt]
\multicolumn{1}{c}{\textbf{$v_p$}} &
\multicolumn{1}{c}{\textbf{MSE}} &
\multicolumn{2}{c}{\textbf{Encoder pairs}} &
\multicolumn{2}{c}{\textbf{Decoder pairs}} &
\multicolumn{2}{c}{\textbf{End-to-end}} \\
\cmidrule(lr){3-4}\cmidrule(lr){5-6}\cmidrule(lr){7-8}
& \multicolumn{1}{c}{} &
\multicolumn{1}{c}{\rpair{(X, Y')}} &
\multicolumn{1}{c}{\rpair{(Y, Y')}} &
\multicolumn{1}{c}{\rpair{(Y', \widehat{X'})}} &
\multicolumn{1}{c}{\rpair{(\widehat{\,X\,}, \widehat{X'})}} &
\multicolumn{1}{c}{\rpair{(X, \widehat{\,X\,})}} &
\multicolumn{1}{c}{\rpair{(X, \widehat{X'})}} \\
\midrule
$10^{-7}$ & 0.035 & 781.71 & 723.36 & 868.93 & 877.61 & 762.46 & 625.97 \\
$10^{-6}$ & 0.036 & 275.65 & 272.43 & 793.28 & 840.97 & 278.31 & 265.03 \\
$10^{-5}$ & 0.041 & 88.56 & 88.33 & 387.96 & 422.73 & 88.86 & 86.90 \\
$10^{-4}$ & 0.046 & 28.87 & 28.88 & 200.74 & 207.90 & 28.87 & 28.87 \\
$10^{-3}$ & 0.051 &  9.83 &  9.83 & 65.84 & 67.29 &  9.83 &  9.73 \\
$10^{-2}$ & 0.058 &  3.77 &  3.77 & 21.78 & 21.95 &  3.77 &  3.73 \\
$10^{-1}$ & 0.063 & 1.84 & 1.84 & 6.96 & 6.96 & 1.84 & 1.81 \\
\bottomrule[1.2pt]
\end{tabular}\vspace{-6pt}
\caption{MNIST: dependence vs.\ noise levels.}
\label{tab:vp_sweep_mnistttt}
\end{table}


\noindent \textbf{Concatenating noise to the inputs does not work.} The paper argues that statistical dependence cannot be meaningfully measured in a static, deterministic, noise-free setting. To make the quantity well-defined, one can assume a small Gaussian ball around the features. A natural alternative is to inject noise through the inputs: we concatenate each data sample with additional dimensions filled with i.i.d. uniform random noise as inputs to the encoder. We wonder if this input-noise concatenation has the same effect as adding Gaussian noise directly to the features. 

We compare dependence estimates for two variable pairs under three conditions: the static noise-free setting, concatenated input noise, and additive feature noise with variance $v_p = 10^{-7}$. Unfortunately, concatenating input noise does not solve the issue of over-determination: the dependence estimates remain essentially identical to the noise-free case and diverge to very large values, making it unclear whether the estimator is biased or simply ill-posed. In contrast, even a small additive feature noise ($v_p = 10^{-7}$) immediately yields smaller, finite, and measurable dependence values, while only slightly degrading reconstruction performance (MSE $0.417 \rightarrow 0.424$). 

We suspect that Gaussian noise is still implicitly associated with the nominally noise-free case. The key difference is that this noise is not injected into the model; instead, it exists ``outside'' the model, assumed rather than explicitly presented, much like the additive noise assumed for reconstructions, whose optimal value corresponds to the MSE. Without such an assumption, there may be no principled way to measure statistical dependence. Learning seeks to shrink the radius of this Gaussian ball. Learning can be viewed as shrinking the radius of this Gaussian ball.




\begin{table}[H]
\centering
\scriptsize
\setlength{\tabcolsep}{2.0pt}
\renewcommand{\arraystretch}{1.10}
\begin{tabular}{c c c c c c c c}
\toprule[1.2pt]
\multicolumn{1}{c}{\textbf{Noise type}} &
\multicolumn{1}{c}{\textbf{MSE}} &
\multicolumn{2}{c}{\textbf{NMF-like cost}} &
\multicolumn{2}{c}{\textbf{MINE}} \\
\cmidrule(lr){3-4}\cmidrule(lr){5-6}
& \multicolumn{1}{c}{} &
\multicolumn{1}{c}{\rpair{(X, Y')}} &
\multicolumn{1}{c}{\rpair{(X, \widehat{\,X\,})}} &
\multicolumn{1}{c}{\rpair{(X, Y')}} &
\multicolumn{1}{c}{\rpair{(X, \widehat{\,X\,})}} \\
\midrule 
Concatenated input noise & 0.417 & 1508.45 & 1793.36 & 2.83 & 4.14  \\
No noise & 0.417 & 1507.52 & 1782.98 & 4.68 & 5.45\\
Additive feature noise $10^{-7}$ & 0.424 & 608.29 & 583.30 & 4.64 & 5.20\\ 
\bottomrule[1.2pt]
\end{tabular}\vspace{-6pt}
\caption{Concatenating input noise does not solve the issue that the variable pairs are overly dependent, which essentially makes the dependence unmeasurable. MINE is very unstable and does not produce a meaningful result.}
\label{tab:table}
\end{table}

\subsection{Eigenanalysis}\label{eigenanalysis}

In the main paper we primarily report numerical results for the proposed statistical dependence measures. Recall that our method is based on an orthonormal decomposition of the density ratio $\frac{p(X,Y)}{p(X)\,p(Y)} \;=\; \sum_{k=1}^{K} \sqrt{\lambda_k} \cdot \widehat{\phi_k}(X) \cdot \widehat{\psi_k}(Y).$ 

Apart from the statistical dependence measure itself (i.e., the sum of the singular values), the decomposition also comes with singular values and singular functions. In this section, we therefore visualize the learned singular spectrum and singular functions.

To extract the singular values and singular functions from the trained networks, we  need the following steps:\vspace{9pt}

\noindent 1. Compute correlation matrices. Estimate the autocorrelation matrices $\mathbf{R}_F = \mathbb{E}\left[ \mathbf{f}(X)\mathbf{f}^\intercal (X)\right]$ and $\mathbf{R}_G = \mathbb{E} \left[\mathbf{g}(Y) \mathbf{g}^\intercal (Y)\right]$, and the cross-correlation matrix $\mathbf{P} = \mathbb{E} \left[ \mathbf{f}(X) \mathbf{g}^\intercal (Y)\right]$. \vspace{5pt}

\noindent 2. Whiten the estimator network outputs. Normalize the network features with $\overline{\mathbf{f}} = \mathbf{R}_F^{-1/2}\mathbf{f}$ and $\overline{\mathbf{g}} = \mathbf{R}_G^{-1/2}\mathbf{g}.$\vspace{5pt}

\noindent 3. Recalculate the cross-correlation in the whitened space: $\overline{\mathbf{P}} \;=\; \mathbb{E}\!\left[\overline{\mathbf{f}}(X)\,\overline{\mathbf{g}}^\intercal(Y)\right]$.\vspace{5pt}

\noindent 4. Perform SVD: $\overline{\mathbf{P}} \;=\; \mathbf{Q}_F\,\mathbf{\Lambda}^{1/2}\,\mathbf{Q}_G^\intercal$.\vspace{5pt}

\noindent 5. Rotations: $\widehat{\,\mathbf{f}\,} = \mathbf{Q}_F^\intercal \overline{\mathbf{f}}$ and $\widehat{\,\mathbf{g}\,} = \mathbf{Q}_G^\intercal \overline{\mathbf{g}}$.\vspace{9pt}

\noindent The resulting $\widehat{\,\mathbf{f}\,}$ and $\widehat{\,\mathbf{g}\,}$ estimates the singular functions of the density ratio. The diagonal entries of $\mathbf{\Lambda}^{1/2}$ from the SVD of $\overline{\mathbf{P}}$ estimate the leading singular values of the density ratio. Formal justification is provided in the next appendix section.\vspace{12pt}

\begin{figure}[H]
  \centering
\begin{subfigure}{.25\textwidth}\includegraphics[width=\linewidth]{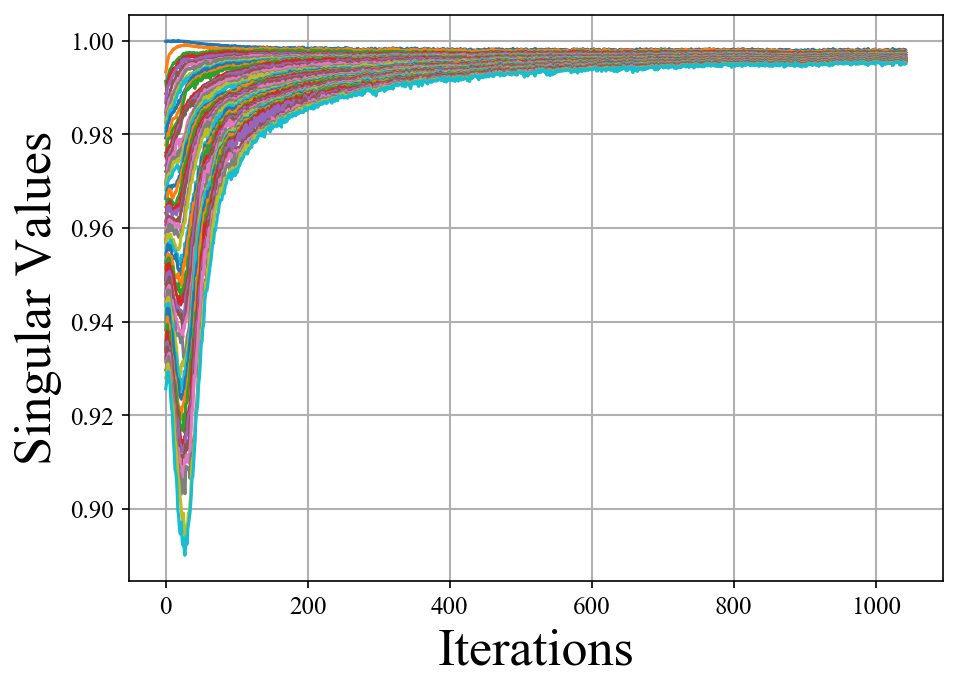}\vspace{-7pt}
\caption{No $v_p$ assumed}
\label{3a}
\end{subfigure}\hspace{-5pt}
\begin{subfigure}{.23\textwidth}\includegraphics[width=\linewidth]{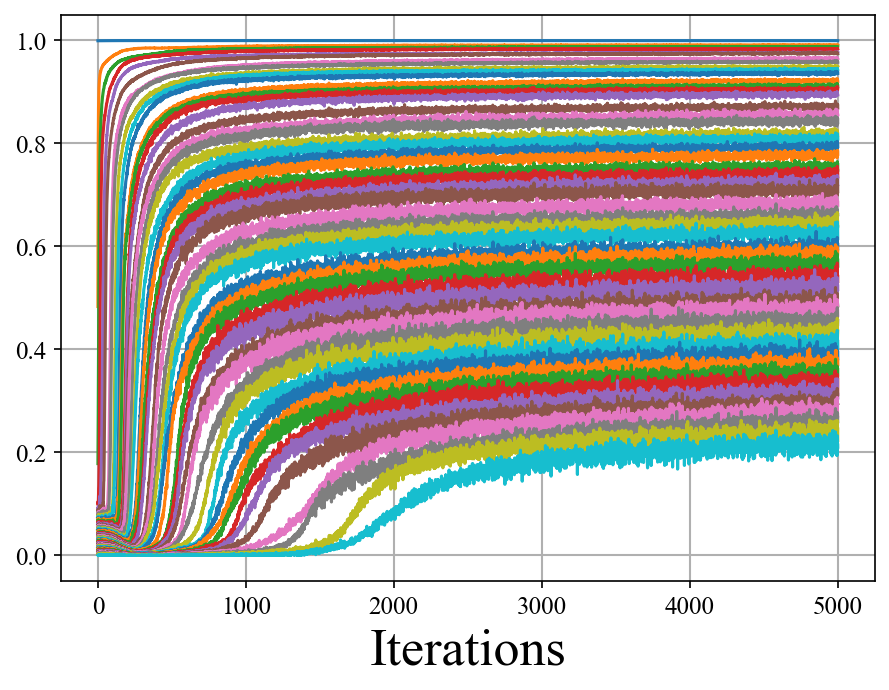}\vspace{-7pt}
\caption{$v_p$ assumed}
\label{3b}
\end{subfigure}\vspace{-5pt}
\caption{Learning curves of singular values.}
\label{learning_curves_singular_values}
\end{figure}

\noindent \textbf{Learning curves of singular values.} Fig.~\ref{learning_curves_singular_values} shows the singular values during training. The result is for the pair $\{X,Y'\}$ for the toy dataset. At each iteration, we estimate the singular values and plot the trajectories of the top 50. We consider two settings: (a) a static, deterministic, noise-free setting; and (b) a setting in which additive noise with \(v_p = 10^{-4}\) assumed and added for the features. Each curve represents one singular value.

In the noise-free setting, the top 50 singular values rapidly collapse to \(1\), the upper bound. This saturation makes it difficult to determine whether the estimator is unbiased because there are too many repetitive singular values. When \(v_p = 10^{-4}\), the spectrum is more reasonable: the singular values are spread over \([0,1]\), and we can say that our estimator is indeed unbiased and accurate. 

\begin{figure}[H]
\centering
\begin{subfigure}{.24\textwidth}
\centering
\includegraphics[width=.75\linewidth]{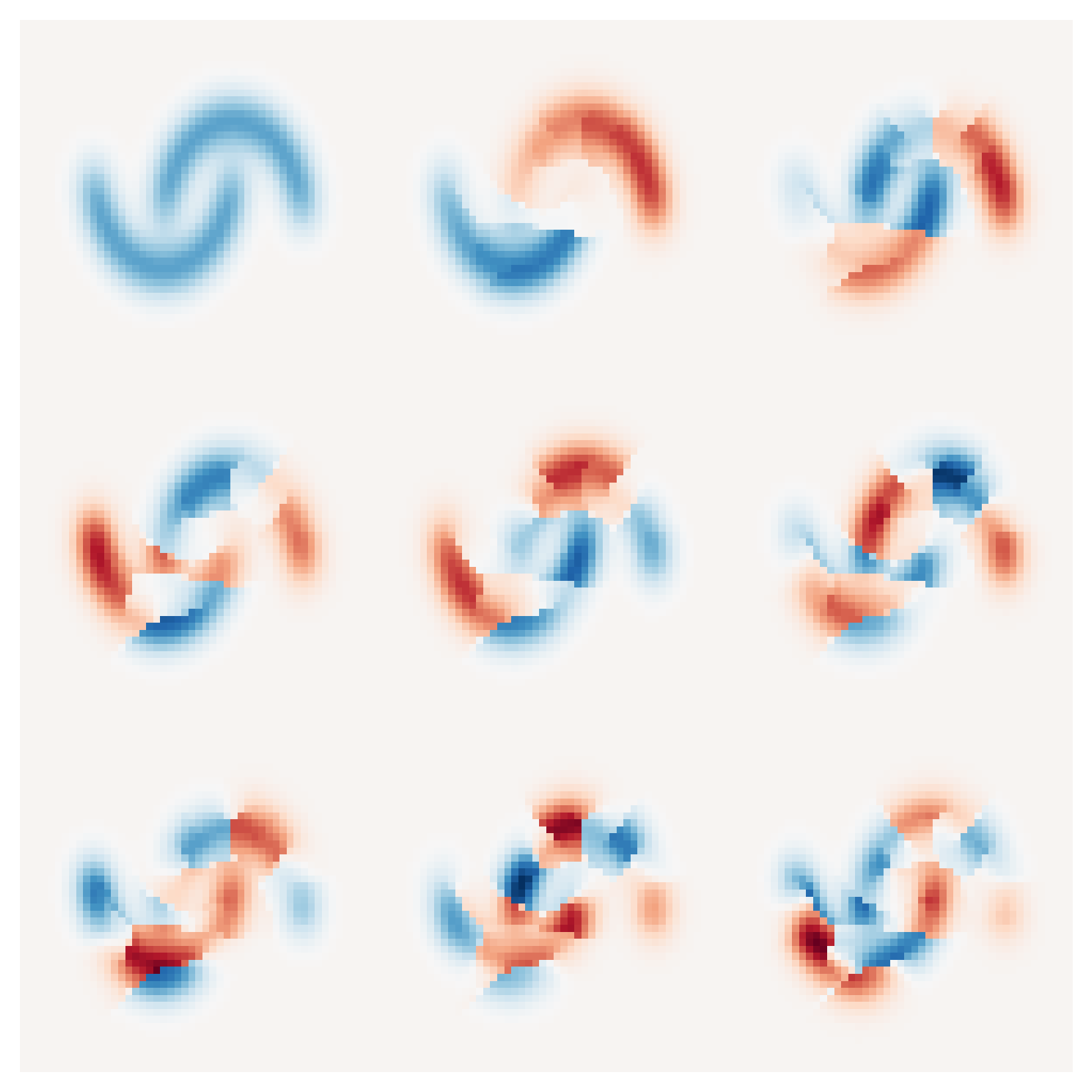}
\caption{\footnotesize Left singular functions $\{X,Y'\}$}
\end{subfigure}%
\begin{subfigure}{.24\textwidth}
\centering
\hspace{-10pt}\includegraphics[width=.75\linewidth]{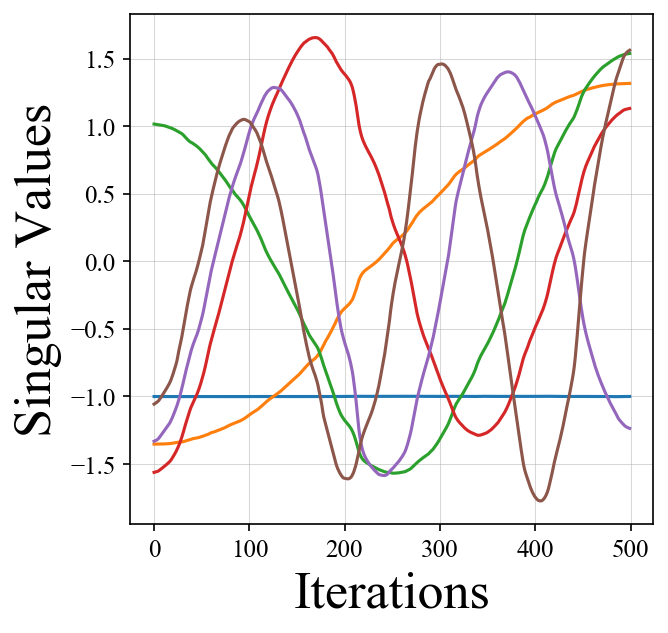}
\caption{\footnotesize Right singular functions $\{X,Y'\}$}
\end{subfigure}
\begin{subfigure}{.24\textwidth}
\centering
\includegraphics[width=.75\linewidth]{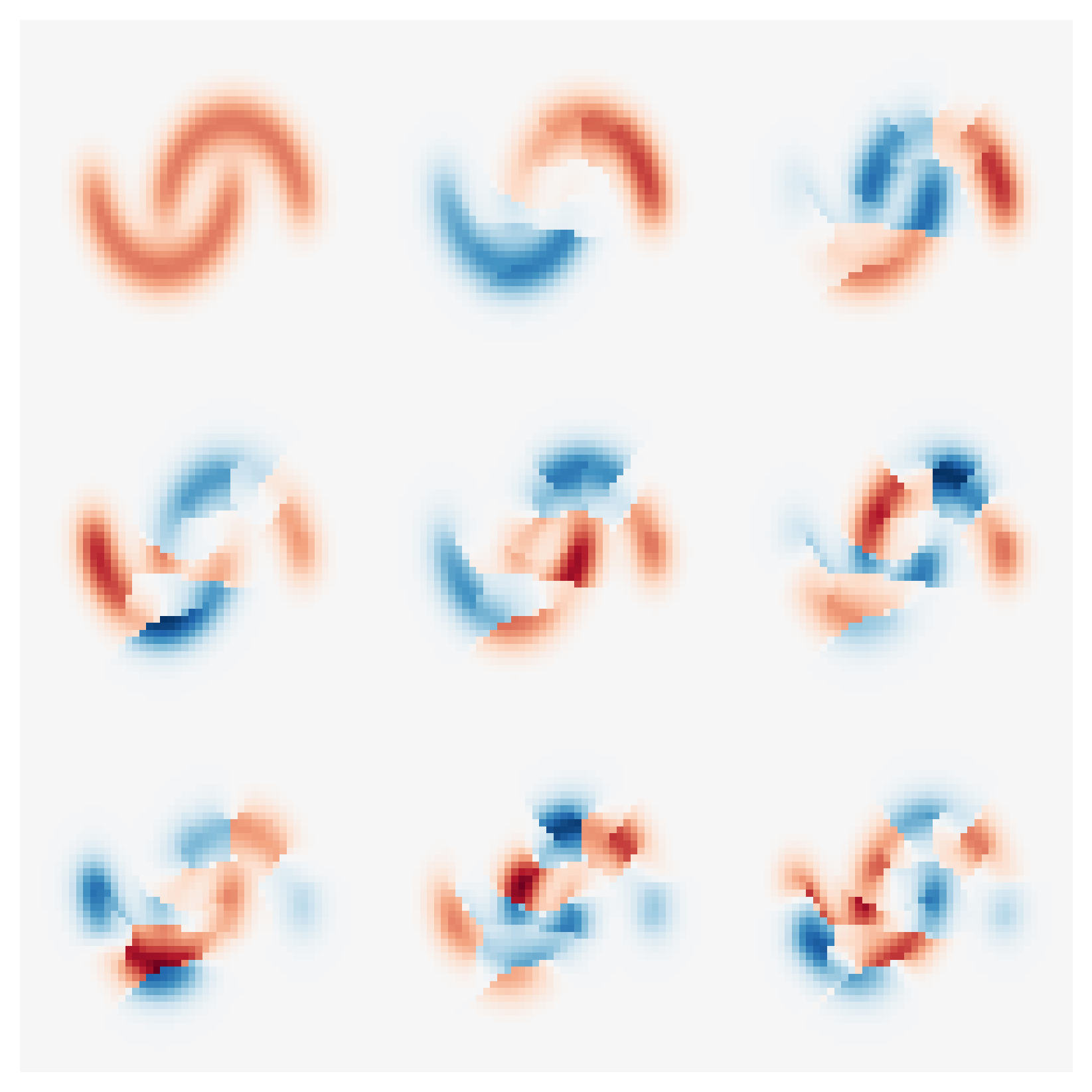}
\caption{\footnotesize Left singular functions $\{X, \widehat{\,X\,}\}$}
\end{subfigure}%
\begin{subfigure}{.24\textwidth}
\centering
\includegraphics[width=.75\linewidth]{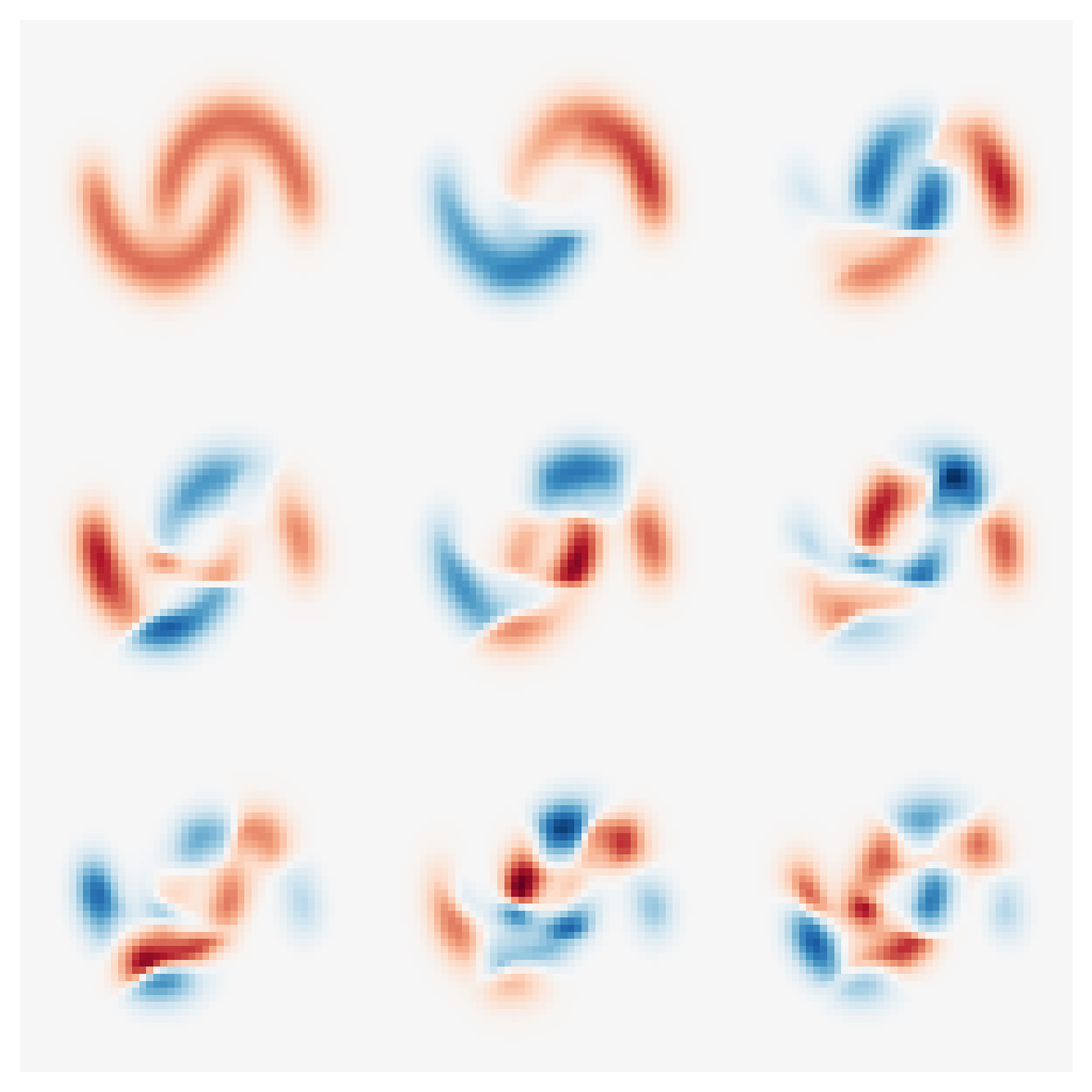}
\caption{\footnotesize Right singular functions $\{X, \widehat{\,X\,}\}$}
\end{subfigure}
\caption{Two-moon dataset: left and right singular functions for the pairs $\{X,Y'\}$ and $\{X,\widehat{\,X\,}\}$. (a), (c), and (d) display 2D singular functions as heatmaps (nine functions shown per panel). (b) displays 1D singular functions as curves (six functions shown).}
\label{interpolated_samples999999999}
\end{figure}

\begin{figure}[H]
  \centering
\begin{subfigure}{.15\textwidth}\includegraphics[width=\linewidth]{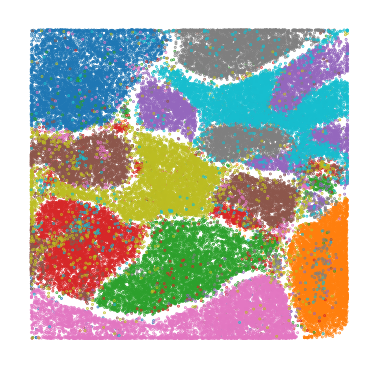}\vspace{-5pt}
\caption*{\footnotesize Visualized features}
\end{subfigure}\hspace{-10pt}
\begin{subfigure}{.15\textwidth}\includegraphics[width=\linewidth]{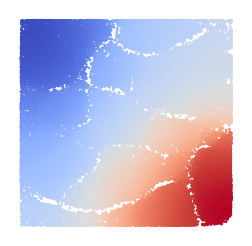}\vspace{-5pt}
\caption*{1st singular}
\end{subfigure}\hspace{-10pt}
\begin{subfigure}{.15\textwidth}\includegraphics[width=\linewidth]{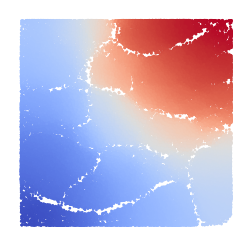}\vspace{-5pt}
\caption*{2nd singular}
\end{subfigure}\hspace{-10pt}
\begin{subfigure}{.15\textwidth}\includegraphics[width=\linewidth]{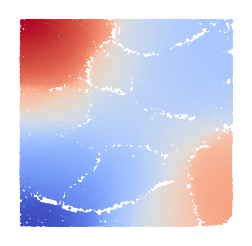}\vspace{-5pt}
\caption*{3rd singular}
\end{subfigure}\hspace{-10pt}
\begin{subfigure}{.15\textwidth}\includegraphics[width=\linewidth]{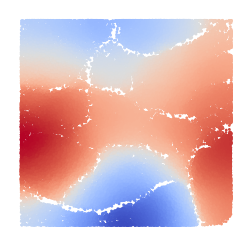}\vspace{-5pt}
\caption*{4th singular}
\end{subfigure}\hspace{-10pt}
\begin{subfigure}{.15\textwidth}\includegraphics[width=\linewidth]{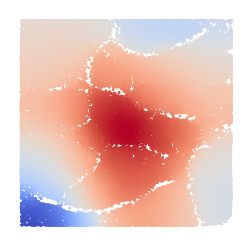}\vspace{-5pt}
\caption*{5th singular}
\end{subfigure}\hspace{-10pt}
\begin{subfigure}{.15\textwidth}\includegraphics[width=\linewidth]{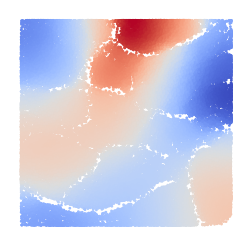}\vspace{-5pt}
\caption*{6th singular}
\end{subfigure}\hspace{-10pt}
\begin{subfigure}{.15\textwidth}\includegraphics[width=\linewidth]{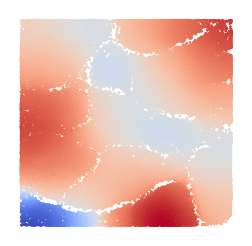}\vspace{-5pt}
\caption*{7th singular}
\end{subfigure}\hspace{-10pt}
\begin{subfigure}{.15\textwidth}\includegraphics[width=\linewidth]{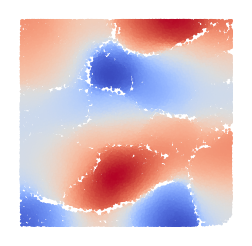}\vspace{-5pt}
\caption*{8th singular}
\end{subfigure}\\
{\footnotesize (a) Right singular functions $\{X,Y'\}$.}\vspace{8pt}
\begin{subfigure}{.25\textwidth}
\centering
\includegraphics[width=\linewidth]{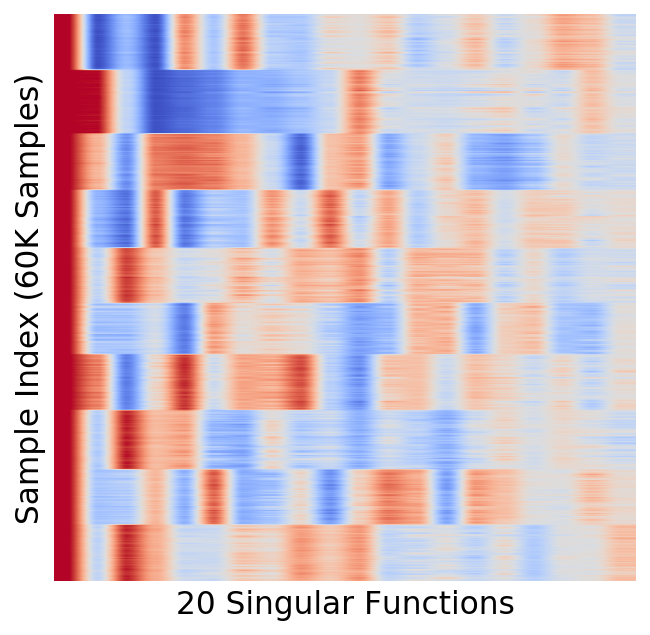}
\caption*{}
\end{subfigure}\\ \vspace{-17pt}
{\footnotesize (b) Left singular functions $\{X,Y'\}$.}\vspace{7pt}
\begin{subfigure}{.245\textwidth}
\centering
\includegraphics[width=\linewidth]{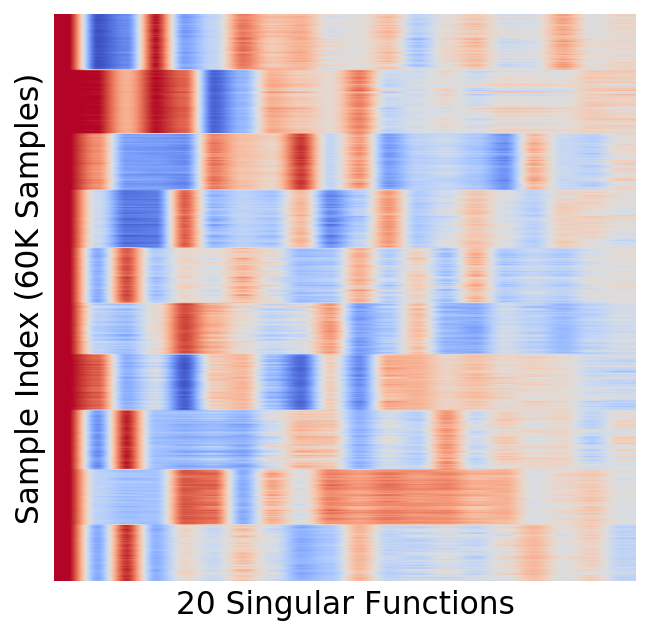}\vspace{-2pt}
\caption{\footnotesize Left singular functions $\{X, \widehat{\,X\,}\}$.}
\end{subfigure}%
\begin{subfigure}{.245\textwidth}
\centering
\includegraphics[width=\linewidth]{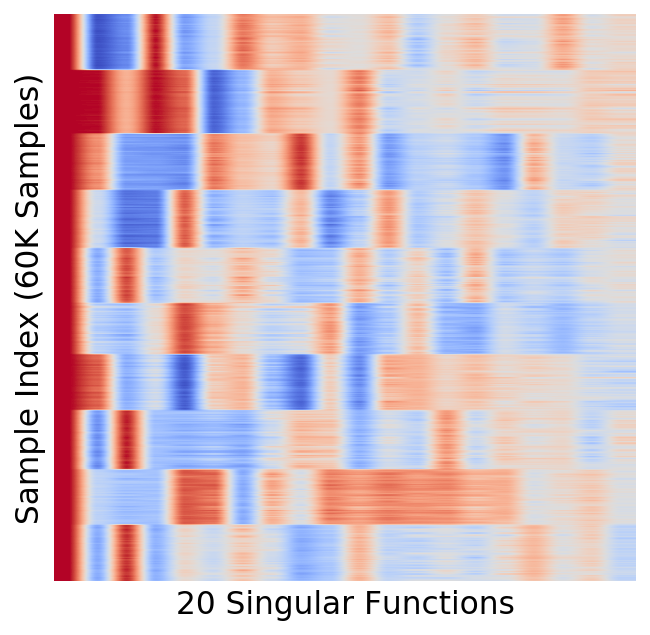}\vspace{-2pt}
\caption{\footnotesize Right singular functions $\{X, \widehat{\,X\,}\}$.}
\end{subfigure}
\caption{MNIST: left and right singular functions for the pairs $\{X,Y'\}$ and $\{X,\widehat{\,X\,}\}$. In (a) we have excluded the trivial constant singular function that always has a singular value $1$.}
\label{singular_functions_mnist}
\end{figure}

\noindent \textbf{Visualizing singular functions.} We next visualize and interpret the learned singular functions. 
Fig.~\ref{interpolated_samples999999999} shows results on the two-moons toy dataset, and Fig.~\ref{singular_functions_mnist} shows results on MNIST. Given a variable pair, the estimated density-ratio operator admits an SVD characterized by left singular functions 
$\widehat{f_1},\widehat{f_2},\ldots,\widehat{f_K}$ and right singular functions 
$\widehat{g_1},\widehat{g_2},\ldots,\widehat{g_K}$. 
We focus on two pairs: $\{X,Y'\}$ (data and noise-corrupted features) and $\{X,\widehat{\,X\,}\}$ (data and noise-free reconstructions). 
Here $Y'$ denotes the decoder input (noise-corrupted latent features), and $\widehat{\,X\,}$ denotes the decoder output (the noise-free reconstructions). 
As shown in many tables above, these two pairs will have the same dependence measure values, because $Y'$ and $\widehat{\,X\,}$ are interchangeable. We therefore compare whether their singular functions also match.\vspace{10pt}

\noindent \textbf{Singular functions: toy two-moon dataset.} Fig.~\ref{interpolated_samples999999999} visualizes the singular functions for both pairs. 
For $\{X,Y'\}$, the left singular functions $\widehat{f_k}(X)$ live on the 2D data domain, while the right singular functions $\widehat{g_k}(Y')$ live on the 1D feature domain. 
After training, we evaluate the learned networks on a dense grid to visualize these functions: a $50\times 50$ grid over $[0,1]\times[0,1]$ for the 2D domain and 100 uniformly spaced points over $[0,1]$ for the 1D domain. 
For $\{X,\widehat{\,X\,}\}$ we follow the same procedure, except that both left and right singular functions are defined on the 2D domain. 
When plotting 2D functions, we additionally multiply each heatmap by the empirical marginal density of the toy dataset to emphasize regions with non-negligible probability mass. \vspace{5pt}

\noindent These visualizations suggest three main observations: \vspace{5pt}

\noindent 1. The 2D left singular functions for $\{X,Y'\}$ closely match the 2D left/right singular functions for $\{X,\widehat{\,X\,}\}$. This supports the claim that, for dependence measurement, with a required appropriate reference variable, $Y'$ (decoder network input) and $\widehat{\,X\,}$ (decoder network output) are interchangeable. \vspace{5pt}

\noindent 2. For $\{X,\widehat{\,X\,}\}$, the left and right singular functions are also visually very similar. Such symmetry is not guaranteed for any SVD, and here likely reflects the symmetry induced by an end-to-end autoencoder mapping.\vspace{7pt}

\noindent 3. The 1D right singular functions for $\{X,Y'\}$ look like Hermite polynomials, consistent with decompositions of a Gaussian density functions. This may be explained either by the Gaussian additive noise used in the feature corruption, or the approximately Gaussian nature of noise in the toy two-moon dataset setup.\vspace{12pt}

\noindent From these results, we may give an interpretation that the decomposition is constructing an explicit alignment, or approximating an isomorphism between a sample-space Hilbert space and a feature-space Hilbert space. Each side admits its own orthonormal basis, and learning seeks to pack as many sample-Hilbert-space basis functions as possible using the basis functions in the feature Hilbert space. The SVD provides a canonical one-to-one matching between a basis for samples and a basis for low-dimensional features. The basis functions match one by one and are paired. Learning seeks to find as many such matched orthonormal pairs as possible.\vspace{-15pt}

\begin{figure}[t]
  \centering
\begin{subfigure}{.15\textwidth}\centering\includegraphics[width=.7\linewidth]{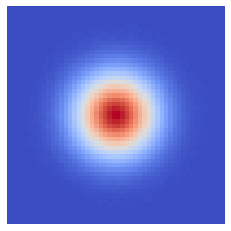}\vspace{-5pt}
\caption*{\footnotesize Trivial function}
\end{subfigure}\hspace{-10pt}
\begin{subfigure}{.15\textwidth}\centering\includegraphics[width=.7\linewidth]{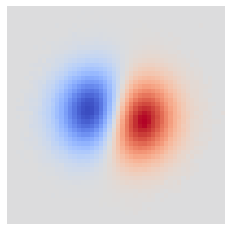}\vspace{-5pt}
\caption*{1st hermite}
\end{subfigure}\hspace{-10pt}
\begin{subfigure}{.15\textwidth}\centering\includegraphics[width=.7\linewidth]{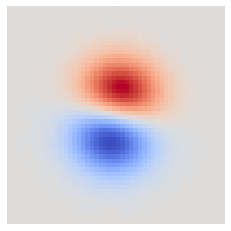}\vspace{-5pt}
\caption*{2nd hermite}
\end{subfigure}\hspace{-10pt}\\
\begin{subfigure}{.15\textwidth}\centering\includegraphics[width=.7\linewidth]{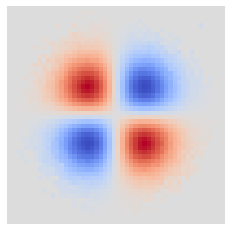}\vspace{-5pt}
\caption*{3rd hermite}
\end{subfigure}\hspace{-10pt}
\begin{subfigure}{.15\textwidth}\centering\includegraphics[width=.7\linewidth]{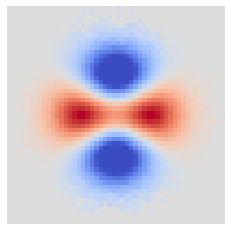}\vspace{-5pt}
\caption*{4th hermite}
\end{subfigure}\hspace{-10pt}
\begin{subfigure}{.15\textwidth}\centering\includegraphics[width=.7\linewidth]{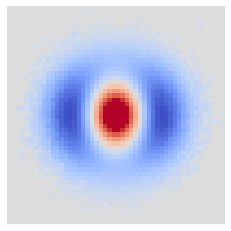}\vspace{-5pt}
\caption*{5th hermite}
\end{subfigure}\hspace{-10pt}\\
\begin{subfigure}{.15\textwidth}\centering\includegraphics[width=.7\linewidth]{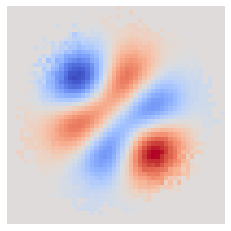}\vspace{-5pt}
\caption*{6th hermite}
\end{subfigure}\hspace{-10pt}
\begin{subfigure}{.15\textwidth}\centering\includegraphics[width=.7\linewidth]{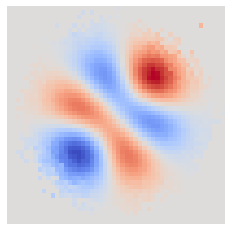}\vspace{-5pt}
\caption*{7th hermite}
\end{subfigure}\hspace{-10pt}
\begin{subfigure}{.15\textwidth}\centering\includegraphics[width=.7\linewidth]{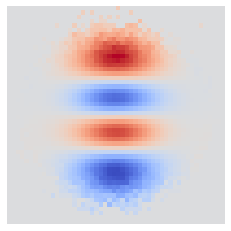}\vspace{-5pt}
\caption*{8th hermite}
\end{subfigure}\\ \vspace{-3pt}
\caption{Top 2D Hermite polynomials obtained by decomposing the standard Gaussian density. The first component coincides with the density itself and is therefore trivial. We include these polynomials to compare with the right singular functions in Fig.~\ref{singular_functions_mnist}a. Although they do not match exactly, the qualitative similarity is clear, consistent with the result in Fig.~\ref{interpolated_samples999999999}b for the toy dataset.}
\label{hermite}
\end{figure}

\noindent \textbf{Singular functions: MNIST.} For MNIST, directly visualizing singular functions $f_k(X)$ interpolated over the input domain is not feasible because the data is 784-dimensional. Instead, in Fig.~\ref{singular_functions_mnist} we evaluate and normalize the singular functions directly on the training samples, not on the interpolated grids. 

For $\{X,Y'\}$, the right singular functions are defined on the feature space. So in Fig.~\ref{singular_functions_mnist}a we first show a 2D projection of features, then visualize the top eight right singular functions by coloring each projected point by its function value. The heatmaps visualize the singular functions. 

The resulting patterns indicate that the singular functions partition the feature space: directions associated with larger singular values correspond to coarser, ``low-frequency'' partitions, while smaller singular values capture finer, ``high-frequency'' variations. Moreover, the leading singular functions clearly correlate with digit classes, indicating that they encode class-relevant structure.


We then compare (b) the left singular functions for $\{X,Y'\}$ with (c) the left and (d) the right singular functions for $\{X,\widehat{\,X\,}\}$.

Motivated by the toy results, we expect these three sets to be very similar, and this is also what we observe. To present this comparison, we construct heatmaps whose horizontal axis indexes the top 20 singular functions and whose vertical axis indexes the 60K MNIST training samples; each heatmap displays the function values across samples. 

Distinct block structures across rows align with digit classes, again confirming that the leading singular functions encode class information, and the three heatmaps are visually close, consistent with the exchangeability of $Y'$ and $\widehat{\,X\,}$ suggested in this paper.\vspace{7pt}

\noindent \textbf{Comparison with Hermite polynomials.} A closer look of the right singular functions of $\{X,Y'\}$ in Fig.~\ref{singular_functions_mnist}a suggests a close connection to Hermite polynomials. We plot the singular functions obtained by decomposing a standard Gaussian joint density, for which the singular functions are the two-dimensional Hermite functions. The leading polynomials up to order $9$ are shown in Fig.~\ref{hermite}. A mode-by-mode comparison of Fig.~\ref{hermite} and Fig.~\ref{singular_functions_mnist} reveals clear qualitative similarities. This implies that, even when the MNIST features projected to 2D appear visually unstructured, their dominant components may still follow Hermite-like patterns, potentially induced by the additive Gaussian noise and $v_p$. Moreover, as $v_p$ increases, the singular functions become more Hermite-like.

\subsection{Visualizing isomorphism}

A shown interesting result is that, for a static neural network, the input and output variables can be treated as interchangeable when we measure the statistical dependence against a proper third reference variable. Let us look at the topology of an autoencoder $X \;\rightarrow\; Y \;\rightarrow\; Y' \;\rightarrow\; \widehat{X}$, where the mapping $X \rightarrow Y$ is a deterministic encoder, $Y \rightarrow Y'$ is through additive Gaussian noise, and $Y' \rightarrow \widehat{X}$ is a deterministic decoder. Under this topology, $Y$ (features) can be viewed as a surrogate representation of $X$ (data), and $Y'$ (noisy features) as a surrogate of $\widehat{X}$ (noise-free reconstructions). This leads to an interesting result: if we view the density ratio as a metric function, the metric induced by the pair $\{Y, Y'\}$ on the space $\mathcal{Y}\times\mathcal{Y}$ should be equivalent to the metric induced by the pair $\{X,\widehat{X}\}$ on $\mathcal{X}\times\mathcal{X}$, and therefore they are isomorphic. 

\begin{figure}[H]
\centering
\begin{subfigure}{0.49\textwidth}
  \centering
  \includegraphics[width=0.7\linewidth]{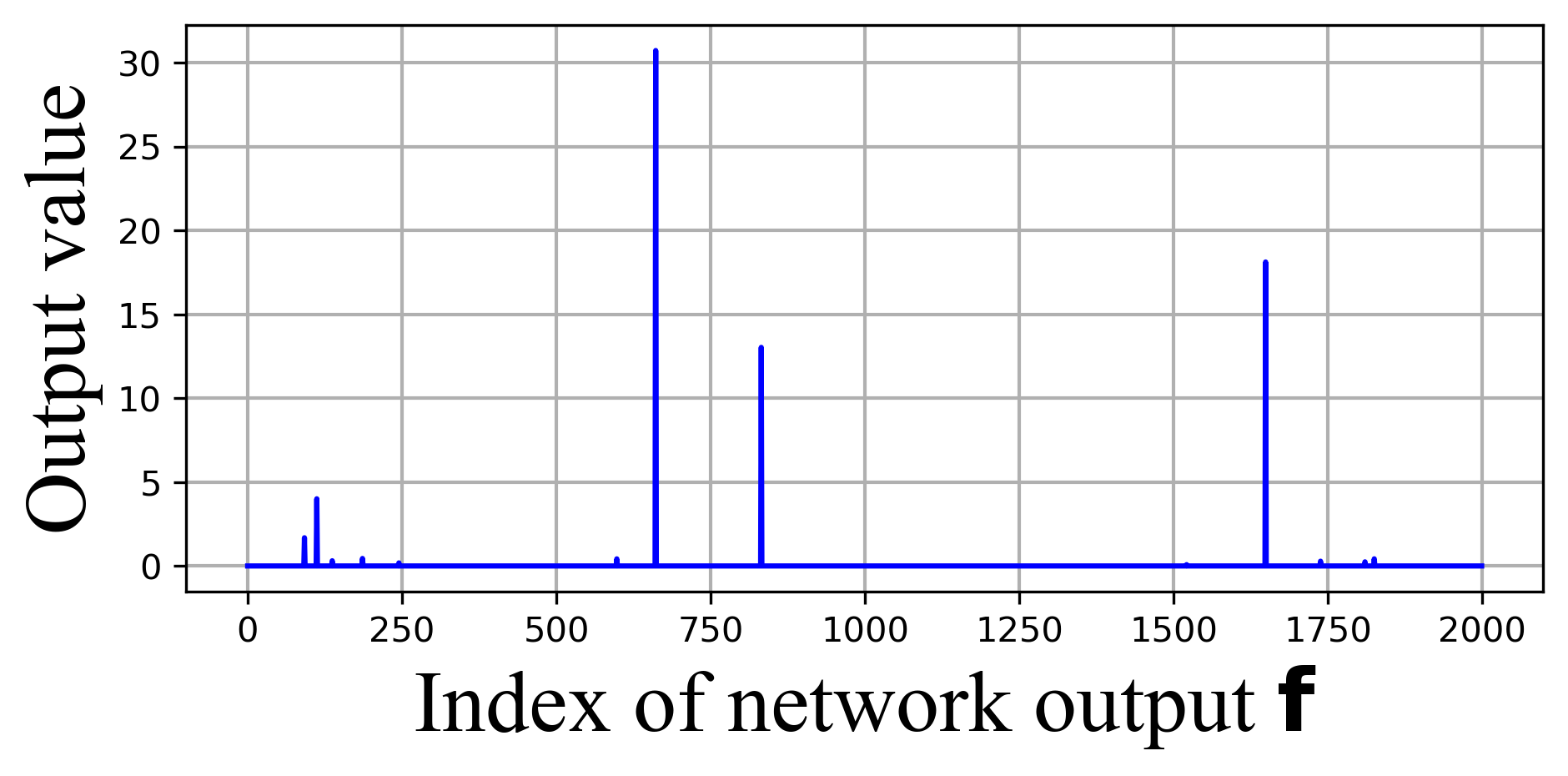}\vspace{-6pt}
  \caption{Example output of the dependence-estimator network $\mathbf{f}(X)$. The network has $K=2000$ nonnegative outputs (x-axis). The activation pattern for a single sample (one curve) is extremely sparse. This behavior is consistent across all samples.}
  \label{figure7a}
\end{subfigure}\vspace{-6pt}
\begin{subfigure}{0.4\textwidth}
  \centering
  \includegraphics[width=.85\linewidth]{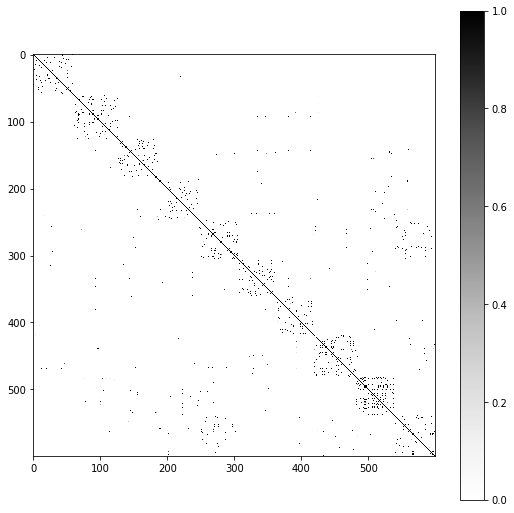}\vspace{-6pt}
  \caption{\footnotesize Metric $\frac{p(Y,Y')}{p(Y)p(Y')}$ on the space of $\mathcal{Y} \times \mathcal{Y}$ where $Y$ is 2D.}
  \label{figure7b}
\end{subfigure}
\begin{subfigure}{0.4\textwidth}
  \centering
  \includegraphics[width=.85\linewidth]{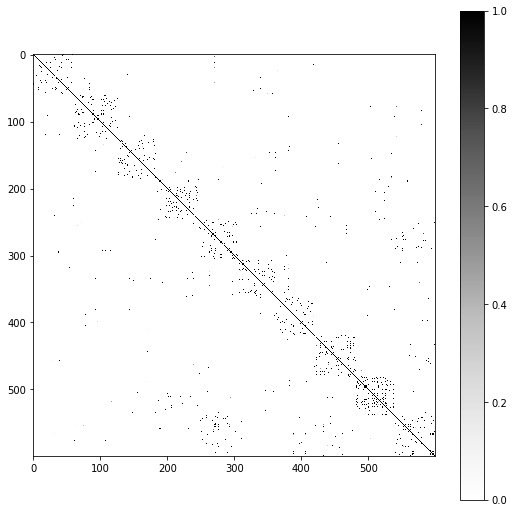}\vspace{-6pt}
  \caption{\footnotesize Metric $\frac{p(X,X')}{p(X)p(X')}$ on the space of $\mathcal{X} \times \mathcal{X}$ where $X$ is 784D.}  \label{figure7c}
\end{subfigure}
\caption{Visualizing the estimator output and comparing the induced metrics. Both the estimator activations and the resulting matrices are highly sparse, and the matrices induced by $\{Y,Y'\}$ and $\{X,\widehat{X}\}$ are visually similar.\vspace{-6pt}}
\end{figure}

Let us visualize the results for MNIST. In this experiment, the data are projected into 2D features. The feature variance is chosen to be $v_p = 10^{-4}$. We use the proposed statistical dependence estimator $\frac{p(X,Y)}{p(X)p(Y)} = \sum_{k=1}^K f_k(X) g_k(Y),$ where the estimator network outputs are nonnegative. First, if we visualize the outputs of the nonnegative, multi-dimensional networks $\mathbf{f}$ and $\mathbf{g}$ in the estimator network, we can see that they are extremely sparse, shown in Fig.~\ref{figure7a}. This sparsity may require further investigation.

Next we compare the density ratio function, the metric. Since MNIST contains $60000$ samples, the full $60000\times 60000$ matrix representing the metric distance is too large to visualize. We therefore subsample by taking every 100-th point, visualizing a $600\times 600$ matrix.Fig.~\ref{figure7b} shows the metric for $\mathcal{Y}\times\mathcal{Y}$, and Fig.~\ref{figure7c} the matrix for $\mathcal{X}\times\mathcal{X}$. The two matrices are visually very similar, supporting the proposed isomorphism.

Both matrices are also highly sparse, with most entries close to zero. Ideally, perfect reconstructions would induce a strongly diagonal structure. Both matrices are expected to be highly diagonal, since perfect reconstructions would induce a strongly diagonal correlation. But because of the insufficient dimensionality and the additive noise, while the matrices remain close to diagonal, some off-diagonal entries become non-negligible. These deviations likely reflect the compression of the model, but a more complete interpretation requires additional investigation.

\section{Formal Proofs}


This section provides the formal proofs underlying the cost functions introduced in the paper. 

Recall from Section~\ref{costs_ratio_sec} that we use the decomposition in Eq.~\eqref{orthonormal_decomposition}. For the proofs, it is convenient to keep track of two closely equivalent decompositions: 
\begin{equation}
\begin{aligned}
\frac{p(X,Y)}{\sqrt{p(X)}\,\sqrt{p(Y)}} &= \sum_{k=1}^K \sqrt{\lambda_k}\,\phi_k(X)\,\psi_k(Y),\\ 
\frac{p(X,Y)}{p(X)\,p(Y)} &= \sum_{k=1}^K \sqrt{\lambda_k}\,\widehat{\phi_k}(X)\,\widehat{\psi_k}(Y).
\end{aligned}
\label{orthonormal_decomposition2}
\end{equation}
The main text presents only the second form. The two expansions are equivalent: they differ only in the choice of base measure, i.e., the measure with respect to which orthonormality is defined. Specifically, \(\phi_k\) and \(\psi_k\) are orthonormal with respect to the Lebesgue measure \(\mu\), whereas \(\widehat{\phi}_k\) and \(\widehat{\psi}_k\) are orthonormal with respect to the probability measures induced by the marginals \(p(X)\) and \(p(Y)\). The two sets of functions differ only by a half-density reweighting:
\begin{equation}
\begin{aligned}
\phi_k(X) = \sqrt{p(X)}\cdot\widehat{\phi_k}(X), \;\; \psi_k(Y) = \sqrt{p(Y)}\cdot\widehat{\psi_k}(Y).
\end{aligned}
\end{equation}
Both forms are useful, and the distinction becomes particularly important when deriving a discrete equivalence of the decomposition in a Nyström-style fashion. The two decomposition will share the same set of singular values. If we look at the defined R\'enyi's mutual information
\begin{equation}
\resizebox{.8\linewidth}{!}{$
\begin{aligned}
\iint \frac{p^2(X,Y)}{p(X)p(Y)}\, dX\, dY
&= \iint \left( \frac{p(X,Y)}{\sqrt{p(X)}\sqrt{p(Y)}} \right)^2\, dX\, dY \\
&=  || \frac{p(X,Y)}{\sqrt{p(X)}\sqrt{p(Y)}} ||_2^2 = \sum_{k=1}^K\lambda_k,
\end{aligned}
$}
\label{equality_norm_term}
\end{equation}
it is the norm of the ratio function $\frac{p(X,Y)}{\sqrt{p(X)}\sqrt{p(Y)}}$, which turns out to be the sum of the square of singular values $\sum_{k=1}^K\lambda_k$. 

In the remainder of this section, we prove that optimizing the proposed NMF-like cost is an estimate of statistical dependence. We also provide proofs for our previous trace and logdet costs, the same as in~\cite{hu2024learning}.

\subsection{Proof for the new NMF-like cost}

The proof has three steps. First write down the inner product of the estimator with $p(X,Y)$. Next, factor out $\sqrt{p(X)}\sqrt{p(Y)}$. Finally, apply the Schwarz inequality.

Since the density ratio $\frac{p(X,Y)}{p(X)p(Y)}$ is nonnegative, we initialize two neural networks with multivariate, nonnegative outputs $f_1,f_2,\cdots, f_K$ and $g_1,g_2,\cdots, g_K$. We can directly construct a function $\widetilde{\rho}(X,Y) = \sum_{k=1}^K f_k(X) g_k(Y)$ without using any matrix inverses or products. Then, we compute the inner product of this function with the joint density $p(X,Y)$ and apply the Schwarz inequality:\vspace{-7pt}
\begin{equation}
\resizebox{.8\linewidth}{!}{
$\begin{aligned}
&\langle \sum_{k=1}^Kf_k(X) g_k(Y), p(X,Y)\rangle^2 \\  
&= \langle \sum_{k=1}^Kf_k(X) g_k(Y) \sqrt{p(X)}\sqrt{p(Y)}, \frac{p(X,Y)}{\sqrt{p(X)}\sqrt{p(Y)}} \rangle^2\\
& \leq \iint (\sum_{k=1}^Kf_k(X) g_k(Y) \sqrt{p(X)}\sqrt{p(Y)})^2 dXdY\\
& \hspace{140pt} \cdot\iint \frac{p^2(X,Y)}{p(X)p(Y)} dXdY
\end{aligned}$}\vspace{-5pt}
\label{equatonnnn}
\end{equation}
The first equality here results from factoring out $\sqrt{p(X)}\sqrt{p(Y)}$ from $p(X,Y)$ when computing the inner product. Then, the inequality in~\eqref{equatonnnn} follows from the Schwarz inequality: the square of the inner product is bounded by the product of the norms. One of the norm terms, $\iint \frac{p^2(X,Y)}{p(X)p(Y)} dXdY$, is exactly the quadratic form of mutual information, i.e., Shannon's mutual information without the $\log$. Rearranging these terms, we can construct a variational bound:
\begin{equation}
\resizebox{1\linewidth}{!}{
$\begin{aligned}
\frac{\langle \sum_{k=1}^Kf_k(X) g_k(Y), p(X,Y)\rangle^2}{ \iint (\sum_{k=1}^Kf_k(X) g_k(Y) \sqrt{p(X)}\sqrt{p(Y)})^2 dXdY} \leq \iint \frac{p^2(X,Y)}{p(X)p(Y)} dXdY. 
\end{aligned}$}
\end{equation}
Further, the terms on the right-hand side can also be written in the form of expectations, which yields:\vspace{-5pt}
\begin{equation}
\begin{aligned}
\frac{\left(\mathbb{E}\left[ \sum_{k=1}^K f_k(X)g_k(Y) \right]\right)^2} {\sum_{i, j = 1}^K \mathbb{E}\left[  f_i(X)f_j(X) \right] \cdot \mathbb{E}\left[  g_i(Y) g_j(Y) \right] } \\
\leq \iint \frac{p^2(X,Y)}{p(X)p(Y)} dXdY.
\end{aligned}
\label{bound6666666}
\end{equation}
If we compute the correlation matrices of the network outputs, with $\mathbf{R}_F = \mathbb{E}\left[ \mathbf{f}(X)\mathbf{f}^\intercal (X)\right]$ and $\mathbf{R}_G = \mathbb{E} \left[\mathbf{g}(Y) \mathbf{g}^\intercal (Y)\right]$, the denominator on the left-hand side, i.e., the variational cost in Eq.~\eqref{bound6666666}, can be written as $\sum_{i,j=1}^K (\mathbf{R}_F)_{i, j} (\mathbf{R}_G)_{i,j}$ or $\sum_{i,j=1}^K (\mathbf{R}_F \cdot \mathbf{R}_G)_{i,j}$. Then we obtain the final cost $c = \frac{\left(\mathbb{E}\left[ \sum_{k=1}^K f_k(X)g_k(Y) \right]\right)^2}{\sum_{i,j=1}^K (\mathbf{R}_F \cdot \mathbf{R}_G)_{i,j}}$, and maximizing this cost with respect to the neural networks $\mathbf{f}$ and $\mathbf{g}$ attains the upper bound, R\'enyi's mutual information of order $2$. This completes the proof. 

\subsection{Proof for the trace and logdet costs}

Next, we provide the formal proof that optimizing our previous trace and logdet costs will also give us the orthonormal decomposition of the density ratio. The proof is the same as in~\cite{hu2024learning} and our previous papers. 

Given estimator networks $\mathbf{f}$ and $\mathbf{g}$, suppose both of their autocorrelation matrices are full-rank. We first apply the normalization steps described in the eigenanalysis (Appendix~\ref{eigenanalysis}). 

In particular, we first whiten the two functions by $\overline{\mathbf{f}} = {\mathbf{R}}_F^{-\frac{1}{2}} \mathbf{f}$ and $\overline{\mathbf{g}} = {\mathbf{R}}_G^{-\frac{1}{2}} \mathbf{g}$, such that the autocorrelation matrices of $\overline{\mathbf{f}}$ and $\overline{\mathbf{g}}$ are both diagonal identity matrices. 

Next, we compute the SVD for $\overline{\mathbf{P}} \;=\; \mathbb{E}\!\left[\overline{\mathbf{f}}(X)\,\overline{\mathbf{g}}^\intercal(Y)\right] = \mathbf{Q}_F\mathbf{\Lambda}^{\frac{1}{2}} \mathbf{Q}_G^\intercal$, and then rotate the functions by $\widehat{\mathbf{f}} = \mathbf{Q}_F^\intercal \overline{\mathbf{f}}$ and $\widehat{\mathbf{g}} = \mathbf{Q}_G^\intercal \overline{\mathbf{g}}$.

Continuing from here, the proof follows as below. We use the trace cost as an example. \vspace{9pt}

\noindent \textbf{{Step 1: Prove the invariance of normalization.}} The costs for functions $\{{\mathbf{f}}, {\mathbf{g}}\}$ and normalized and rotated functions $\{\widehat{\mathbf{f}}, \widehat{\mathbf{g}}\}$ have the same values. First, the cost for the original functions $\{{\mathbf{f}}, {\mathbf{g}}\}$ can be written as
\begin{equation}
\begin{aligned}
Trace(\mathbf{R}_F^{-1} \,\mathbf{P}\, \mathbf{R}_G^{-1} \,\mathbf{P}\,^\intercal) & = Trace(\mathbf{R}_F^{-\frac{1}{2}} \,\mathbf{P}\, \mathbf{R}_G^{-1} \,\mathbf{P}\,^\intercal \mathbf{R}_G^{-\frac{1}{2}})\\
& =  Trace(\overline{\,\mathbf{P}\,} \,\overline{\,\mathbf{P}\,}^\intercal) \\
& = \sum_{k=1}^K \sigma_k^2 (\overline{\,\mathbf{P}\,}).
\end{aligned}
\end{equation}
Here \(\sigma_k(\cdot)\) denotes the \(k\)-th singular value of a matrix. The trace of the matrix \(\overline{\,\mathbf{P}\,}\,\overline{\,\mathbf{P}\,}^\intercal\) is the sum of its eigenvalues. Thus, it is also the sum of the squares of the singular values of $\overline{\,\mathbf{P}\,}$, by simple algebra. 


By the SVD $\overline{\,\mathbf{P}\,} = \mathbf{Q}_F\mathbf{\Lambda}^{\frac{1}{2}} \mathbf{Q}_G^\intercal$, we have $\mathbf{Q}_F^\intercal \overline{\,\mathbf{P}\,} \mathbf{Q}_G = \mathbf{\Lambda}^{\frac{1}{2}}$. Also note that $\mathbf{Q}_F^\intercal \overline{\,\mathbf{P}\,} \mathbf{Q}_G = \mathbf{\Lambda}^{\frac{1}{2}}$ is exactly the cross-correlation matrix between the rotated functions $\widehat{\mathbf{f}} = \mathbf{Q}_F^\intercal\, \overline{\mathbf{f}}$ and $\widehat{\mathbf{g}} = \mathbf{Q}_G^\intercal\, \overline{\mathbf{g}}$. To see this, since $\overline{\,\mathbf{P}\,} = \mathbb{E}\!\left[\overline{\mathbf{f}}(X)\,\overline{\mathbf{g}}^\intercal(Y)\right]$, multiplying $\overline{\,\mathbf{P}\,}$ by $\mathbf{Q}_F^\intercal$ and $\mathbf{Q}_G$ on the left and right, respectively, gives $\mathbf{Q}_F^\intercal \overline{\,\mathbf{P}\,} \mathbf{Q}_G
= \mathbb{E}\!\left[\mathbf{Q}_F^\intercal \overline{\mathbf{f}}(X)\,\overline{\mathbf{g}}^\intercal(Y)\mathbf{Q}_G\right]
= \mathbb{E}\!\left[\widehat{\mathbf{f}}\,\widehat{\mathbf{g}}^\intercal\right].$

Therefore, we have $\mathbb{E}[\widehat{\mathbf{f}}\,\widehat{\mathbf{g}}^\intercal] = \mathbf{\Lambda}^{\frac{1}{2}}$, i.e., the cross-correlation matrix for the rotated functions is the diagonal matrix $\mathbf{\Lambda}^{\frac{1}{2}}$.

Denote $\mathbb{E}[\widehat{\mathbf{f}}\, \widehat{\mathbf{g}}^\intercal] = \mathbf{\Lambda}^{\frac{1}{2}}  := \widehat{\,\mathbf{P}\,}$, then
\begin{equation}
\begin{aligned}
\sum_{k=1}^K \sigma_k^2 (\widehat{\,\mathbf{P}\,}) = \sum_{k=1}^K \sigma_k^2 (\overline{\,\mathbf{P}\,}).
\end{aligned}
\end{equation}
This means that the original functions $\{{\mathbf{f}}, {\mathbf{g}}\}$ and the rotated functions $\{\widehat{\mathbf{f}}, \widehat{\mathbf{g}}\}$ have the same cost values.\vspace{9pt}

\noindent\textbf{{Step 2: Rewrite the cost.}} Next, we rewrite the cost in the form of the new expectations:
\begin{equation}
\begin{aligned}
Trace(\mathbf{R}_F^{-1} \mathbf{P} \mathbf{R}_G^{-1} \mathbf{P}^\intercal) & = \sum_{k=1}^K \sigma_k^2 (\widehat{\,\mathbf{P}\,}) \\
& = \sum_{k=1}^K \mathbb{E}^2 \left[ (\widehat{\mathbf{f}}(\mathbf{X}))_k (\widehat{\mathbf{g}}(\mathbf{Y}))_k \right]. 
\end{aligned}
\label{eq_20}
\end{equation}
This is because $\widehat{\,\mathbf{P}\,}$ is the diagonal matrix $\mathbf{\Lambda}^{\frac{1}{2}}$. Thus, its $k$-th diagonal element is the inner product between the $k$-th elements of the functions $(\widehat{\mathbf{f}}(X))_k$ and $(\widehat{\mathbf{g}}(Y))_k$.\vspace{9pt}

\noindent \textbf{{Step 3: Apply the Schwarz inequality.}} Suppose the ratio has the decomposition $\frac{p(X, Y)}{p(X)p(Y)} = \sum_{k=1}^K \sqrt{\lambda_k} {\widehat{\phi_k}} (X) {\widehat{\psi_k}} (Y)$.

We simplify the notation using $(\widehat{\mathbf{f}}(X))_k:= (\widehat{\mathbf{f}})_k$, $(\widehat{\mathbf{g}}(Y))_k:= (\widehat{\mathbf{g}})_k$, ${\widehat{\phi_k}} (X):= \widehat{\phi_k}$, and ${\widehat{\psi_k}} (X):= \widehat{\psi_k}$.

The expectation (each term in the sum) in Eq.~\eqref{eq_20} can be written as
\begin{equation}
\resizebox{1\linewidth}{!}{
$\begin{aligned}
\mathbb{E}[(\widehat{\mathbf{f}})_k (\widehat{\mathbf{g}})_k] &= \int (\widehat{\mathbf{f}})_k (\widehat{\mathbf{g}})_k p(X,Y) dX dY\\
&= \int (\widehat{\mathbf{f}})_k (\widehat{\mathbf{g}})_k \sum_{q=1}^K \sqrt{\lambda_q} \phi_q \psi_q \;p(X)p(Y) dX dY \\
&= \sum_{q=1}^K \sqrt{\lambda_q} \cdot \int (\widehat{\mathbf{f}})_k \phi_q p(X) dX \cdot \int (\widehat{\mathbf{g}})_k \psi_q p(Y) dY \\
&= \sum_{q=1}^K \langle (\widehat{\mathbf{f}})_k,  \phi_q \rangle_{p(X)} \cdot \langle (\widehat{\mathbf{g}})_k,  \psi_q \rangle_{p(Y)}.
\end{aligned}$}
\end{equation}

\noindent We then apply the Schwarz inequality to each inner product term in the equation:
\begin{equation}
\begin{aligned}
\langle (\widehat{\mathbf{f}})_k,  \phi_q \rangle_{p(X)} \leq 1,\; \langle (\widehat{\mathbf{g}})_k,  \psi_q \rangle_{p(Y)} \leq 1.
\end{aligned}
\label{eq_22}
\end{equation}
Eq.~\eqref{eq_22} holds because the norms of these functions are all~$1$, due to normalization.

To achieve the maximum value of each $\mathbb{E}\left [(\widehat{\mathbf{f}})_k (\widehat{\mathbf{g}})_k \right]$, each $(\widehat{\mathbf{f}})_k$ has to be one of the left singular functions $\phi$, and similarly each $(\widehat{\mathbf{g}})_k$ has to be one of the right singular functions $\psi$. By induction, each $(\widehat{\mathbf{f}})_k$ has to match $\widehat{\phi_k}$, and each $(\widehat{\mathbf{g}})_k$ has to match $\widehat{\psi_k}$. This completes the proof, as we have now shown that the normalized rotated functions match the singular functions of the density ratio.


\section{Feature Learning: Beyond Feature Analysis}
\label{sec:feature_learning_beyond_analysis}

In the previous sections, we used the statistical-dependence estimator primarily as an analysis tool: after training an autoencoder, we measured the statistical dependence between different variable pairs. The results consistently show that statistical dependence increases during training, most notably for the pair $\{X,Y'\}$, which directly reflects the dependence between the input data and the learned features. This raises a natural question: Can we learn features by maximizing statistical dependence alone, without training a decoder?

We find that this is possible, but only with additive noises for the data: we must introduce additive Gaussian noise not only to the features but also to the data. Under this double-Gaussian assumption, maximizing the statistical dependence between noise-corrupted inputs and noise-corrupted features yields meaningful features, using an encoder alone, without a decoder. We describe the framework below.

\subsection{Noisy data and noisy features}
Let $X$ denote data samples with density $p(X)$. We introduce an auxiliary variable $X'$ by assuming a conditional Gaussian density:
\begin{equation}
p(X'|X) = \mathcal{N}(X'-X; v_X),
\end{equation}
We still define a Gaussian conditional density for features $Y$ given $X$ using an encoder $\mathbf{E}$:
\begin{equation}
p(Y|X) = \mathcal{N}(Y-\mathbf{E}(X); v_p).
\end{equation}
No decoder is used. Given $p(X'|X)$ and $p(Y|X)$, the induced joint density between $\{X',Y\}$ is
\begin{equation}
p(X',Y) = \int p(X'|X)\,p(X)\,p(Y|X)\,dX,
\end{equation}
i.e., we marginalize out $X$ to obtain a joint density between noise-corrupted inputs and learned features.

From $p(X',Y)$ we define the density ratio and the statistical dependence measure. We aim to maximize the overall statistical dependence:
\begin{equation}
\max_{p(Y|X)}\;\;
c \;:=\; \iint \frac{p^2(X',Y)}{p(X')\,p(Y)}\,dX'\,dY.
\label{eq:dependence_objective}
\end{equation}
Empirically, both variances matter: $v_X$ (data corruption) and $v_p$ (feature noise) must be chosen appropriately to obtain generalized features.

We observe that larger $v_X$ typically improves generalization of the learned representation. On MNIST, we choose $v_X > 1$, and values as large as $10^3$ produce more generalized features. The feature variance $v_p$ should be small; we find $v_p = 5\times 10^{-5}$ to be effective.

\subsection{Interpretation via the substitution pattern}

Throughout the paper, we discussed a substitution pattern: for a deterministic network, its input and output are (approximately) interchangeable when measuring statistical dependence against a reference variable $X'$. Here, the joint density underlying $\{X',X\}$ is fixed by $p(X'|X)$, so the dependence induced by $p(X',X)$ is fixed as well. When $v_p$ is small and we “substitute” $X$ by $Y=\mathbf{E}(X)$, the induced joint $p(X',Y)$ can approach the same dependence level as $p(X',X)$. From this perspective, training the encoder amounts to finding features whose induced dependence with $X'$ matches the fixed dependence between $X'$ and $X$.

\subsection{Two practical approaches}

We find two practical ways to optimize this dependence objective. The first uses a kernel density estimator (KDE) and therefore requires no additional estimator networks $\mathbf{f}$ and $\mathbf{g}$. It only defines a loss in terms of the encoder input \(X\) and the learned features \(Y\) (no decoder is needed). The second introduces two neural estimator networks and optimizes the NMF-like cost proposed in this paper by joint gradient ascent over the two estimators and the encoder (three networks together). Both approaches perform well.\vspace{9pt}


\noindent\textbf{KDE cost approach.} Given a minibatch $\{X_n\}_{n=1}^N$, a deterministic encoder produces $Y_n=\mathbf{E}(X_n)$. We construct a biased kernel density estimator (KDE) for the joint and marginals:
\begin{equation}
\resizebox{1\linewidth}{!}{
$\begin{aligned}
p(X',Y) &\approx \frac{1}{N}\sum_{n=1}^N \mathcal{N}(X'-X_n;v_X)\,\mathcal{N}(Y-Y_n;v_p), \\
p(X')p(Y) &\approx \left(\frac{1}{N}\sum_{n=1}^N \mathcal{N}(X'-X_n;v_X)\right)
            \left(\frac{1}{N}\sum_{n=1}^N \mathcal{N}(Y-Y_n;v_p)\right).
\end{aligned}$}
\label{equation_density_kde}
\end{equation}

In addition to estimating the joint density $p(X',Y)$, we also construct noisy samples by adding Gaussian noise:
${X_n}'= X_n + \sqrt{v_X}\cdot noise$ and ${Y_n}' = Y_n + \sqrt{v_p}\cdot noise$.
Using the KDE estimates together with these noise-corrupted samples, the cost function can be written as
\begin{equation}
\resizebox{1\linewidth}{!}{
$\begin{aligned}
c &= \iint \frac{p^2(X',Y)}{p(X')p(Y)} \, dX' dY \\
& = \iint \frac{p(X',Y)}{p(X')p(Y)} \cdot p(X',Y) \, dX' dY \\
&  \approx \iint \frac{ \frac{1}{N}\sum_{n=1}^N \mathcal{N}(X-X_n)\mathcal{N}(Y-Y_n)}{ \frac{1}{N^2}\sum_{n=1}^N \mathcal{N}(X-X_n) \sum_{n=1}^N \mathcal{N}(X-X_n) } \cdot p(X',Y) \, dX' dY \\
& \approx \frac{1}{N}\sum_{m=1}^N \frac{\frac{1}{N}\sum_{n=1}^N \mathcal{N}({X_m}'-X_n)\mathcal{N}({Y_m}'-Y_n)}{\frac{1}{N^2}\sum_{n=1}^N \mathcal{N}({X_m}'-X_n) \sum_{n=1}^N \mathcal{N}({Y_m}'-Y_n) }.
\end{aligned}$}
\label{c1}
\end{equation}

The derivation is as follows. First, we factor one $p(X',Y)$ out of the squared term inside the double integral. The remaining density ratio, $\frac{p(X',Y)}{p(X')p(Y)}$, can be directly estimated from Eq.~\eqref{equation_density_kde}. However, we still need to take the expectation with respect to the factored-out $p(X',Y)$. To do so, we use the noisy samples $X_n',Y_n'$ constructed above: sampling from the joint density $p(X',Y)$ yields exactly these noise-perturbed samples. Therefore, we can approximate the expectation (i.e., the double integral) by the sample average, leading to Eq.~\eqref{c1}.

One additional step is needed in practice. When the data are high-dimensional (e.g., images), a relatively large noise level may be required, in which case the Gaussian difference term $\mathcal{N}({X_m}'-X_n)$ can easily vanish. We found that substituting the Gaussian noisy samples $X_n',Y_n'$ with the original noise-free samples $X_n,Y_n$ resolves this issue. Although this substitution produces a biased estimator, we found empirically that using the noisy samples does not work, whereas replacing them by the noise-free samples in the cost does. With this modification, the final cost becomes
\begin{equation}
\resizebox{1\linewidth}{!}{
$\begin{aligned}
c = \frac{1}{N}\sum_{m=1}^N \frac{\frac{1}{N}\sum_{n=1}^N \mathcal{N}({X_m}-X_n;v_X)\mathcal{N}({Y_m}-Y_n;v_p)}{\frac{1}{N^2}\sum_{n=1}^N \mathcal{N}({X_m}-X_n;v_X) \sum_{n=1}^N \mathcal{N}({Y_m}-Y_n;v_p) }.
\end{aligned}$}
\label{c2}
\end{equation}

\vspace{9pt}

\noindent\textbf{Neural NMF (neural decomposer) approach.} Alternatively, we estimate the density ratio with the proposed neural decomposer (Algorithm~\ref{alg1}). A deterministic encoder produces $Y=\mathbf{E}(X)$. We then feed the noise-corrupted $X_n'$ and $Y_n'$ into two $K$-output networks $\mathbf{f}$ and $\mathbf{g}$ (ReLU activation), with $K=300$ for MNIST. Defining the correlation matrices $\mathbf{R}_F$ and $\mathbf{R}_G$, we then maximize the cost function $c = \frac{\left(\mathbb{E}\left[ \sum_{k=1}^K f_k(X)g_k(Y) \right]\right)^2}{\sum_{i,j=1}^K (\mathbf{R}_F \odot \mathbf{R}_G)_{i,j}}$ by gradient ascent jointly over the encoder net $\mathbf{E}$ and the two estimator nets $\mathbf{f}$ and $\mathbf{g}$.\vspace{9pt}

\subsection{Empirical comparison and discussion}
\label{section_emperical_estimation}
\begin{figure}[H]
\begin{subfigure}{0.235\textwidth}
  \centering
  \includegraphics[width=\linewidth]{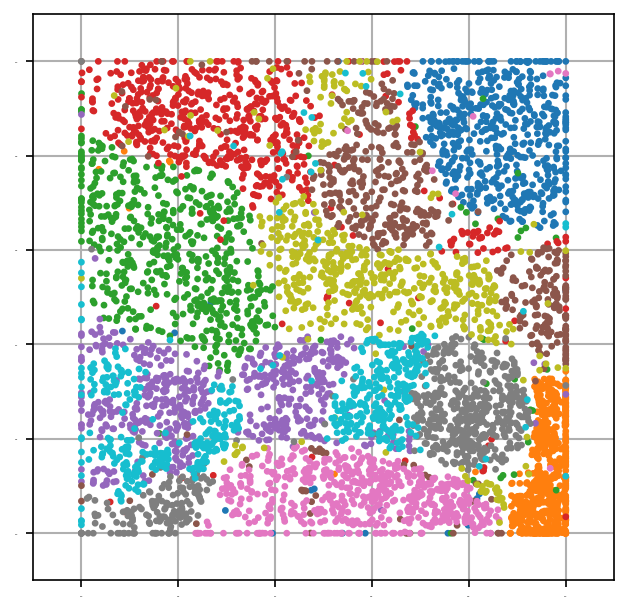}\vspace{-6pt}
  \caption{KDE cost}
\end{subfigure}
\begin{subfigure}{0.235\textwidth}
  \centering
  \includegraphics[width=\linewidth]{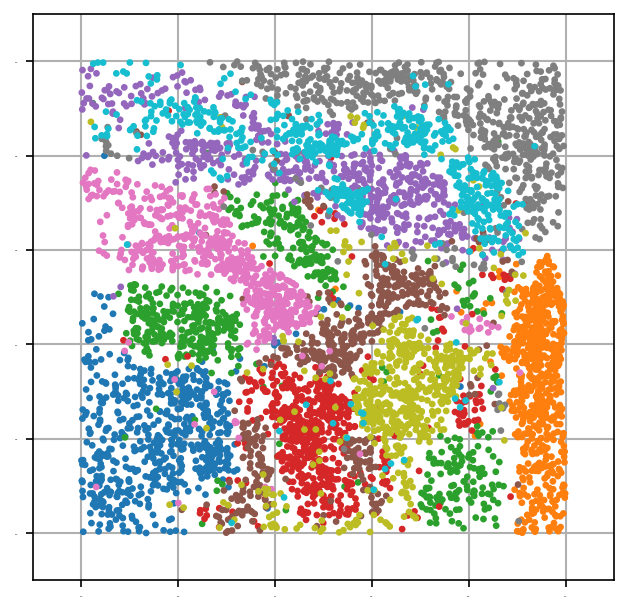}\vspace{-6pt}
  \caption{Neural NMF-like cost}
  \label{figure8bbbb}
\end{subfigure}
\caption{MNIST feature projections learned by maximizing statistical dependence directly as an objective without a decoder. In this example the Gaussian assumption on the sample data is needed.}
\label{figure_4kkkkkkkkk}
\end{figure}

Fig.~\ref{figure_4kkkkkkkkk} compares MNIST features learned using the KDE cost and the neural decomposer cost. 

\noindent Both approaches produce features that generalize and visually resemble those learned by an end-to-end autoencoder, suggesting that maximizing statistical dependence (with the proposed noise assumption) can recover autoencoder-like features without a decoder.

We also observe the following trade-offs:\vspace{3pt}

\noindent \textbf{KDE cost approach:} computational cost is dominated by kernel evaluations; in practice it is less demanding and often produces better results. It does not require explicit additive noise during training when using the biased objective~\eqref{c1}, but this bias appears necessary to avoid a vanishing denominator in high dimensions.\vspace{5pt}

\noindent \textbf{Neural decomposer approach:} unbiased in principle but requires explicit noise, which increases estimator variance and limits how large $v_X$ can be chosen.\vspace{5pt}

We also note that, for the KDE cost case, a sigmoid activation in the encoder network is required to regulate the feature range, and several points saturate at the boundary of the feature space $[0,1]\times [0,1]$. Removing the final sigmoid activation does not work for the KDE cost case, for reasons we do not yet fully understand. By contrast, the neural decomposer does not require this additional sigmoid constraint, and the learned features remain generalized when this sigmoid is removed.\vspace{9pt}

\begin{figure}[H]
\centering
\begin{subfigure}{0.3\textwidth}
  \centering
  \includegraphics[width=1\linewidth]{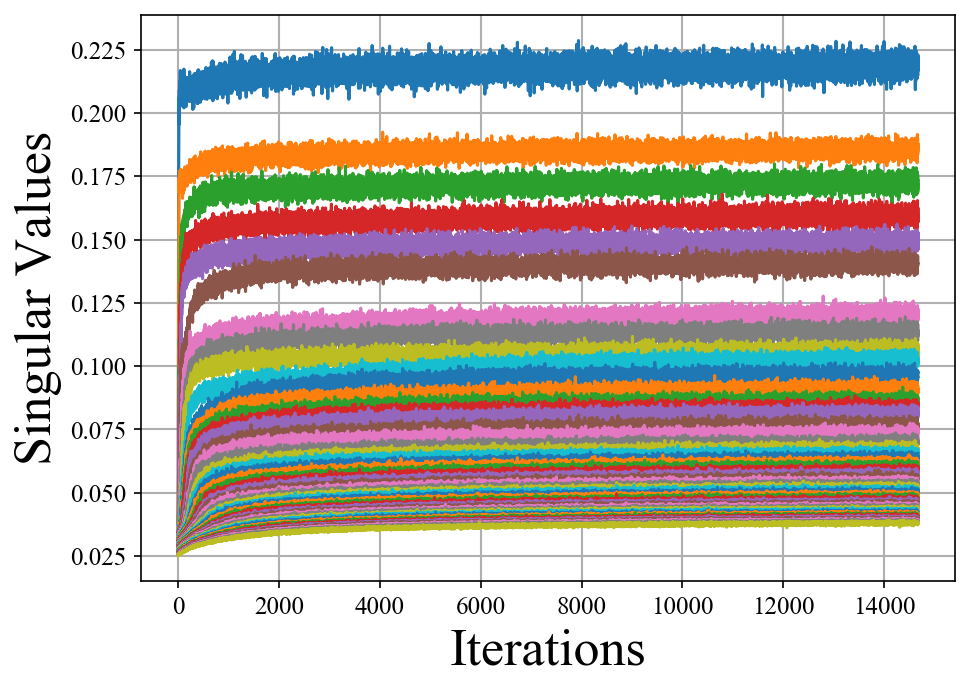}
  \caption*{ }
\end{subfigure}\\ \vspace{-19pt}
{(a) Learning curves of the top 50 singular values.}\\ \vspace{10pt}
\begin{subfigure}{0.125\textwidth}
  \centering
  \includegraphics[width=1\linewidth]{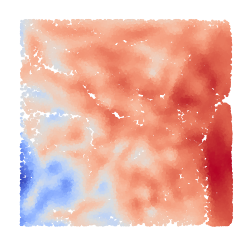}
  \caption*{ }
\end{subfigure}\hspace{-8pt}
\begin{subfigure}{0.125\textwidth}
  \centering
  \includegraphics[width=1\linewidth]{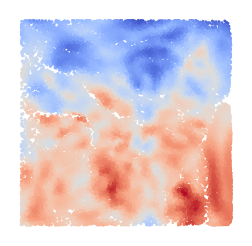}
  \caption*{ }
\end{subfigure}\hspace{-8pt}
\begin{subfigure}{0.125\textwidth}
  \centering
  \includegraphics[width=1\linewidth]{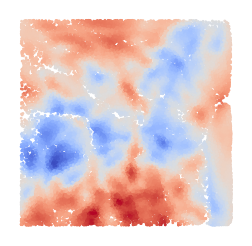}
  \caption*{ }
\end{subfigure}\hspace{-8pt}
\begin{subfigure}{0.125\textwidth}
  \centering
  \includegraphics[width=1\linewidth]{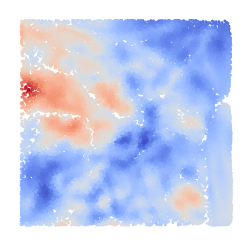}
  \caption*{ }
\end{subfigure}\vspace{-15pt}
\begin{subfigure}{0.125\textwidth}
  \centering
  \includegraphics[width=1\linewidth]{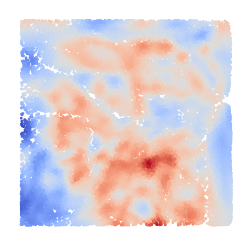}
  \caption*{ }
\end{subfigure}\hspace{-8pt}
\begin{subfigure}{0.125\textwidth}
  \centering
  \includegraphics[width=1\linewidth]{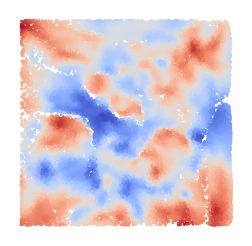}
  \caption*{ }
\end{subfigure}\hspace{-8pt}
\begin{subfigure}{0.125\textwidth}
  \centering
  \includegraphics[width=1\linewidth]{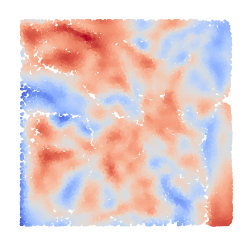}
  \caption*{ }
\end{subfigure}\hspace{-8pt}
\begin{subfigure}{0.125\textwidth}
  \centering
  \includegraphics[width=1\linewidth]{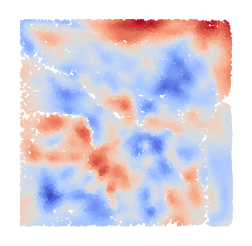}
  \caption*{ }
\end{subfigure}\vspace{-15pt}
{(b) Visualizing the top 8 singular functions.}\\ \vspace{0pt}
\caption{Visualization of the singular values and singular functions associated with Fig.~\ref{figure8bbbb}. A comparison can be made to Fig.~\ref{singular_functions_mnist}a.}
\label{singularvaluesfunctions}
\end{figure}

\noindent\textbf{Singular values and singular functions.} For the neural NMF-like case, visualizing the singular values at each iteration reveals a clear pattern: the singular values increase over time (Fig.~\ref{singularvaluesfunctions}). Thus, the objective is clearly, and underlyingly related to singular value decomposition, in the sense that singular values appear to be maximized sequentially. 

However, when we examine the singular functions, in the same way as in Fig.~\ref{singular_functions_mnist}a, the singular functions obtained by updating the encoder from gradient ascent and maximizing the statistical dependence are less interpretable than those in Fig.~\ref{singular_functions_mnist}a, where we apply the estimator to a trained encoder. 

This may be expected, since in the NMF-like training procedure we add substantial noise (with large variance) to the data samples, whereas the standard encoder measurement does not involve this noise addition. Nevertheless, this still requires further investigation.





\section{Extended Quantitative Evaluation}

Finally, an important goal of this paper is a quantitative analysis of autoencoder features. For readers interested in quantitative analysis, this section provides a more systematic approach and describes the source of our inspiration. 

We focus on the following questions. First, how can we validate that the estimated statistical dependence estimations are accurate and unbiased? Second, we assume a feature variance $v_p$. Does this Gaussian variance meaningfully exist in the learned features, and how can we validate that assumption?

To validate that our estimates are accurate and unbiased, and that the Gaussian assumption on the features is reasonable, we investigate a discrete equivalence of the autoencoder. Specifically, for the simple 2D toy dataset, we construct discrete matrices as the encoder and decoder conditional densities in a Nyström-style fashion, allowing us to reproduce the exact optimal solutions and dependence values of an autoencoder, without using neural networks. 

We encourage readers to experiment with alternative approaches, but the only approach we have found that reliably reproduces the autoencoder’s solution without using neural networks is to assume that the encoder and decoder Markov transition matrices are parameterized with Gaussianity.

\subsection{Inspirations from training an autoencoder}

We begin by training a standard autoencoder on the two-moon dataset. The encoder maps data to a 1D features and the decoder maps it back for reconstruction. Training an encoder-decoder neural network gives us the following observations.\vspace{7pt}

\begin{figure}[t]
  \centering
\begin{subfigure}{.1\textwidth}\includegraphics[width=\linewidth]{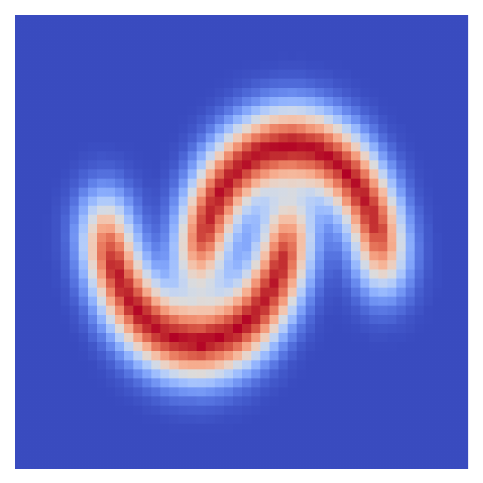}\vspace{-3pt}
\caption*{}
\end{subfigure}\vspace{8pt}
\begin{subfigure}{.11\textwidth}\includegraphics[width=\linewidth]{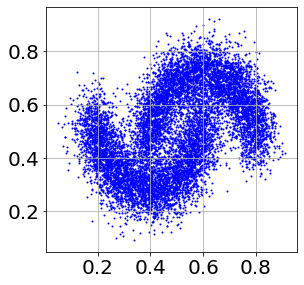}\vspace{-5pt}
\caption*{}
\end{subfigure}\vspace{-24pt}
\caption{The pdf and data samples of the two-moon distribution we use throughout the paper, with a variance of $0.04$.}
\label{interpolated_samples}
\end{figure}

\begin{figure}[t]
\centering
\begin{subfigure}{.1\textwidth}\includegraphics[width=\linewidth]{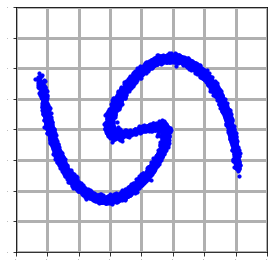}\vspace{-5pt}
\caption{{1K (Iter)}}
\label{figure22222a}
\end{subfigure}
\begin{subfigure}{.1\textwidth}\includegraphics[width=\linewidth]{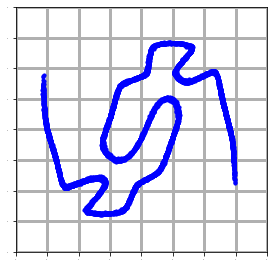}\vspace{-5pt}
\caption{{9K}}
\end{subfigure}
\begin{subfigure}{.1\textwidth}\includegraphics[width=\linewidth]{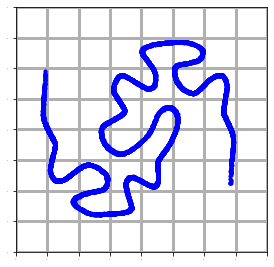}\vspace{-5pt}
\caption{{20K}}
\end{subfigure}
\begin{subfigure}{.1\textwidth}\includegraphics[width=\linewidth]{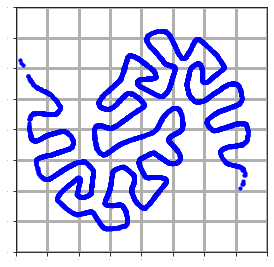}\vspace{-5pt}
\caption{{200K}}
\label{figure22222d}
\end{subfigure}
\begin{subfigure}{.1\textwidth}\includegraphics[width=\linewidth]{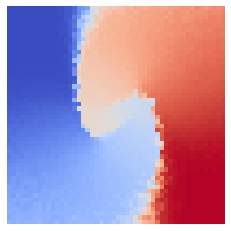}\vspace{-5pt}
\caption{1K}
\label{figure22222e}
\end{subfigure}
\begin{subfigure}{.1\textwidth}\includegraphics[width=\linewidth]{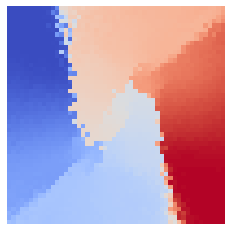}\vspace{-5pt}
\caption{9K}
\end{subfigure}
\begin{subfigure}{.1\textwidth}\includegraphics[width=\linewidth]{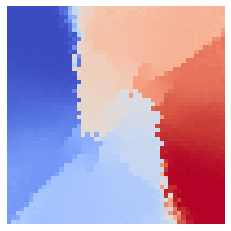}\vspace{-5pt}
\caption{20K}
\end{subfigure}
\begin{subfigure}{.1\textwidth}\includegraphics[width=\linewidth]{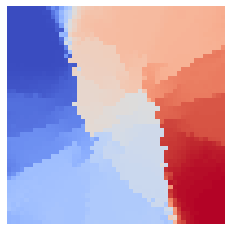}\vspace{-5pt}
\caption{200K}
\label{figure22222h}
\end{subfigure}\vspace{-4pt}
\caption{Regular autoencoders: reconstructions (Fig.~\ref{figure22222a}--Fig.~\ref{figure22222d}) and feature heatmaps (Fig.~\ref{figure22222e}--Fig.~\ref{figure22222h}) at iterations 1K, 9K, 20K, and 200K. The features are latently continuous, so the reconstructions appear as 1D curves in 2D, i.e., a manifold. Training also appears to have distinct stages: reconstructions and the implied decision boundary are initially coarse and become progressively more fine-grained.}
\label{figure22222result_autoencoder}
\end{figure}

\noindent \textbf{1. The solution is most likely unique.} The learned solution is consistent across architectural choices and hyperparameters. This motivates the eigen-expansion analysis. Such uniqueness may be explained by Mercer's theorem or by convexity.\vspace{7pt}

\noindent \textbf{2. The features are latently continuous.} Visualizing decoder reconstructions in 2D (Fig.~\ref{figure22222a}--Fig.~\ref{figure22222d}) produces 1D curves embedded in 2D, i.e., a manifold. This suggests continuity in the features $Y$.\vspace{7pt}

\noindent \textbf{3. Convergence appears stage-wise.} Over training (Fig.~\ref{figure22222a}--Fig.~\ref{figure22222d}), reconstructions evolve continuously from coarse to increasingly detailed approximations. Feature outputs (Fig.~\ref{figure22222e}--Fig.~\ref{figure22222h}), visualized as heatmaps, has a similar progression from coarse to fine structure.

Moreover, changes in reconstructed curves correspond closely to changes in the feature heatmaps, even in detailed regions. This further motivates the eigen-expansion perspective, as we want to relate stage-wise convergence to the sequential convergence of singular values.\vspace{-3pt}

\subsection{Nyström-style analysis: discrete equivalence for training an autoencoder}


We return to the autoencoder objective and its discrete form:
\begin{equation}
\resizebox{.8\linewidth}{!}{
$\begin{gathered}
\max_{\mathbf{P}_{Y|X}, \mathbf{Q}_{X|Y}} \;\; {Trace}\left({diag}(\mathbf{P}_X) \cdot \mathbf{P}_{Y|X} \cdot \log \mathbf{Q}_{X|Y}\right).\\
\max_{p(Y|X), p(X|Y)}\; \iint p(X) \cdot p(Y|X) \cdot \log q(X|Y) dX dY.\vspace{6pt}
\end{gathered}$}
\label{continuous_case}
\end{equation}
In the discrete case (Eq.~\eqref{continuous_case}), we start from the marginal density vector $\mathbf{P}_X$, apply an encoder Markov matrix $\mathbf{P}_{Y|X}$ to map probability mass into a lower-dimensional feature space, and then apply $\log \mathbf{Q}_{X|Y}$, where $\mathbf{Q}_{X|Y}$ is a decoder Markov matrix mapping features back to the original space. The trace computes the overall objective, corresponding to mean-squared reconstruction error in the neural setting.

If we parameterize the encoder and decoder directly as Markov transition matrices, we can solve the optimization exactly and obtain full control over the optimal conditional densities. This provides a direct comparison between the Markov-matrix solution and the neural-network solution.

The steps are as follows.

We use the two-moon dataset (Fig.~\ref{interpolated_samples}). We estimate the pdf beforehand on a $50 \times 50$ histogram grid, constructing a $50 \times 50$ matrix (equivalently, a $2500 \times 1$ vector $\mathbf{P}_X$), representing the marginal pdf of the data. 

We discretize and parameterize $p(Y|X) = \mathcal{N}(Y-\textbf{E}(X);v_p)$ and $q(Y|X) = \mathcal{N}(X-\textbf{D}(Y);v_q)$ as matrices $\mathbf{P}_{Y|X}$ and $\mathbf{Q}_{X|Y}$, and optimize them. This procedure of parameterizing $\mathbf{P}_{Y|X}$ and $\mathbf{Q}_{X|Y}$ is in Algorithm~\ref{algorithm22222222}.

This construction removes the need for neural networks or other universal function approximators. The only nonlinearity is a sigmoid that constrains the optimizable vector, after which we perform gradient ascent.

\begin{algorithm}
\small
\caption{\small Parameterize $\mathbf{P}_{Y|X}$.}
\begin{algorithmic}[1]\vspace{3pt}
\State Pick the grid size of 2D $X$: $50 \times 50$ ($2500 \times 1$) for $[0, 1]\times [0,1]$;\vspace{3pt}
\State Pick the grid size of 1D $Y$: $500\times 1$ for $[-0.01, 1.05]$;\vspace{3pt}
\State Initialize a vector of size $2500 \times 1$ followed by a sigmoid function to regulate its range. This vector is optimizable, representing $\textbf{E}(X)$;\vspace{3pt}
\State Create a vector of size $500 \times 1$ filled with $500$ uniformly interpolated points between $[-0.01, 1.05]$, representing $Y$;\vspace{3pt}
\State Pick a variance $v_p$, compute the Gaussian differences between the two vectors created above, forming a matrix $\mathbf{P}_{Y|X}$ of size $2500\times 500$, representing $p(Y|X) = \mathcal{N}(Y-\textbf{E}(X))$;\vspace{3pt}
\State To parameterize $q(X|Y)$, we apply the same procedure. The only difference is that the optimizable vector is of size $500\times 1$ for $Y$ (1D) and the interpolation vector is of size $2500\times 1$ (2D) for $X$. \vspace{3pt}
\end{algorithmic}
\label{algorithm22222222}
\end{algorithm}

We then optimize the objective in Eq.~\eqref{continuous_case}, i.e., the trace the matrix ${diag}(\mathbf{P}_X) \cdot \mathbf{P}_{Y|X} \cdot \log \mathbf{Q}_{X|Y}$. By derivation, this trace still reduces to mean-squared error and is irrelevant to $v_q$ of the decoder. This allows us to isolate and study the impact of the encoder variance $v_p$.

\begin{figure}[t]
\centering
\begin{subfigure}[b]{0.2\textwidth}
  \centering
  \includegraphics[width=0.48\linewidth]{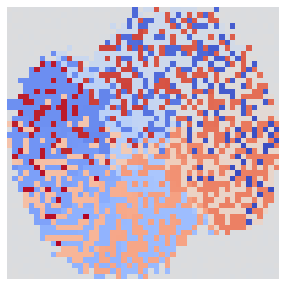}\hfill
  \includegraphics[width=0.48\linewidth]{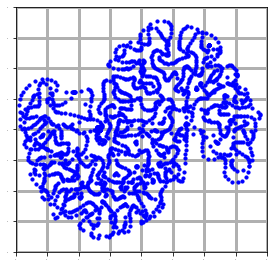}\vspace{-5pt}
  \caption{$v_p = 10^{-6}$}
\end{subfigure}\hspace{10pt}
\begin{subfigure}[b]{0.2\textwidth}
  \centering
  \includegraphics[width=0.48\linewidth]{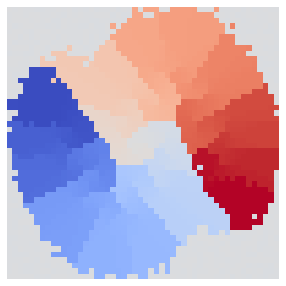}\hfill
  \includegraphics[width=0.48\linewidth]{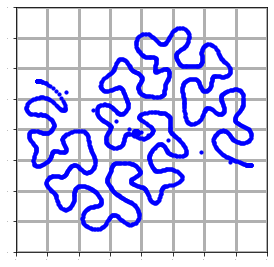}\vspace{-5pt}
  \caption{$v_p = 10^{-5}$}
\end{subfigure}

\begin{subfigure}[b]{0.2\textwidth}
  \centering
  \includegraphics[width=0.48\linewidth]{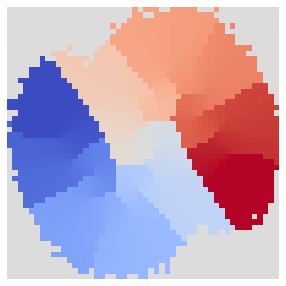}\hfill
  \includegraphics[width=0.48\linewidth]{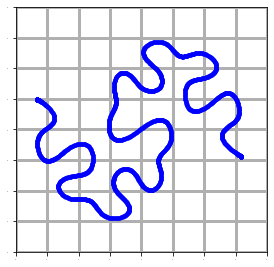}\vspace{-5pt}
  \caption{$v_p = 10^{-4}$}
\end{subfigure}\hspace{10pt}
\begin{subfigure}[b]{0.2\textwidth}
  \centering
  \includegraphics[width=0.48\linewidth]{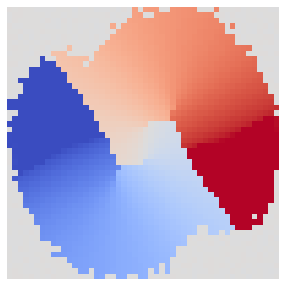}\hfill
  \includegraphics[width=0.48\linewidth]{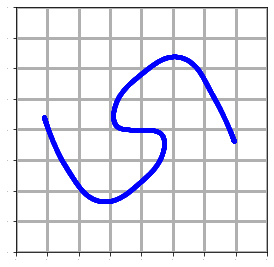}\vspace{-5pt}
  \caption{$v_p = 10^{-3}$}
\end{subfigure}
\caption{Producing the end-to-end autoencoder without neural networks, but with discretized Markovian matrices parameterized simply by two optimizable vectors. The results are consistent with autoencoder features. It also shows the importance of the encoder variance $v_p$ for features. If it is lower than a certain threshold, the model loses generalization.}
\label{figure3map1}
\end{figure}

With this setup, for simple toy datasets we can reproduce the stage-wide convergence behavior in Fig.~\ref{figure22222result_autoencoder} using only two optimizable vectors, rather than nonlinear neural networks. We vary $v_p$ and visualize the learned feature heatmaps and reconstructions in Fig.~\ref{figure3map1}. As $v_p$ decreases, reconstructions and heatmaps become progressively more fine-grained detailed.

For a grid size of $50\times 50$, the model loses generalization at a variance level of $10^{-6}$. Comparing with the regular autoencoder results (Fig.~\ref{figure22222result_autoencoder}), the effective $v_p$ in a standard autoencoder appears to fall between $10^{-5}$ and $10^{-4}$, approximately $5\times 10^{-5}$.

\begin{figure}[t]
\begin{subfigure}{0.35\textwidth}
  \centering
  \includegraphics[width=0.28\linewidth]{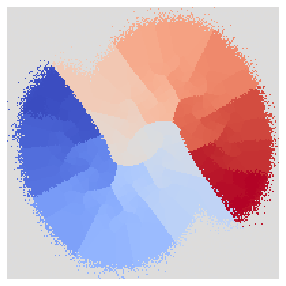}\hspace{3pt}
  \includegraphics[width=0.28\linewidth]{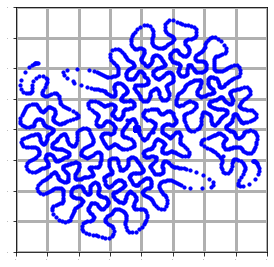}\vspace{-5pt}
  \caption{$v_p = 10^{-6}$, grid size = $200\times 200$}
\end{subfigure}\hspace{-5pt}%
\begin{subfigure}{0.1\textwidth}
  \centering
  \includegraphics[width=1\linewidth]{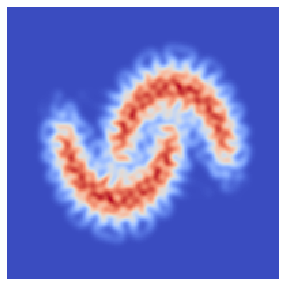}\vspace{-5pt}
  \caption{$q(X)$}
\end{subfigure}
\caption{The threshold of the lowest feature variance $v_p$ is tied to the grid size of the sample space, which can be lowered to $10^{-6}$ if we increase the grid size from $50\times 50$ (Fig.~\ref{figure3map1}) to $200\times 200$. The produced mean-squared error is around $2\times 10^{-4}$, which, combined with our finding that the error is the decoder variance $v_q$, gives us the marginal $q(X)$ in (b). }
\label{figure_4}
\end{figure}

The minimum viable $v_p$ is constrained by grid resolution and can be reduced to $10^{-6}$ by increasing the grid size from $50\times 50$ to $200\times 200$ (Fig.~\ref{figure_4}). This suggests that the smallest feasible $v_p$ is tied to the resolution of the sample space and the precision of the estimated pdf. We also find that the practical precision of a neural autoencoder is lower than what is shown in this discretized setting.\vspace{12pt}

\begin{figure}[t]
\centering
\begin{subfigure}{.235\textwidth}
\centering
\includegraphics[width=.9\linewidth]{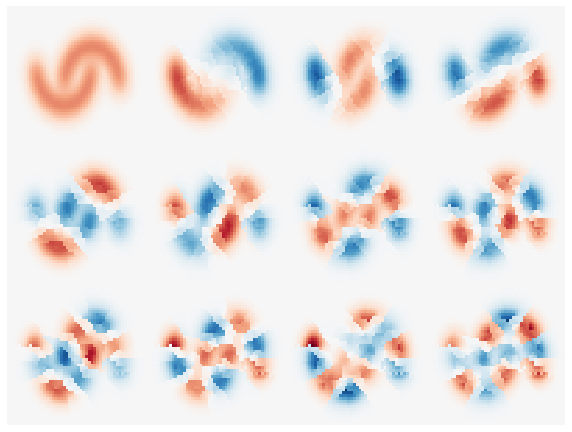}
\caption{\footnotesize $\rho_{1/2}(X, \widehat{\,X\,})$ left singular func.}
\end{subfigure}
\begin{subfigure}{.235\textwidth}
\centering
\includegraphics[width=.9\linewidth]{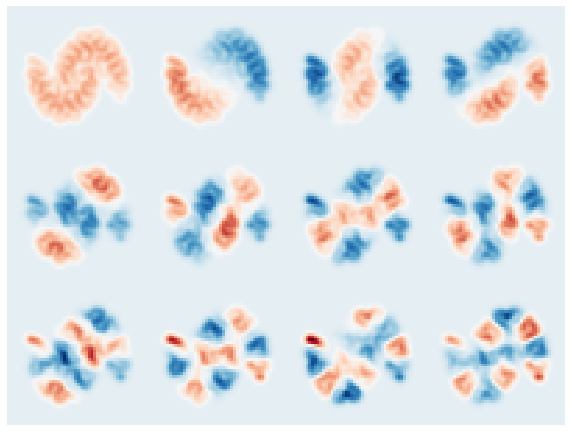}
\caption{\footnotesize $\rho_{1/2}(X, \widehat{\,X\,})$ right singular func.}
\end{subfigure}\vspace{5pt}
\begin{subfigure}{.235\textwidth}
\centering
\includegraphics[width=.9\linewidth]{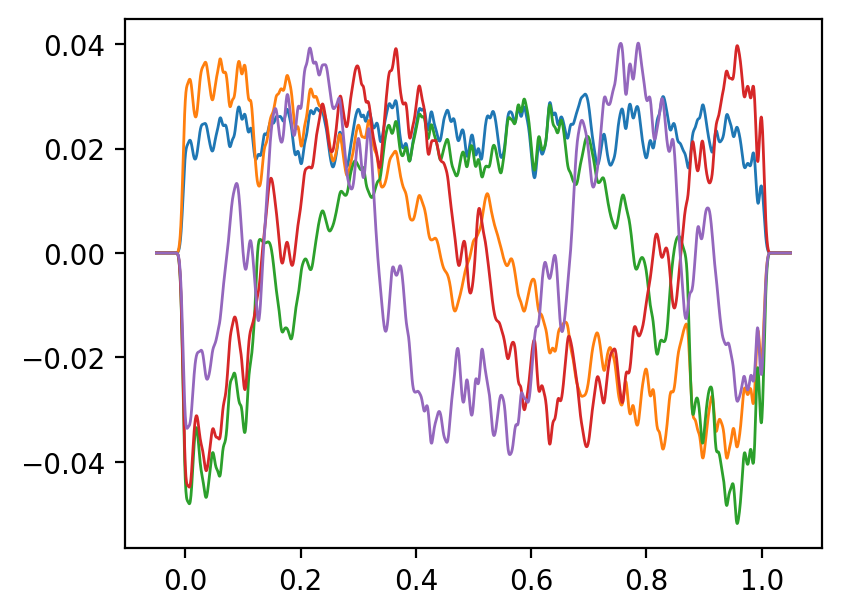}
\caption{\footnotesize $\rho_{1/2}(Y, \widehat{\,X\,})$ left singular func.}
\end{subfigure}
\begin{subfigure}{.235\textwidth}
\centering
\includegraphics[width=.9\linewidth]{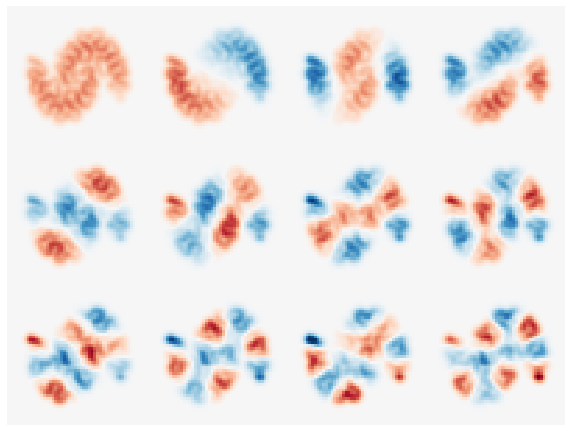}
\caption{\footnotesize $\rho_{1/2}(Y, \widehat{\,X\,})$ right singular func.}
\end{subfigure}\vspace{7pt}
\begin{subfigure}{.5\textwidth}
\centering
\includegraphics[width=.55\linewidth]{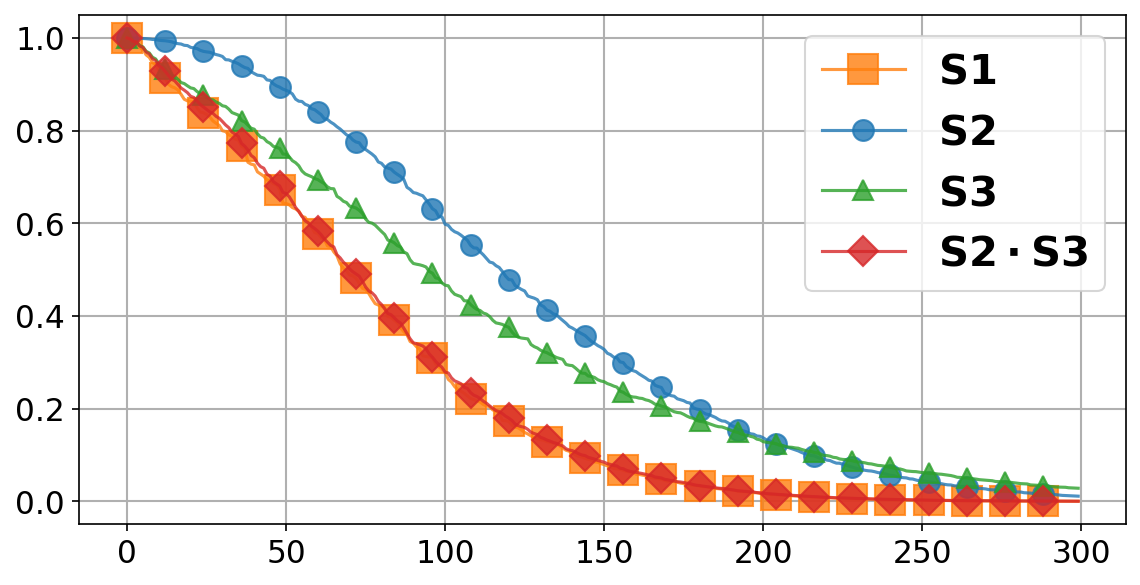}
\caption{Singular-value spectra: $\mathbf{S1}$ for $\rho_{1/2}(X,\widehat{\,X\,})$, $\mathbf{S2}$ for $\rho_{1/2}(X,Y)$, and $\mathbf{S3}$ for $\rho_{1/2}(Y,\widehat{\,X\,})$. $\mathbf{S2\odot S3}$ denotes the componentwise product between every singular value of $\rho_{1/2}(X,Y)$ and $\rho_{1/2}(Y,\widehat{\,X\,})$.}
\end{subfigure}
\caption{A Nyström-style analysis of the autoencoder.}
\label{result_autoencoder33333}
\end{figure}

\noindent \textbf{Eigenanalysis (Fig.~\ref{result_autoencoder33333}).} Now with discretized estimates of $p(X)$, $p(Y|X)$, and $q(X|Y)$ in hand, we have the total control of the probability densities in this autoencoder system. Recall that the statistical dependence can be measured by the SVD of the ratio function $\frac{p(X,Y)}{\sqrt{p(X)}\sqrt{p(Y)}}$, which we denote as $\rho_{1/2}(X,Y)$. Using the encoder-decoder conditionals, we can first form an autoencoder recurrence density function 
\begin{equation}
\begin{aligned}
p(X,\widehat{\,X\,}) = \int p(X)\cdot p(Y|X)\cdot q(\widehat{\,X\,}|Y) \,dY.
\end{aligned}
\label{equation_autoencoder}
\end{equation}
Here $\widehat{\,X\,}$ denotes the reconstruction produced by the decoder. This joint density is between the original data samples $X$ and the reconstructions $\widehat{\,X\,}$. 

From \eqref{equation_autoencoder} it follows that the associated ratio functions satisfy the corresponding composition (recurrence) relation
\begin{equation}
\rho_{1/2}(X,\widehat{\,X\,})
=\int \rho_{1/2}(X,Y)\,\rho_{1/2}(Y,\widehat{\,X\,})\,dY,
\end{equation}
where each factor is itself a ratio function. We then compute the SVD of each density ratio function $\rho_{1/2}$, and interpret its singular values and singular functions as dependence components. Empirically, shown in Fig.~\ref{result_autoencoder33333}, we found that
\begin{itemize}[leftmargin=*]
\item The singular functions of $\rho_{1/2}(X,Y)$ are not meaningful;
\item The right singular functions of $\rho_{1/2}(X,\widehat{\,X\,})$ coincide with the right singular functions of $\rho_{1/2}(Y,\widehat{\,X\,})$;
\item The left singular functions of $\rho_{1/2}(Y,\widehat{X})$, viewed as functions of the 1D feature variable, are very similar to Hermite polynomials, because of the Gaussian assumption.
\item The singular values follow a termwise (Hadamard) product rule: after matching components by index, the $k$th singular value of $\rho_{1/2}(X,X')$ equals the product of the $k$th singular values of $\rho_{1/2}(X,Y)$ and $\rho_{1/2}(Y,\widehat{X})$, for all $k$.
\end{itemize}

\subsection{Nyström-style analysis: filling the gap between two singular-value spectra}\label{twospectra}


In the previous section and Fig.~\ref{result_autoencoder33333}, we showed that the SVD of the density-ratio functions associated with the encoder and decoder can be highly informative.

Nevertheless, this autoencoder setup and the discretized analysis reveal a potential limitation. As shown in Fig.~\ref{result_autoencoder33333}(c), the singular values of the encoder ratio function $\rho_{1/2}(X,Y)$ (curve \textbf{S2}) are consistently larger than those of the decoder ratio function $\rho_{1/2}(Y,\widehat{\,X\,})$ (curve \textbf{S3}), leaving a clear gap between the two spectra. This is at odds with the expected symmetry of an ideal autoencoder: the decoder should be an inverse mapping of the encoder, which means that, at the optimum, the two singular-value spectra would coincide. Instead, Fig.~\ref{result_autoencoder33333}(c) shows a persistent spectral mismatch. Moreover, a direct ratio decomposition of the encoder joint $p(X,Y)$ (between data $X$ and features $Y$) has singular functions that are not meaningful.

This motivates a concrete objective: to reduce or eliminate the spectral gap between the encoder and decoder, so that the decoder becomes an exact inverse of the encoder in the spectral sense, restoring the symmetry implied by the autoencoder architecture.

We adopt two modifications.\vspace{9pt}

\noindent \textbf{Fix 1: increase decoder capacity via a mixture model.} If $q(X|Y)$ is not an inverse of the encoder, a plausible cause is that a single Gaussian decoder $q(X|Y)=\mathcal{N}\bigl(X-\mathbf{D}(Y);\, v_p\bigr)$ may be too restrictive to represent the inverse of the encoder.

A minimal extension is to replace the single Gaussian with a mixture:
\begin{equation}
\begin{gathered}
q(X|Y)
=\int \mathcal{N}\!\bigl(X-\mathbf{D}(Y,c);v_p(c)\bigr)\,p(c)\,dc,
\end{gathered}
\end{equation}
where $c$ is a latent component index (in practice, a discrete categorical variable is sufficient). The resulting training objective becomes
\begin{equation}
\resizebox{1\linewidth}{!}{$
\begin{gathered}
\iint p(X)\,p(Y|X)\,
\log \int \mathcal{N}\!\bigl(X-\mathbf{D}(Y,c);v_p(c)\bigr)\,p(c)\,dc\; dX\, dY .
\end{gathered}$}
\label{trick1}
\end{equation}
Unlike the single-Gaussian case, the loss is no longer equivalent to the MSE because the logarithm does not cancel the mixture integral; the resulting objective involves a log-sum-exp structure.

Still using two discretized optimizable Markov transition matrices, this modification corresponds to replacing the optimizable mean vector (size $2500\times 1$) with a matrix of size $2500\times C$ (one mean vector per mixture component). This mixture matrix instead of the previous mean vector will be optimizable. When calculating the cost, we take the $\log$ of this optimizable matrix and then marginalize by taking the average over $C$. In short, relative to the previous experiment, the only change is to replace the optimizable vector with an optimizable matrix (with $C$ columns) and to marginalize over the $C$ components by averaging before taking the $\log$.

We have to admit, however, that the mixture decoder is computationally more expensive and often harder to optimize than the standard autoencoder decoder, partly because the objective becomes more non-convex and less well-behaved.\vspace{9pt}

\begin{figure}[t]
\centering
\begin{subfigure}{.235\textwidth}
\centering
\includegraphics[width=.9\linewidth]{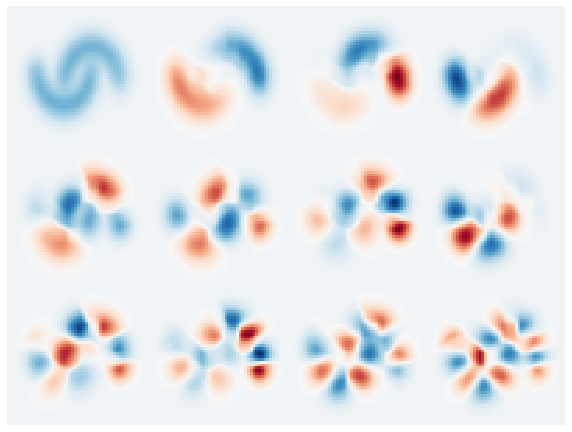}
\caption{\footnotesize \footnotesize $\rho_{1/2}(X', Y)$ left singular func.}
\end{subfigure}
\begin{subfigure}{.235\textwidth}
\centering
\includegraphics[width=.9\linewidth]{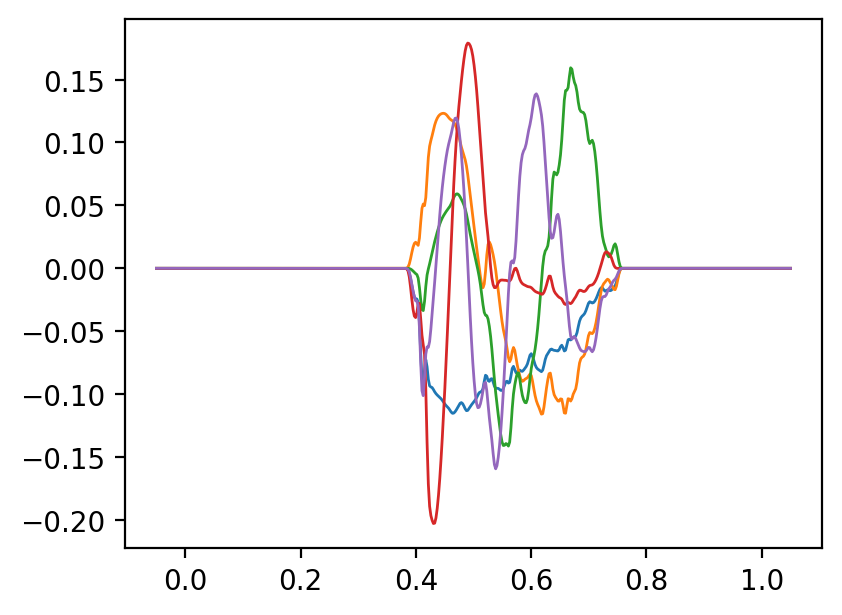}
\caption{\footnotesize $\rho_{1/2}(X', Y)$ right singular func.}
\end{subfigure}
\begin{subfigure}{.235\textwidth}
\centering
\includegraphics[width=.9\linewidth]{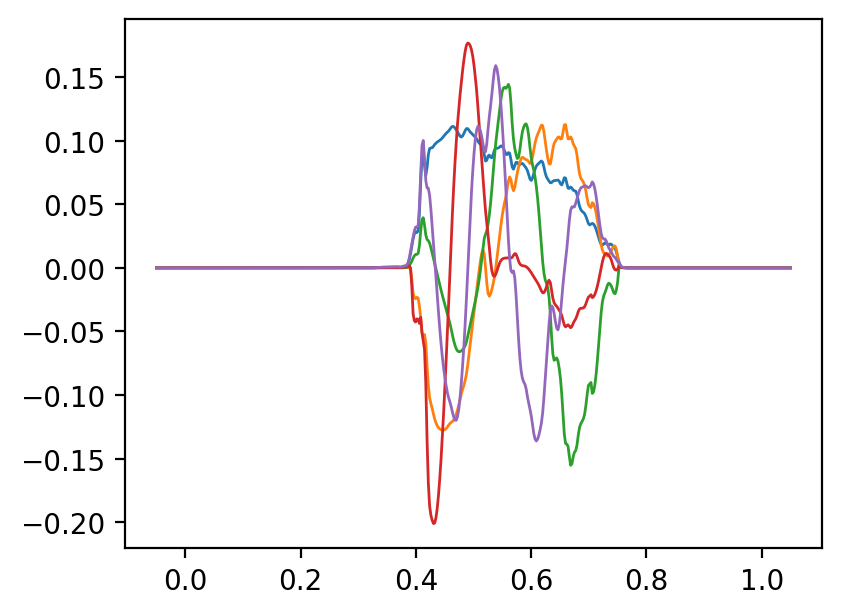}
\caption{\footnotesize $\rho_{1/2}(Y, \widehat{\,X\,})$ left singular func.}
\end{subfigure}
\begin{subfigure}{.235\textwidth}
\centering
\includegraphics[width=.9\linewidth]{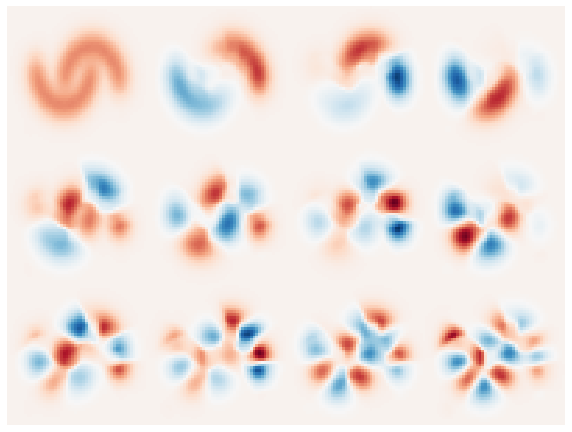}
\caption{\footnotesize $\rho_{1/2}(Y, \widehat{\,X\,})$ right singular func.}
\end{subfigure}\vspace{10pt}
\begin{subfigure}{.5\textwidth}
\centering
\includegraphics[width=.55\linewidth]{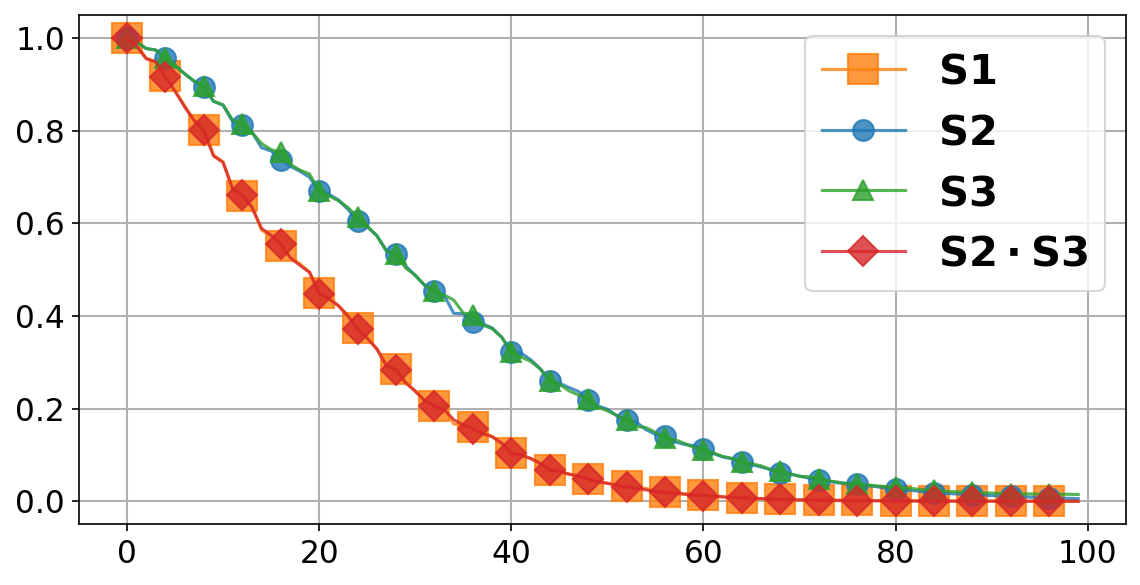}
\caption{Singular-value spectra: $\mathbf{S1}$ for $\rho_{1/2}(X,\widehat{\,X\,})$, $\mathbf{S2}$ for $\rho_{1/2}(X',Y)$, and $\mathbf{S3}$ for $\rho_{1/2}(Y,\widehat{\,X\,})$. $\mathbf{S2\odot S3}$ denotes the componentwise product between every singular value of $\rho_{1/2}(X',Y)$ and $\rho_{1/2}(Y,\widehat{\,X\,})$.}
\end{subfigure}
\caption{The Nyström-style analysis after applying the two modifications from Section~\ref{twospectra}. In Fig.~\ref{result_autoencoder33333}, the encoder spectrum (curve \textbf{C2}) consistently exhibits larger singular values than the decoder spectrum (curve \textbf{C3}), revealing a spectral mismatch between the two operators. After applying the fixes, the two spectra coincide, indicating a symmetry: the encoder and decoder become inverse mappings of each other (in the sense that they share the same singular values and have swapped left/right singular functions).}
\label{result_autoencoder}
\end{figure}

\noindent \textbf{Fix 2: regularize deterministic dependence by the input noise.} Another plausible cause is that the statistical dependence between the data $X$ and features $Y$ becomes overly dependent, such that directly measuring the dependence between them becomes ill-posed. To regularize the problem, we introduce a conditional Gaussian density on the input data: $p(X'|X) = \mathcal{N}(X'-X;\, v_X)$, and optimize the smoothed objective
\begin{equation}
\iiint p(X'|X)\cdot p(X) \cdot p(Y|X) \cdot \log q(X|Y)\,dX'\,dX\,dY.
\label{trick2}
\end{equation}
After training, we analyze the joint pdf $p(X',Y)$ and $p(X',\widehat{X})$, equivalently the SVD of the associated density-ratio functions $\rho_{1/2}(X',Y)$ and $\rho_{1/2}(X',\widehat{X})$.

For the discretization example, this smoothing corresponds to precomputing a kernel matrix $\mathbf{K}\in\mathbb{R}^{2500\times 2500}$, where $\mathbf{K}_{ij}$ is the Gaussian kernel evaluated between the $i$-th and $j$-th grid points in $X$-space. The objective can then be expressed as
\begin{equation}
Trace\Bigl(\mathbf{K}\,\operatorname{diag}(\mathbf{P}_X)\,\mathbf{P}_{Y|X}\,\log \mathbf{Q}_{X|Y}\Bigr),
\end{equation}
i.e., we left-multiply by $\mathbf{K}$ before taking the trace. Empirically, this minor modification greatly stabilizes training when the decoder is modeled by a mixture.\vspace{9pt}

\noindent \textbf{Result.} Combining Eq.~\eqref{trick1} (mixture decoder) and Eq.~\eqref{trick2} (input smoothing) closes the spectral gap: the decoder becomes (approximately) the inverse mapping of the encoder, and the learned representation can be interpreted as a maximal mutual-information solution.

Fig.~\ref{result_autoencoder} shows the eigenanalysis after applying both fixes. The singular values and singular functions now match across the encoder $p(X',Y)$ and the decoder $q(Y,\widehat{\,X\,})$, indicating an exact inverse relationship: the left singular functions of $\rho_{1/2}(X',Y)$ coincide with the right singular functions of $\rho_{1/2}(Y,\widehat{\,X\,})$, and vice versa, with identical singular values. We also observe that this setting can be more difficult to train and may yield fewer effective singular values than the vanilla autoencoder setting.

\subsection{Motivation for additive input noise in statistical-dependence feature learning} 

The previous section shows that, with a mixture decoder and input noises, feature learning can be framed as an exact mutual-information maximization problem. In practice, however, this procedure may provide limited benefits relative to its computational cost: parameterizing and training a mixture decoder can be expensive, and its optimization can be more delicate.

Since the (modified) decoder’s main role is to approximate an inverse mapping of the encoder, a natural question is whether we can remove the decoder altogether, keep only the encoder, and maximize statistical dependence directly.

We found that this becomes feasible only with the Gaussian input noise assumption.  Assume $p(X)$ is given and the encoder is $p(Y|X)=\mathcal{N}\!\bigl(Y-\mathbf{E}(X);v_p\bigr)$, with $v_p$ arbitrarily small. Introduce the same conditional Gaussian density for the data and define
\begin{equation}
\begin{gathered}
p(X',Y) = \int p(X'|X)\,p(X)\,p(Y|X)\,dX .
\end{gathered}
\label{trick3}
\end{equation}
We then directly maximize the mutual information of $p(X',Y)$:
\begin{equation}
\iint p(X',Y)\log \frac{p(X',Y)}{p(X')p(Y)}\, dX'\, dY,
\end{equation}
or, more conveniently for our ratio-based analysis,
\begin{equation}
\iint \frac{p^2(X',Y)}{p(X')p(Y)}\, dX'\, dY.
\end{equation}
From Eq.~\eqref{trick3}, the corresponding ratio function satisfies the composition rule
\begin{equation}
\rho_{1/2}(X',Y) = \int \rho_{1/2}(X',X)\,\rho_{1/2}(X,Y)\, dX.
\end{equation}
Let the singular values of $\rho_{1/2}(X',Y)$ be $\sqrt{\lambda_k(X',Y)}$. Then
\begin{equation}
\begin{gathered}
\iint \frac{p^2(X',Y)}{p(X')p(Y)}\, dX'\, dY
= \|\rho_{1/2}(X',Y)\|_2^2
= \sum_{k=1}^K \lambda_k(X',Y),
\end{gathered}
\label{maximization}
\end{equation}
so maximizing mutual information corresponds to maximizing the sum of squared singular values of $\rho_{1/2}(X',Y)$.

\begin{figure}[t]
\centering
\begin{subfigure}{.235\textwidth}
\centering
\includegraphics[width=.87\linewidth]{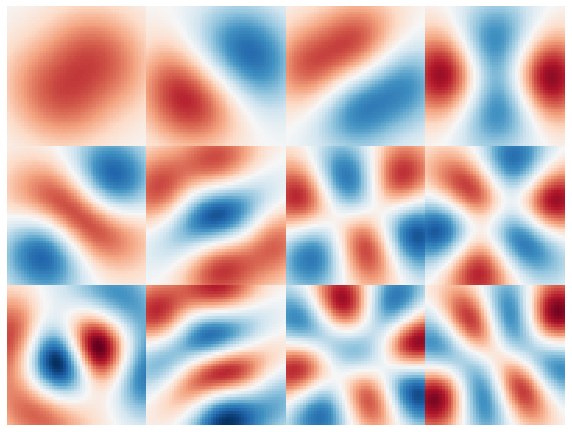}
\caption{\footnotesize $\rho_{1/2}(X', X)$ left singular func.}
\end{subfigure}
\begin{subfigure}{.24\textwidth}
\centering
\includegraphics[width=.9\linewidth]{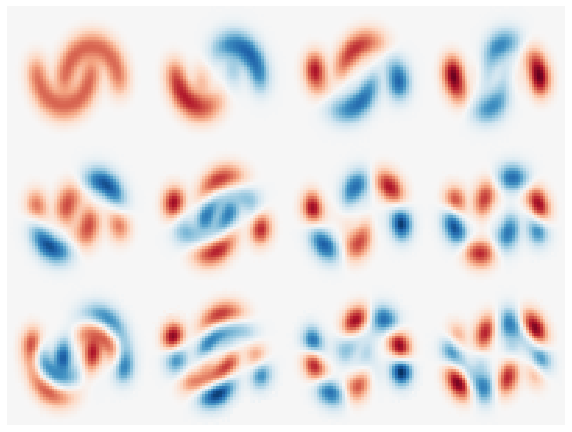}
\caption{\footnotesize $\rho_{1/2}(X', X)$ right singular func.}
\end{subfigure}
\begin{subfigure}{.235\textwidth}
\centering
\includegraphics[width=.87\linewidth]{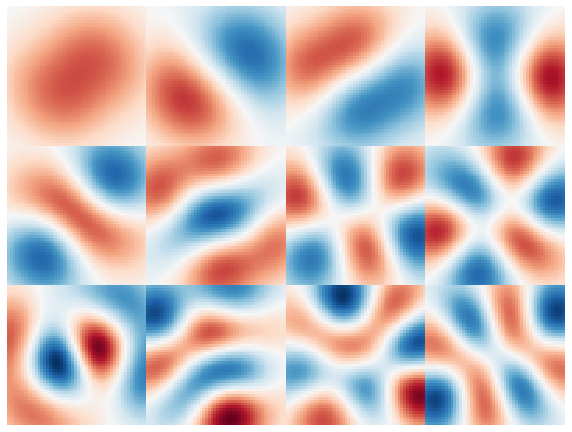}
\caption{\footnotesize $\rho_{1/2}(X', Y)$ left singular func.}
\end{subfigure}
\begin{subfigure}{.24\textwidth}
\centering
\includegraphics[width=.9\linewidth]{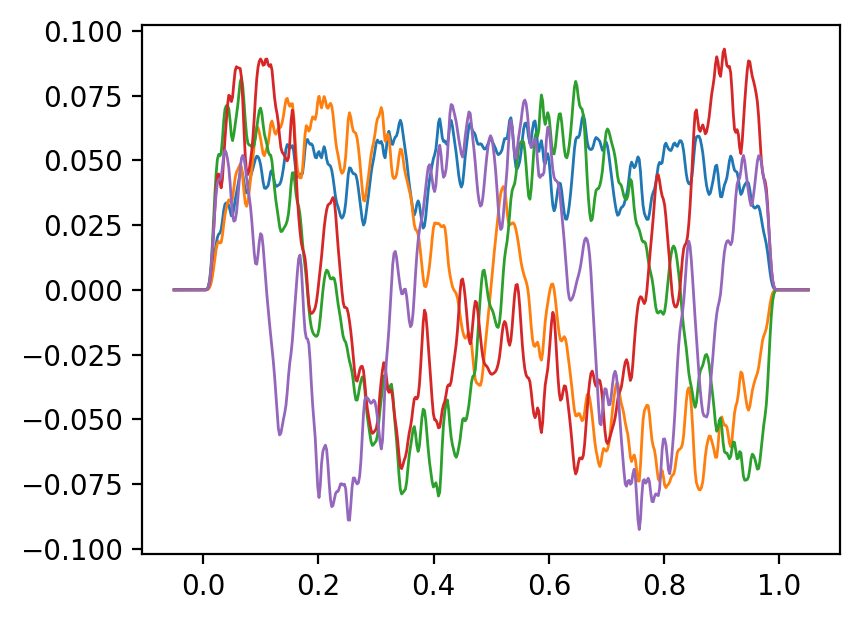}
\caption{\footnotesize $\rho_{1/2}(X', Y)$ right singular func.}
\end{subfigure}\vspace{10pt}
\begin{subfigure}{.5\textwidth}
\centering
\includegraphics[width=.55\linewidth]{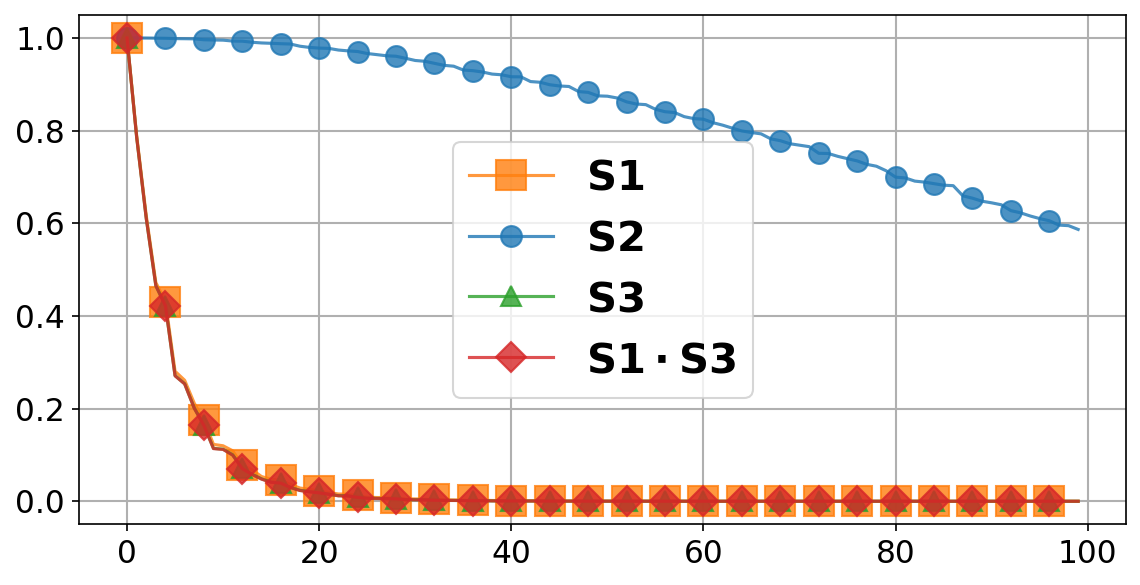}
\caption{Singular-value spectra: $\mathbf{S1}$: $\rho_{1/2}(X',X)$ (fixed); $\mathbf{S2}$: $\rho_{1/2}(X',Y)$ (learned); $\mathbf{S3}$: $\rho_{1/2}(X, Y)$ (almost strictly dependent); $\mathbf{S2\odot S3}$ denotes the componentwise product between every singular value of $\rho_{1/2}(X',X)$ and $\rho_{1/2}(X, Y)$.}
\end{subfigure}
\caption{The Nyström-style analysis of the setting where we remove the decoder and maximize statistical dependence for the encoder alone. Generalized features can still be learned, but only under the assumption of additive Gaussian input noise.}
\label{result_singular_functions_new}
\end{figure}

Fig.~\ref{result_singular_functions_new} reports the results of this direct maximization without a decoder. The key finding is that dependence maximization is stable and produces meaningful projections only when the input Gaussian noise assumption is present. Ideally, the density-ratio functions associated with $p(X',X)$, $p(X',Y)$, and $p(X,Y)$ satisfy:
\begin{itemize}[leftmargin=*]
    \item The singular values of the ratios for $\rho_{1/2}(X',X)$ and $\rho_{1/2}(X',Y)$ match, i.e., substituting $X$ by $Y$ does not reduce dependence.
    \item The singular values of the ratio for $\rho_{1/2}(X,Y)$ are as close to $1$ as possible whenever the singular values for $\rho_{1/2}(X',X)$ are nonnegative.
    \item The right singular functions of $\rho_{1/2}(X',X)$ match the left singular functions of $\rho_{1/2}(X,Y)$ so that, under composition, the intermediate modes cancel appropriately.  
\end{itemize}

Empirically, larger input noise variance $v_X$ tends to improve generalization. For sufficiently large $v_X$, the learned features becomes extremely close to standard autoencoder features.

This example serves as a primary motivation for our feature-learning approach based on statistical dependence maximization (see the MNIST experiment in Appendix~\ref{sec:feature_learning_beyond_analysis}).\vspace{10pt}

We have considered three cases in this section: (i) the standard autoencoder (Fig.~\ref{result_autoencoder33333}), (ii) an autoencoder with a mixture decoder (Fig.~\ref{result_autoencoder}), and (iii) removing the decoder and optimizing only the encoder (Fig.~\ref{result_singular_functions_new}). These are precisely the regimes in which the learned autoencoder features can be reproduced using a discretized Nyström-style analysis, as can be verified by directly comparing the corresponding feature heatmaps. For the toy 2D datasets, all three cases yield results equivalent to those obtained by training a standard end-to-end autoencoder.

\section{Disadvantage of our method}

A limitation of our method is that, although our series of costs avoids the explicit estimation of expectations under the product of marginals required by MINE, it does not directly estimate Shannon's mutual information $\iint p(X,Y)\log \frac{p(X,Y)}{p(X)p(Y)} dXdY$ but $\iint \frac{p^2(X,Y)}{p(X)p(Y)} dXdY$, the quantity without the $\log$. For convenience, we refer to this quantity as a R\'enyi's mutual information. More precisely the order-2 R\'enyi's mutual information. The main question, then, is whether the two forms of statistical dependence  behave differently in practice when estimated or maximized. 

\subsection{Statistical dependence in parameter initialization}

To investigate this, we revisit the experiments in Appendix~\ref{section_extra_tables}, where we showed that the statistical dependence between the data $X$ and the learned features $Y$ increases during training of the autoencoder, even when the optimization is simply minimizing the reconstruction MSE. In those experiments, we estimated the dependence at each training iteration and observed that the estimate increases and eventually converges. Here we examine this experiment more closely.

A natural starting point to investigate is the network at initialization. Normally, we should expect the dependence between $X$ and $Y$ to be small before any training of the autoencoder, since the features have not yet been optimized to encode useful information about the data. Under independence, Shannon's mutual information should attain its minimum value of $0$, whereas R\'enyi's mutual information should attain its minimum value of $1$ since spectrum will contain one and only one positive singular value at $1$. Therefore, if the features are statistically independent of the data at initialization, the estimated dependence should be close to $0$ for Shannon's mutual information and close to $1$ for R\'enyi's mutual information (without the $\log$).

However, this expectation does not hold in the fully deterministic setting in practice. When there is no injected noise and the network is fixed at random initialization, the feature $Y$ is still a deterministic function of $X$, which makes the statistical dependence value between $X$ and $Y$ very large. This is shown in Table~\ref{table9difference} for the two-moon dataset under the deterministic setting. Before any training, our estimator of R\'enyi's mutual information is already very large, around 1500, far from its lower bound of 1. This contradicts the naive expectation that an untrained network should only show weak dependence between inputs and features.

MINE, which estimates Shannon mutual information, behaves somewhat better in this regard (Table~\ref{table9difference}): at initialization it gives a value of about 1.15, which is closer to its lower bound of 0. Nevertheless, its estimates at later iterations converges to around 4 and show little separation across training stages. Thus, even though the Shannon-based estimate is somewhat more interpretable at initialization, it is still unclear whether it provides a more reliable quantification of dependence in a completely static setting.

It is reasonable to take the $\log$ of R\'enyi's mutual information, which gives us a value around $\log 1500 \approx 7.31$. But this transformation changes only the scale, not the qualitative behavior. The estimates still fail to differentiate the later stages of training, because the underlying values remain clustered around 1500.

\begin{table}[H]
\centering
\scriptsize
\setlength{\tabcolsep}{2.5pt}
\renewcommand{\arraystretch}{1.3}
\begin{tabular}{c c c c c c c c}
\toprule[1.2pt]
\multicolumn{1}{c}{\textbf{Measurement}} &
\multicolumn{7}{c}{\textbf{Iterations ($\times 10^3$)}}  \\
\cmidrule(lr){2-8}
& 0 & 1 & 2 & 3 & 4 & 5 & 6 \\
\midrule 
MSE ($\times 10^{-3}$) & 29.9 & 1.28 & 1.51 & 1.00 &  0.95 & 1.08 & 0.98\\
Shannon’s MI & 1.15 & 4.08 & 4.10 & 3.93 & 3.82 &
       3.84 & 3.85 \\ 
R\'enyi's MI & 1479.41 & 1499.53 & 1521.31 & 1515.33 &
       1485.18 & 1481.55 & 1544.81 \\
\bottomrule[1.2pt]
\end{tabular}\vspace{-6pt}
\caption{Results on the two-moon dataset in a fully deterministic setting without injected noise. At initialization, the MINE estimate of Shannon mutual information is closer to its lower bound of $0$, whereas the Rényi's mutual information is far from its lower bound of $1$. However, at later training iterations (e.g., 1000, 2000, and beyond), the Shannon's mutual information estimates produced by MINE show little clear separation and appear noisy, even though the dependence is expected to increase as the reconstruction MSE decreases.}
\label{table9difference}
\end{table}

We have found two practical modifications that can make the estimated dependence value start near its lower bound of $1$ at initialization. Their results are summarized in Table~\ref{proposed_fix222}.

The first possible modification, which is also the approach used in this paper, is to add Gaussian noise to the features. This successfully brings the initial dependence estimate close to 1, but it also introduces a trade-off: larger noise variance tends to limit the best achievable reconstruction accuracy, as reflected in the MSE values in Table~\ref{proposed_fix222}. 

The second possible modification is to concatenate the input with a sufficiently large number of independent uniform noise dimensions. At random initialization, the resulting features are influenced heavily by these noise coordinates, which dilutes the dependence between the data and the learned representation. This also makes the initial dependence estimate close to its minimum value of 1.

As shown in Table~\ref{proposed_fix222}, using either of the modifications, additive Gaussian noises or concatenated input noises, can bring the statistical dependence value at the initialization of the autoencoder to the expected lower bound value.


\begin{table}[H]
\centering
\scriptsize
\setlength{\tabcolsep}{2.0pt}
\renewcommand{\arraystretch}{1.3}
\begin{tabular}{r c c c c c c c c}
\toprule[1.2pt]
\multicolumn{1}{c}{\textbf{Proposed fix}} &
\multicolumn{7}{c}{\textbf{Iterations (\(\times 10^3\))}}  \\
\cmidrule(lr){2-9}
& 0 & 1 & 2 & 3 & 4 & 5  & 20 & 50\\
\midrule 
\textbf{(i) Additive feature noise} \\ 
MSE (\(\times 10^{-3}\)) & 30.2 & 2.16 & 1.65 & 1.64 & 1.70 & 1.51 & 1.20 & 0.99 \\
R\'enyi's MI & 1.01 & 27.89 & 27.84 & 27.87 & 27.95 & 27.88 & 28.49 & 28.78  \\
\midrule
\textbf{(ii) Concatenated input noise} \\
MSE (\(\times 10^{-3}\)) & 37.3 & 1.75 & 1.49 & 1.37 & 1.27 & 1.16 & 0.83 & 0.54  \\
R\'enyi's MI & 1.02 & 39.01 & 49.56 & 55.00 & 57.90 & 64.10 & 79.46 & 97.79 \\
\bottomrule[1.2pt]
\end{tabular}\vspace{-6pt}
\caption{Two modifications that correct the high dependence estimate at initialization in the deterministic setting. Both additive feature noise and concatenated input noise make the estimated R\'enyi's mutual information start close to its lower bound of 1 before training.}
\label{proposed_fix222}
\end{table}

The common characteristic of both fixes is the presence of noise that is resampled at every iteration. In all these experiments, we fix the sample size and the batch size to 10000, and assume that only these 10000 samples are available at each iteration. At each iteration, the network receives the same dataset. The only source of variation across iterations is the injected noise, either added to the features or concatenated to the inputs. Such noises that are unfixed and changing at each iteration, independent of the data, is what is needed to make the measure meaningful. 

\subsection{Empirical pdf estimator: estimation bias and sample efficiency}

This still leaves open the question of why the estimator fails in the fully deterministic setting. There can only be two potential reasons why the estimator gives a overly high value, either the two variables are overly dependent or they are too discrete.  However, since in the beginning of training the parameters are randomized, the features are expected to not contain any information from the data without training, thus independent, so the second potential reason that the sample space is too discrete seems more plausible that the joint distribution is insufficiently smooth for this estimator. Both additive and concatenated noise may be viewed as smoothing mechanisms that regularize the joint distribution and make the estimator better behaved. 

We next investigate whether Shannon's mutual information is more robust than the R\'enyi's measure when the empirical samples are overly discrete, or whether both measures behave similarly.

\begin{figure}[H]
\centering
\begin{subfigure}{0.16\textwidth}
  \centering
  \includegraphics[width=1\linewidth]{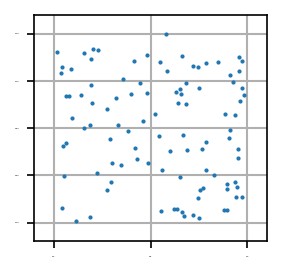}\vspace{-5pt}
  \caption{100 samples}
  \label{17a}
\end{subfigure}\hspace{-7pt}
\begin{subfigure}{0.16\textwidth}
  \centering
  \includegraphics[width=1\linewidth]{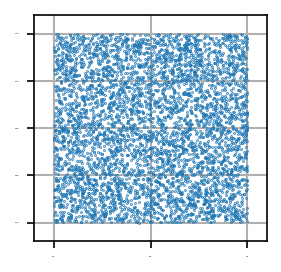}\vspace{-5pt}
  \caption{1000 samples}
  \label{17b}
\end{subfigure}\hspace{-5pt}%
\begin{subfigure}{0.16\textwidth}
  \centering
  \includegraphics[width=1\linewidth]{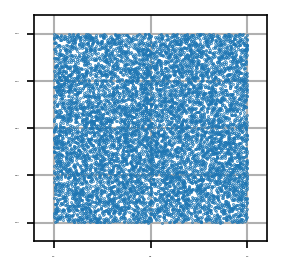}\vspace{-5pt}
  \caption{10000 samples}
  \label{17c}
\end{subfigure}\vspace{-5pt}
\caption{Toy example illustrating the effect of sample discreteness on dependence estimation. Although the two variables are truly independent, finite-sample estimates can be biased when the empirical distribution is too sparse or discrete. As the number of samples increases, the estimated dependence value may first increase and then gradually decrease toward its true value at the minimum.}
\label{figure_4444444}
\end{figure}

To illustrate this, we consider a simple toy example in which two random variables are independently drawn from uniform distributions. Since the variables are independent, the true statistical dependence is 0 for Shannon's mutual information, and 1 for the R\'enyi's measure. Fig.~\ref{figure_4444444} shows the empirical samples of different sizes.

Let us illustrate this effect in this simple setting. Consider two independent uniformly distributed random variables. Suppose we have only \(100\) joint samples from them (Fig.~\ref{17a}). With so few available samples, these samples may effectively be treated as \(100\) isolated discrete points. As a result, the statistical dependence can be severely overestimated: although the true joint distribution is independent, the estimator may assign a large value, potentially around \(100\), because the singular value decomposition of the estimated density ratio may produce \(100\) singular values close to \(1\), as if every sample point were fully distinct. This is far from the minimal value of \(1\). This situation is different from repeatedly drawing fresh batches of \(100\) samples during optimization. Here, we assume that only a fixed set of \(100\) samples is available. In that case, the statistical dependence estimate can be strongly biased upward.

If we increase the sample size, we may initially expect the estimate to increase further, since the estimator may still interpret the additional samples as more discrete points at the current resolution (Fig.~\ref{17b}). However, this trend cannot continue indefinitely. Once the sample size becomes sufficiently large, the density can be estimated more smoothly and accurately, and the bias should decrease, causing the estimate to approach its true value of \(1\) (for example, Fig.~\ref{17c}). This is essentially a finite-sample resolution effect: the estimate becomes less biased only after enough samples are available relative to the model resolution.

This analysis suggests that the dependence estimate may first increase and then decrease, eventually approaching the lower bound of \(1\) in this independent uniform example. The shape of this curve, including the location of its peak, depends on the \textit{\textbf{minimum resolution}} of the estimator. In this sense, R\'enyi's mutual information, defined here as the sum of the squared singular values, may be interpreted as the total number of samples that are effective, or equivalently the effective degrees of freedom.

Next, We would like to visualize whether this increase-then-decrease behavior indeed occurs in this simple case of

\begin{figure}[H]
\centering
\begin{subfigure}{0.45\textwidth}
  \centering
  \includegraphics[width=.7\linewidth]{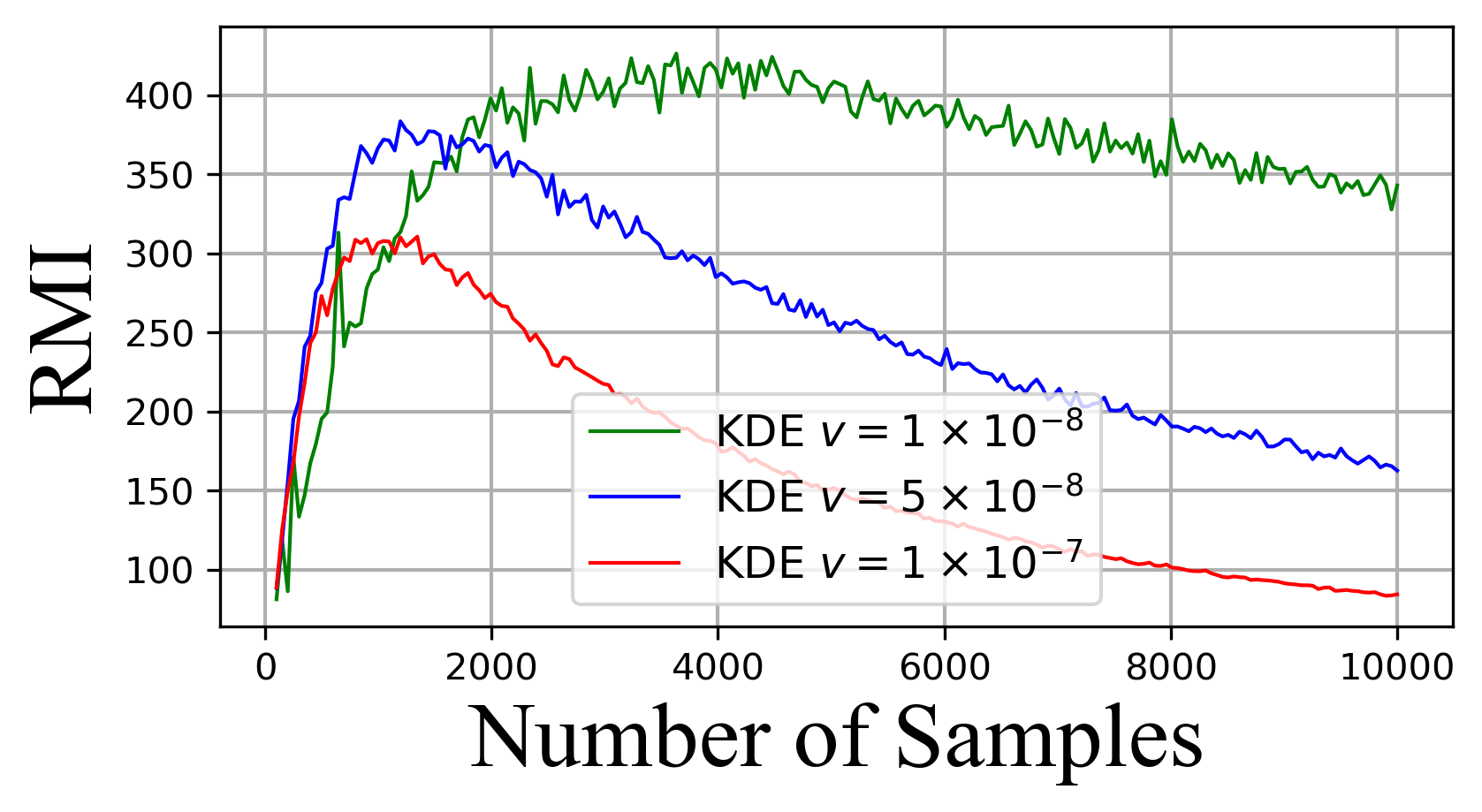}\vspace{-5pt}
  \caption{KDE: evolution of R\'enyi's mutual information as the sample size increases.}
\end{subfigure}\vspace{-1pt}
\begin{subfigure}{0.45\textwidth}
  \centering
  \includegraphics[width=.7\linewidth]{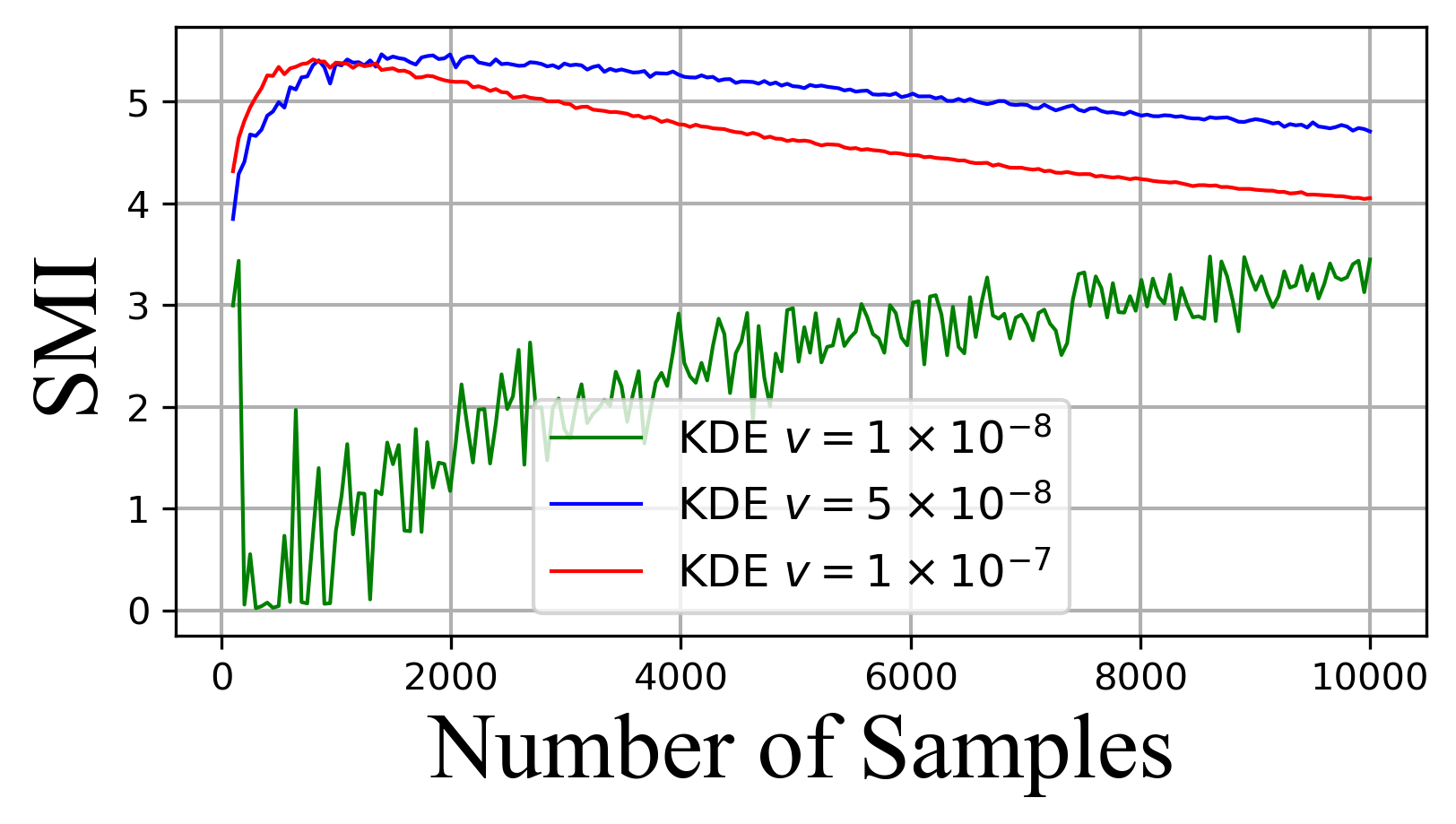}\vspace{-5pt}
  \caption{KDE: evolution of Shannon's mutual information as the sample size increases.}
\end{subfigure}\vspace{-1pt}
\begin{subfigure}{0.45\textwidth}
  \centering
  \includegraphics[width=.7\linewidth]{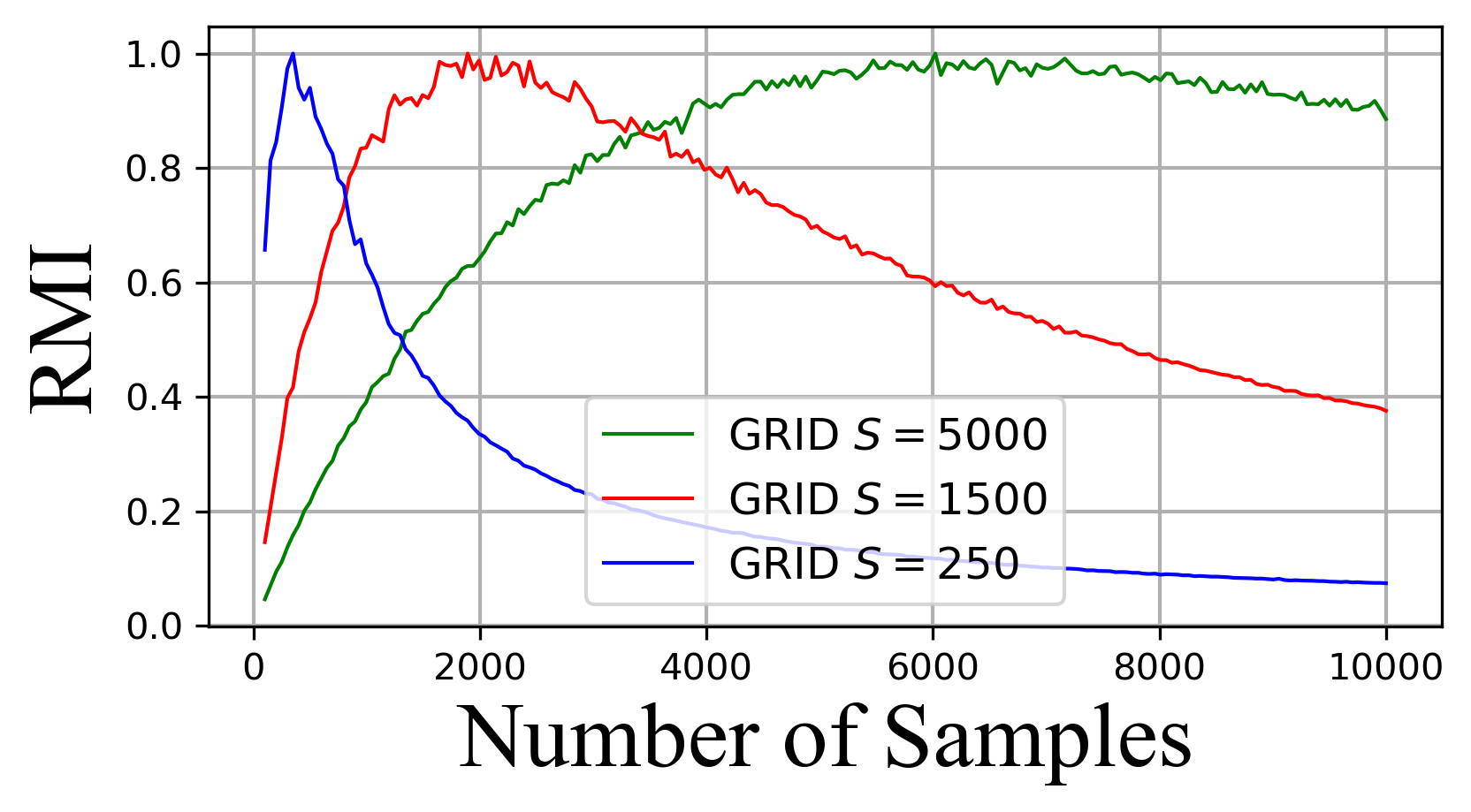}\vspace{-5pt}
  \caption{Grid-based estimation: evolution of R\'enyi's mutual information (RMI) as the sample size increases. For visualization, each curve is normalized by its maximum value; the same normalization is used in (d).}
  \label{figure18c}
\end{subfigure}\vspace{-1pt}
\begin{subfigure}{0.45\textwidth}
  \centering
  \includegraphics[width=.7\linewidth]{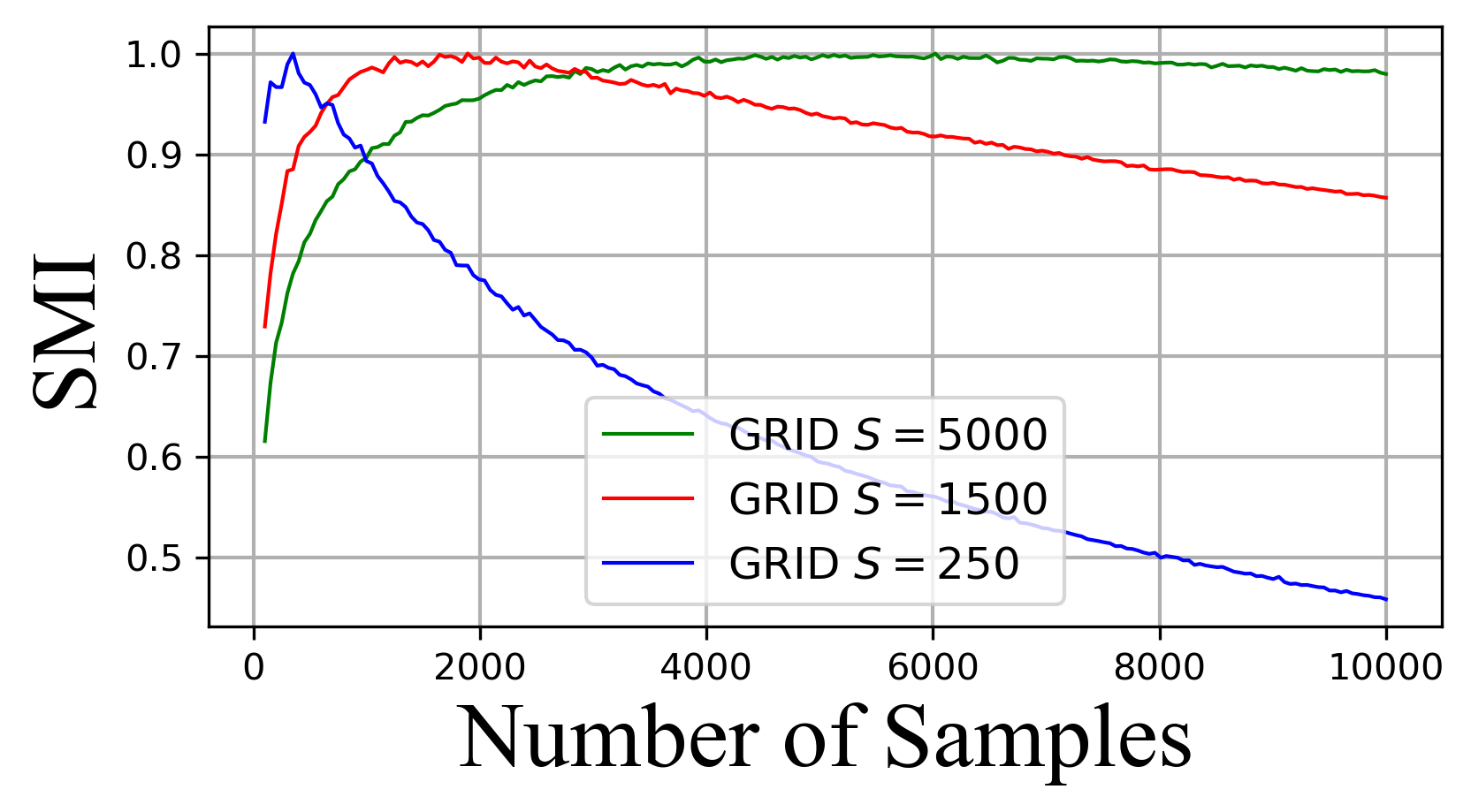}\vspace{-5pt}
  \caption{Grid-based estimation: evolution of Shannon's mutual information (SMI) as the sample size increases.}
  \label{figure18d}
\end{subfigure}\vspace{-5pt}%
\caption{Experiments on sample sufficiency versus parameters controlling estimator resolution: the kernel variance \(v\) in KDE and the grid size \(S\) in the grid-based estimator.}
\label{figure_18_results}
\end{figure}

\noindent independent uniformly distributed variables in 2D. Since the distribution is simple, we can study this effect without neural networks by using standard pdf estimators. There are two conventional approaches: partitioning the 2D sample space with grid lines, or using a Kernel Density Estimator (KDE).

In both cases, we estimate the joint pdf \(p(X,Y)\) from samples, marginalize it to obtain \(p(X)\) and \(p(Y)\), construct the density ratio $\frac{p(X,Y)}{p(X)p(Y)}$, and then compute directly the dependence measure from this ratio.

The first approach uses kernels. We choose a Gaussian kernel with variance \(v\), and approximate the pdf by placing one kernel at each sample point. The variance \(v\) controls the resolution. If \(v\) is large, only a small number of samples is needed to obtain a smooth estimate, but the estimate may be heavily biased. If \(v\) is small, the estimator has higher resolution, but many more samples are required to fill the gaps between points and avoid a highly discrete estimate.

The second approach uses a discrete grid of size \(S \times S\). Here, the grid size \(S\) controls the resolution. If \(S\) is very large, the space is partitioned into many cells, and a large number of samples is needed to obtain a smooth estimate. If \(S\) is small, fewer samples are required, but the estimate becomes coarse and can again be strongly biased.

We present the results in Fig.~\ref{figure_18_results}. We estimate both Shannon's mutual information and R\'enyi's mutual information, and track how their estimated values evolve as the sample size increases. We vary the parameters that control the estimator resolution: the kernel variance \(v\) for KDE, and the grid size \(S\) for the \(S \times S\) partition of the sample space.\vspace{9pt}

The results are consistent with our hypothesis, especially in showing the characteristic non-monotonic behavior: the estimated values first increase and then decrease. The resolution parameters determine both the shape of the curve and the location of its peak, and therefore determine how many samples are needed for the estimate to become reliable.

For any fixed resolution, once a sufficient number of samples is available, the estimate approaches the value expected for independent variables. 

These experiments also suggest that Shannon's mutual information and R\'enyi's mutual information have broadly similar sample-size requirements. Shannon's mutual information appears to behave slightly better in this example, but neither measure eliminates the bias caused by insufficient samples. Thus, for the purpose of analyzing autoencoder features, the practical difference between them appears limited.\vspace{7pt}

Another third possible approach is to use a Gaussian Mixture Model (GMM). In this case, we first fit a GMM to the samples, and then compute the statistical dependence measure from the estimated pdf. Unlike KDE, which places one kernel at each sample point, a GMM approximates the density using a smaller number of learned Gaussian components. The number of mixture components therefore may play a role of controlling the resolution: the more components the GMM uses, the finer the structure it can represent, and the more samples may be required for estimating the statistical dependence. We leave this example to the readers.

\subsection{Neural network dependence estimator: estimation bias and sample efficiency}

Next, the question is whether the same conclusion also holds when the dependence measure is estimated by the neural network estimator. The previous Fig.~\ref{figure_18_results} shows the behavior obtained with model-free pdf estimators. Here, we investigate whether a neural estimator trained with the NMF-like cost introduced in this paper has a similar intrinsic minimal resolution.

\begin{figure}[H]
\centering
\includegraphics[width=.75\linewidth]{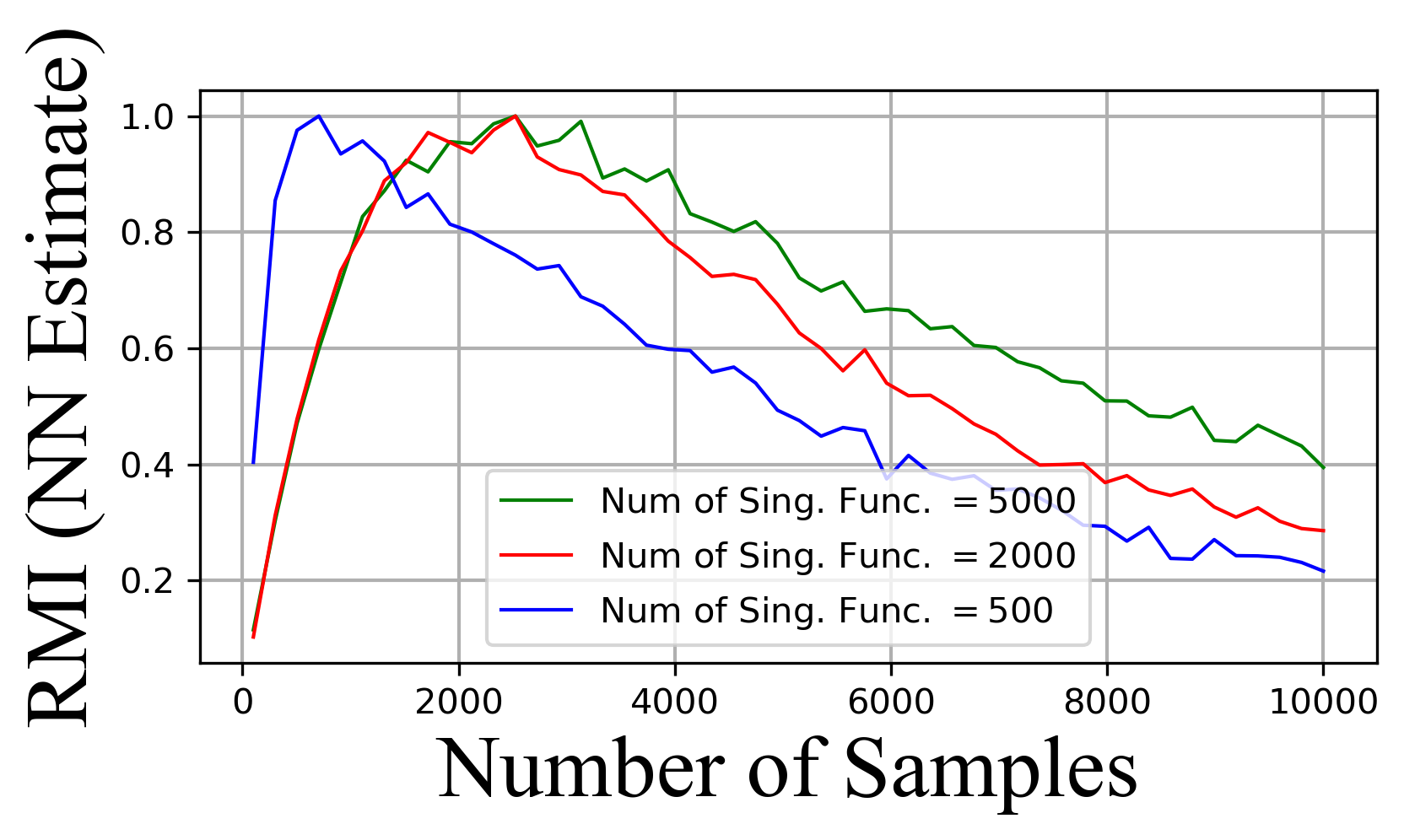}\vspace{-5pt}
\caption{Neural estimation of statistical dependence for the 2D uniformly distributed independent variables in Fig.~\ref{figure_4444444}. Similar Fig.~\ref{figure_18_results}, the estimated value first increases and then decreases as the sample size grows, indicating that the neural estimator also suffers from a resolution-induced bias when the number of samples is insufficient.}
\label{network_curve}
\end{figure}

We again consider the 2D example of two independent uniformly distributed variables in Fig.~\ref{figure_4444444}. We gradually increase the number of samples from $100$ to $10000$, and train the two neural networks on the corresponding sample pairs drawn from the two independent variables. At each iteration, the entire sample set is fed to the network as a full batch. The resulting curves are shown in Fig.~\ref{network_curve}. Similar to Fig.~\ref{figure18c} and Fig.~\ref{figure18d}, each curve is normalized by its maximum value so that all curves can be plotted on the same scale.

From Fig.~\ref{network_curve}, we again observe the same qualitative behavior: the estimated dependence first increases and then decreases as the sample size grows. In these experiments, training was stopped after $50000$ iterations, since the cost continued to improve slowly even after prolonged optimization. Nevertheless, for a fixed training number of iterations, the final converged estimated value clearly decreases when more samples are provided. In fact, with around $100000$ samples, the optimized value will remain close to the lower bound of $1$.

These results suggest that the neural estimator also has an effective minimal resolution, similar to that of the two pdf estimators. When the network is too small relative to the model capacity, the sample space is effectively treated as overly discrete, leading to an overestimation of statistical dependence. We also find that the number of singular functions used in the NMF approximation, i.e., the number of outputs in the two multi-output networks, affects the shape of these curves. Although such behavior is not ideal, since we would like the estimator to be as unbiased as possible, it indicates that the number of singular functions acts as a capacity or resolution parameter.

We can further compare the curves in Fig.~\ref{network_curve} obtained from the neural estimators with those obtained from the pdf estimators in Fig.~\ref{figure_18_results}. For example, the curve obtained with $2000$ singular functions is quite similar to the curve in Fig.~\ref{figure18c} corresponding to grid size $S=1500$ for the R\'enyi mutual information estimator. This suggests that a neural estimator with $2000$ outputs has an effective resolution comparable to that of a fine discretization of the 2D sample space, and that this resolution largely determines the model capacity. As expected, given sufficiently many samples, the estimate eventually decreases toward $1$.\vspace{9pt}

This analysis suggests that estimating statistical dependence in a deterministic, static network may be ill-posed when the estimator resolution is much higher than the number of available samples. In that case, the sample space is not smooth enough, and the estimator behaves as though it were fitting isolated discrete points, similar to approximating a density from only a few samples using an excessively fine grid.


The previous experiments examined only simple independent uniform variables. We now investigate how these findings extend to measuring statistical dependence between autoencoder inputs and features.

The insufficient sample size may prevent the autoencoder from achieving the behavior desired, namely, shrinking local Gaussian neighborhoods until the largest universal radius is found for which the density-ratio matrix becomes approximately diagonal across all samples. We therefore perform one final test by returning to the original two-moon autoencoder example, where the goal is to measure the statistical dependence between the data and the learned features.

Based on the previous analysis, we expect that increasing the number of available training samples should reduce the bias of the statistical dependence estimate. A second possibility is to use a large dataset and, at each iteration, sample only a small subset from it, i.e., to perform stochastic mini-batch optimization for dependence measurement. Both strategies should yield lower, and presumably more accurate, estimates of the dependence between the inputs and the autoencoder features at initialization. The results are shown in Table~\ref{table11difference}, where the number of available samples ranges from $100$ to $200000$. The autoencoder is untrained, initialized randomly. 

\begin{table}[t]
\centering
\scriptsize
\setlength{\tabcolsep}{2.5pt}
\renewcommand{\arraystretch}{1.3}
\begin{tabular}{c c c c c c}
\toprule[1.2pt]
\multicolumn{1}{c}{\textbf{Training Setting}} &
\multicolumn{5}{c}{\textbf{Available Samples}}  \\
\cmidrule(lr){2-6}
& 100 & 1000 & 10000 & 100000 & 200000 \\
\midrule 
All Samples in a Batch & 90.95 & 530.51 & 558.99 & 344.61 & 325.86 \\       
Fix the Batch Size at $1000$ & 93.16 & 186.26 & 134.47 & 131.23 & 131.26 \\
\bottomrule[1.2pt]
\end{tabular}\vspace{-6pt}
\caption{Estimated statistical dependence between the two-moon data and the corresponding autoencoder features as the number of available training samples increases. Two settings are considered: full-batch training using all available samples at each iteration, and stochastic optimization with a fixed batch size of $1000$. In both cases, the estimates eventually decrease as more samples become available, consistent with the sample-resolution analysis above. Training was stopped after $5000$ iterations for the full-batch case and after $50000$ iterations for the fixed-batch case, so some residual bias due to incomplete optimization may remain. No noise is used in this example.}
\label{table11difference}
\end{table}

The same qualitative trend can be shown: the estimated dependence initially increases and then decreases as the available sample size becomes sufficiently large. Moreover, when the total dataset is large, using a fixed mini-batch size of $1000$ gives lower estimates than training on all available samples in a single batch, indicating that repeated exposure to different subsets at each iteration helps reduce the discretization bias, although the estimated dependence still does not converge to the lower bound $1$ at initialization.

\subsection{Further investigating the concatenated input noises for autoencoders}

Among the experiments in this section, the most informative are those in which random noise is concatenated to the inputs. This fix is particularly appealing because it makes the dependence estimate start from its independence lower bound at initialization, while still allowing the model to achieve a low reconstruction error after training the autoencoder. For the two-moon example, we therefore perform one final experiment in which we carefully examine the singular values and singular functions of this case.

In Table~\ref{table9difference}, we showed that in the deterministic, static setting the estimated dependence starts at roughly $1400$ and increases to about $1500$. This value is obviously overestimated for a task of this complexity (mapping 2D two-moon data to 1D features), and suggests that the estimator already produces overestimated values at initialization. By contrast, when we concatenate $50$ dimensions of uniform random noise to the input, so that the encoder input becomes $52$-dimensional ($2$ data dimensions plus $50$ noise dimensions), and resample this noise at every iteration, the estimate starts from the independence lower bound and rises only to about $60$ during training. Importantly, this fox does not materially degrade reconstruction performance: the MSE remains approximately $0.0005$ ($5 \times 10^{-4}$). 

Without concatenated noise, the estimated dependence reaches a level of about $1500$; with concatenated noise, it reaches only about $60$, while the MSE decreases very little. This indicates that we cannot directly estimate the statistical dependence between the inputs and the learned features by naively applying the estimator to deterministic data-feature pairs, since doing so may lead to severe overestimation. 

\begin{figure}[H]
\centering
\hspace{-20pt}\begin{subfigure}{0.45\textwidth}
  \centering
  \includegraphics[width=.6\linewidth]{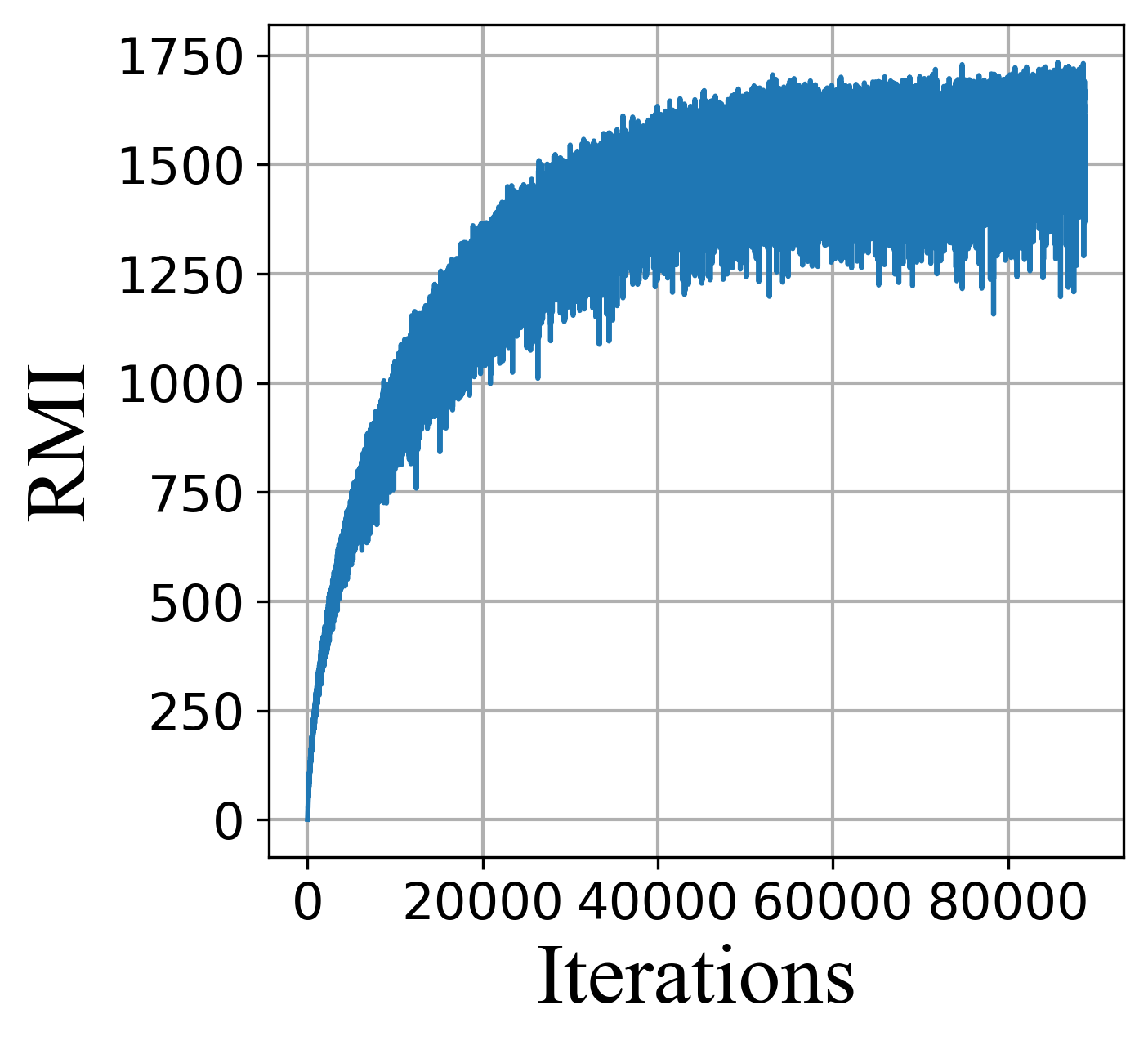}\vspace{-6pt}
  \caption{Static setting: learning curve of the dependence estimator after training the autoencoder.}
  \label{figure20a}
\end{subfigure}
\begin{subfigure}{0.45\textwidth}
  \centering
  \includegraphics[width=.7\linewidth]{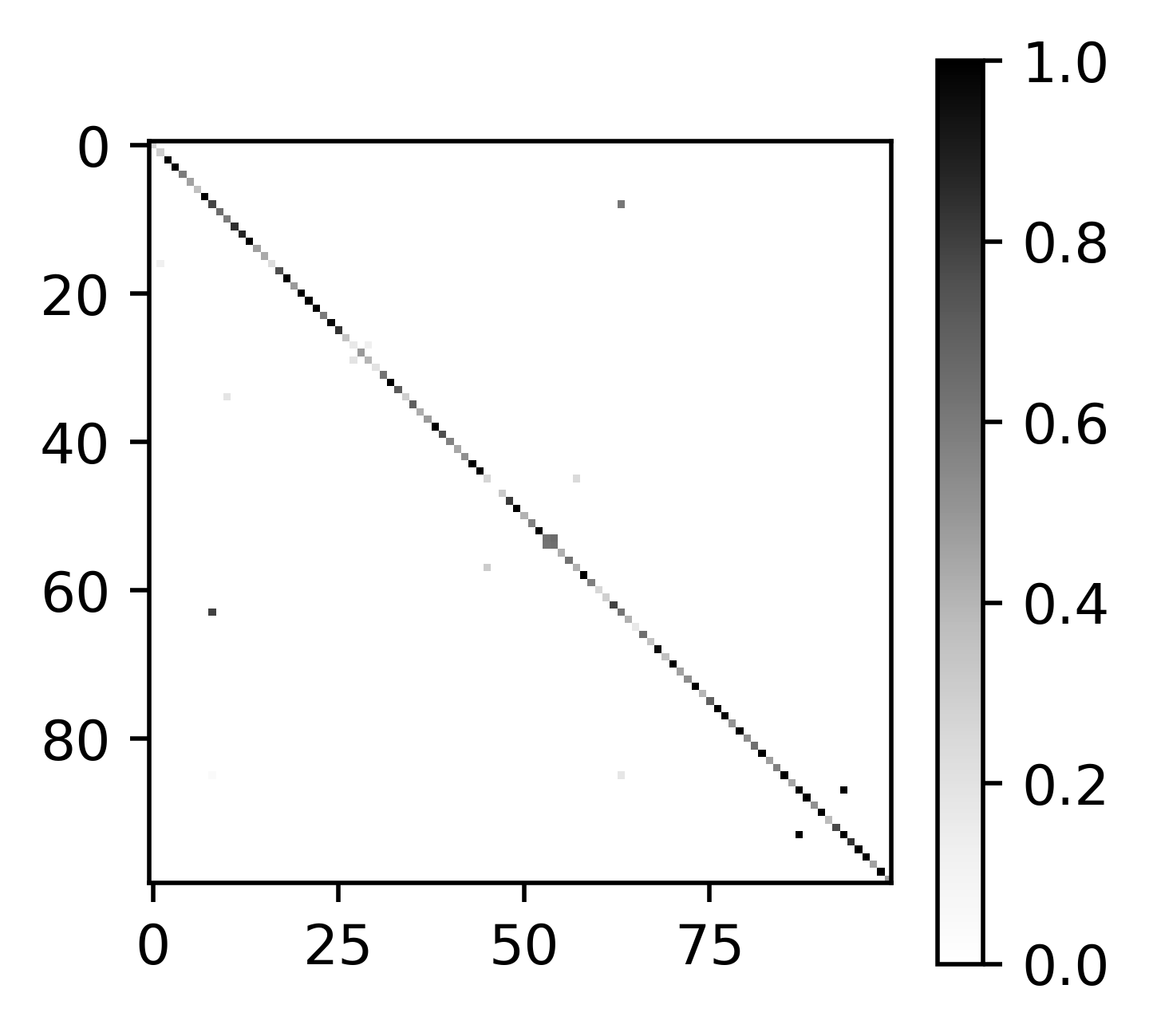}\vspace{-5pt}
  \caption{Static setting: learned pairwise density-ratio matrix.}
  \label{figure20b}
\end{subfigure}\vspace{5pt}
\begin{subfigure}{0.45\textwidth}
  \centering
  \includegraphics[width=.85\linewidth]{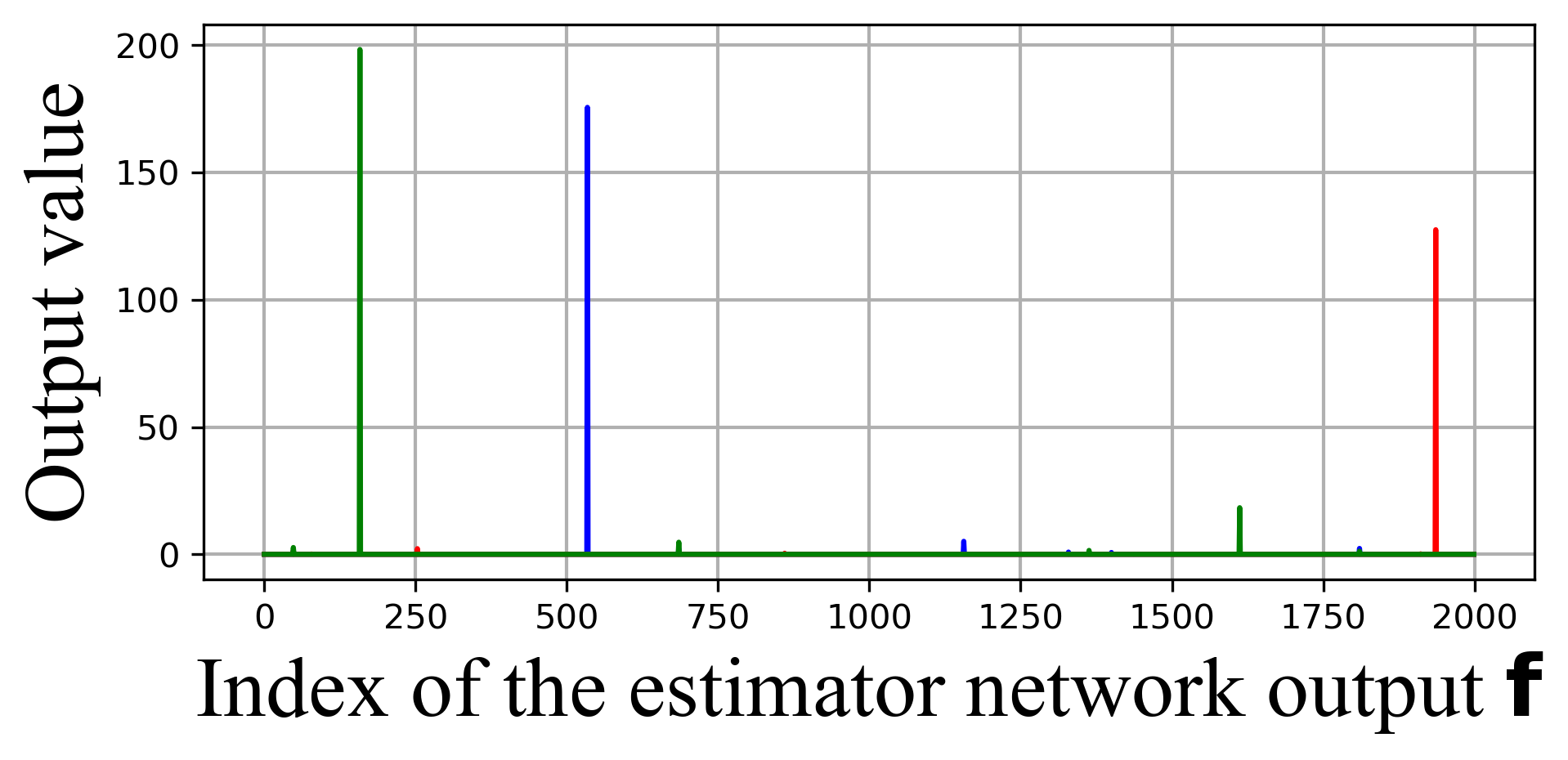}
  \caption{Static setting: the $2000$ post-ReLU activations of the network $\mathbf{f}$ for three representative samples. The activations are nonnegative because of the ReLU and extremely sparse.}
  \label{figure20c}
\end{subfigure}
\caption{Dependence estimation in the fully static setting, with no presence of noise. The autoencoder is first trained to convergence, after which the dependence estimator is trained between the input data and the learned features. (a) Learning curve of the estimated dependence (R\'enyi's mutual information), which converges to a very large value; the number of singular functions is set to $2000$. (b) Learned density-ratio matrix $\frac{p(X,Y)}{p(X)p(Y)}$, visualized using $100$ selected samples from the full set of $10000$ samples, resulting in a $100 \times 100$ matrix. (c) Direct outputs of the estimator network $\mathbf{f}$. The activations are extremely sparse, and the learned density-ratio matrix is close to diagonal.}
\label{figure20plots}
\end{figure}

\begin{figure}[H]
\centering
\hspace{-20pt}\begin{subfigure}{0.45\textwidth}
  \centering
  \includegraphics[width=.6\linewidth]{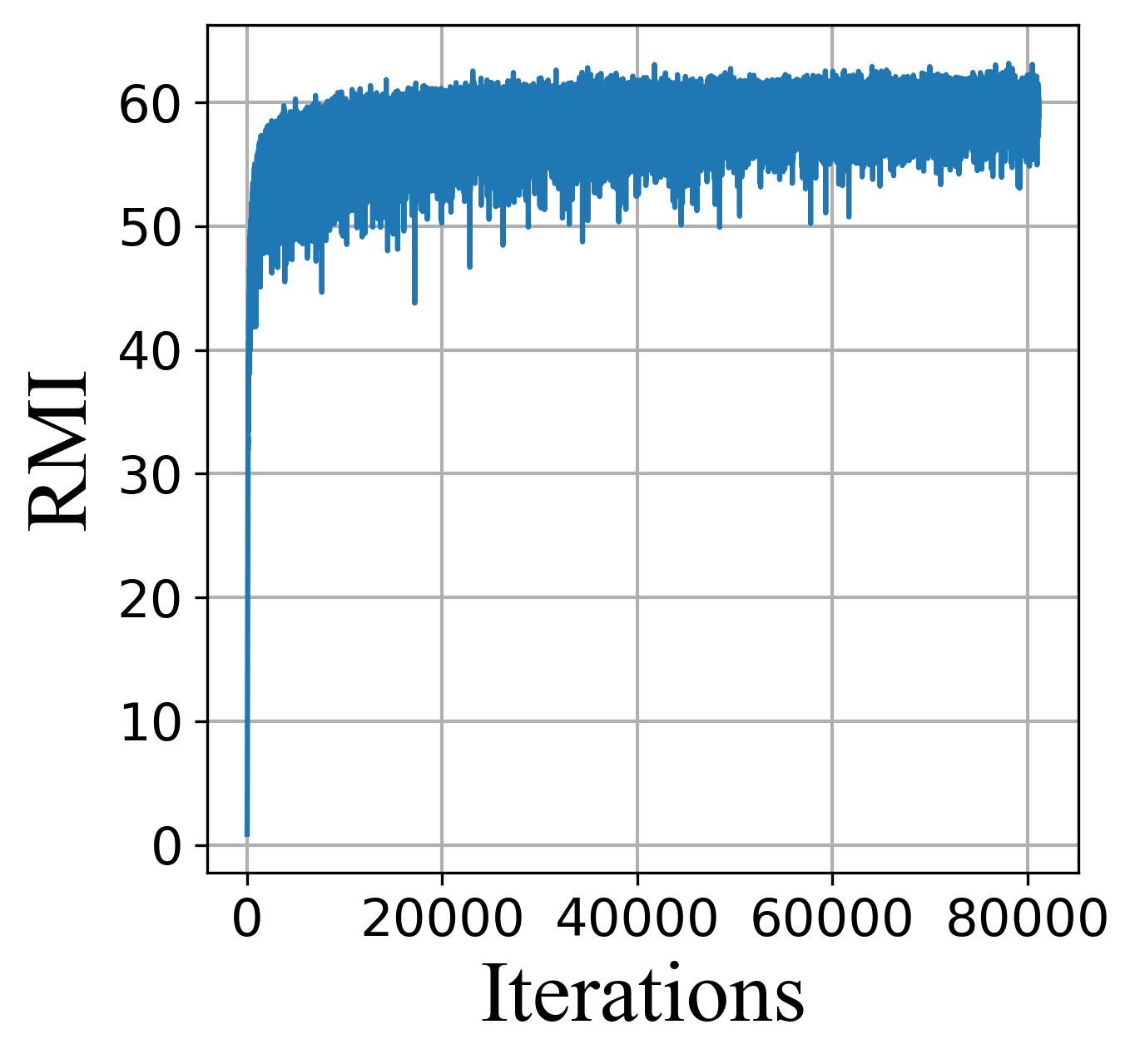}\vspace{-6pt}
  \caption{Stochastic setting: learning curve of the dependence estimator. Compared with Fig.~\ref{figure20a}, the converged value is much smaller while the training remains stable.}
\end{subfigure}
\begin{subfigure}{0.45\textwidth}
  \centering
  \includegraphics[width=.65\linewidth]{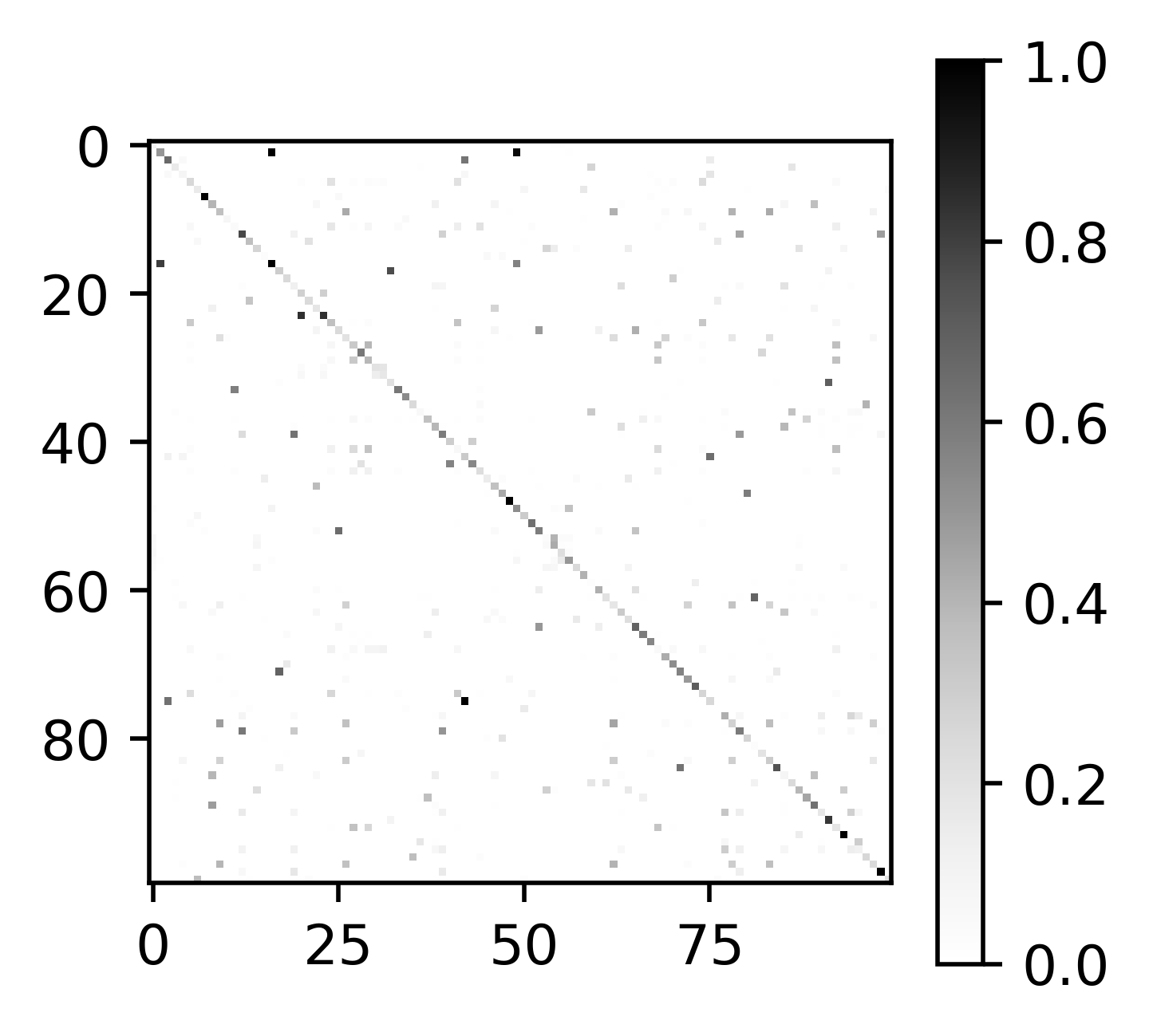}\vspace{-5pt}
  \caption{Stochastic setting: learned pairwise density-ratio matrix. Compared with Fig.~\ref{figure20b}, the matrix is less diagonal and noisier, with many more nonzero entries.}
  \label{21bbbb}
\end{subfigure}\vspace{5pt}
\begin{subfigure}{0.45\textwidth}
  \centering
  \includegraphics[width=0.85\linewidth]{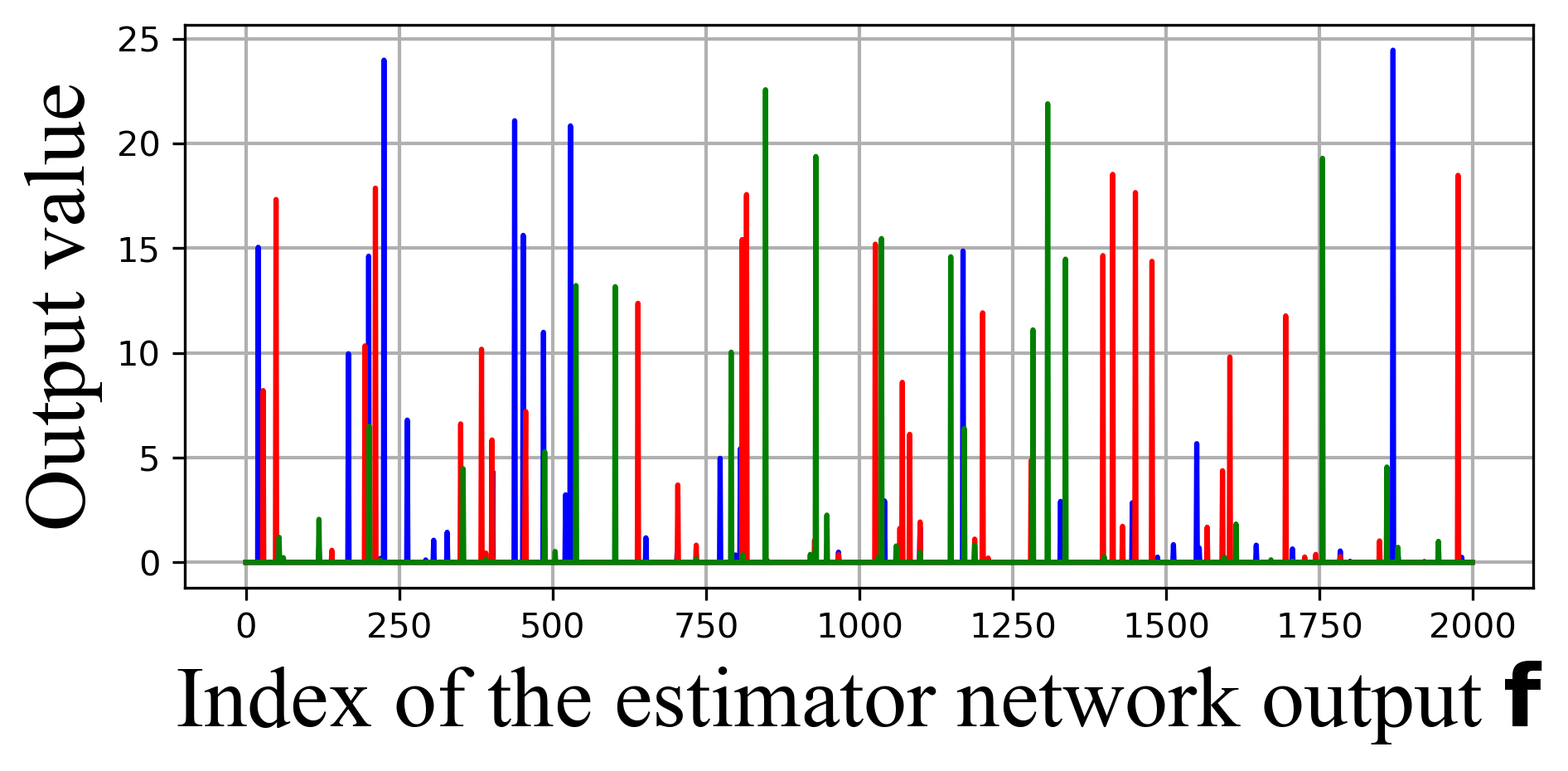}
  \caption{Stochastic setting: the $2000$ post-ReLU activations of the network $\mathbf{f}$ for three representative samples. Compared with Fig.~\ref{figure20c}, the activations are much less sparse.}
  \label{figure21c}
\end{subfigure}
\caption{Direct comparison with Fig.~\ref{figure20plots}, but now $50$ dimensions of uniform random noise are concatenated to the inputs. Since the original data are 2D, the encoder input becomes $52$-dimensional. In this setting, the estimated dependence starts from its independence baseline and converges to about $60$, which is much smaller than the value of about $1500$ in Fig.~\ref{figure20plots}. The learned metric, which is represented by the pairwise density-ratio matrix computed from $100$ selected exemplars out of the full set of $10000$ samples, is much noisier and less diagonal. The outputs of the estimator network $\mathbf{f}$ are also much less sparse. This fix does not sacrifice reconstruction performance, as the MSE remains very similar to that of the noise-free case.}
\label{figure21plots}
\end{figure}

We visualize more of these results in Fig.~\ref{figure20plots} and Fig.~\ref{figure21plots}. We focus on three quantities: (a) the learning curve of the estimated statistical dependence at the end of training, (b) the learned metric, i.e., the pairwise density-ratio metric, and (c) the outputs of the estimator network $\mathbf{f}$, whose output dimension is set to $2000$.

\newpage

Fig.~\ref{figure20plots} corresponds to the fully static setting: we first train the autoencoder, and then train the two estimator networks Fig.~to measure the statistical dependence between the input data and the learned features. Fig.~\ref{figure21plots} shows the corresponding experiment when the inputs are concatenated with uniform random noise that is resampled at each iteration.

Comparing these two figures, we see that the stochastic setting converges to a much smaller dependence estimate (about $60$) than the static setting (about $1400$). In addition, the learned metric and the post-ReLU outputs of the estimator network $\mathbf{f}$ are extremely sparse in the static setting, whereas in the stochastic setting they are noticeably noisier and much less sparse. For the simple two-moon dataset, this shows that concatenating random uniform noise to the inputs may alleviate the overestimation problem.\vspace{10pt}

\begin{figure}[H]
  \centering
\begin{subfigure}{.43\textwidth}
\includegraphics[width=\linewidth]{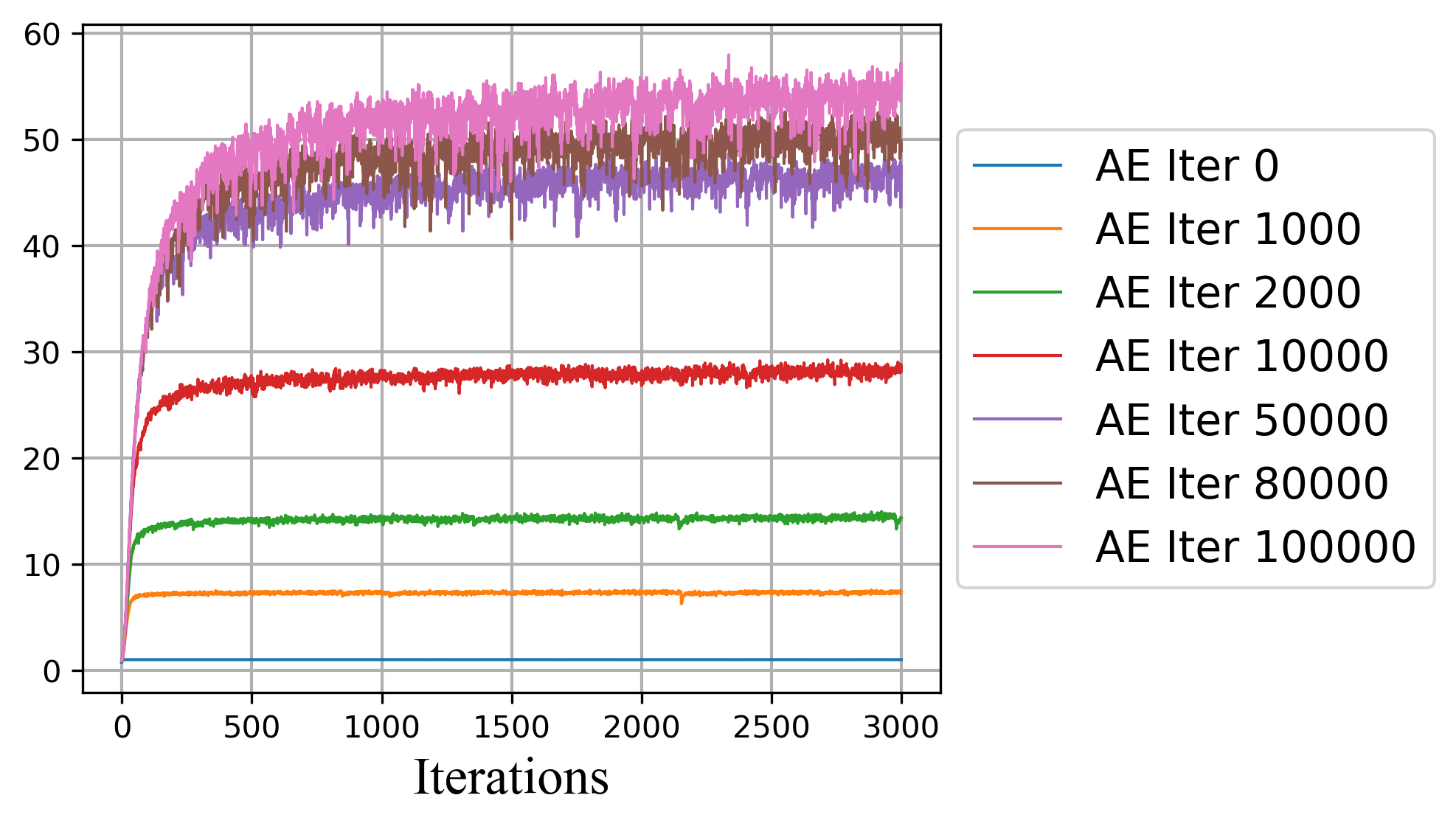}\vspace{-3pt}
\caption{Two-moon with concatenated input noise: learning curves of the dependence estimator at different autoencoder checkpoints. The autoencoder is stopped at iterations ranging from $0$ to $100000$, and the dependence estimator is then trained between the input data and the learned features for this checkpoint. The converged dependence increases gradually and approaches about $60$ after full autoencoder training.}
\label{22a}
\end{subfigure}\vspace{3pt}
\begin{subfigure}{.43\textwidth}
\centering
\includegraphics[width=.8\linewidth]{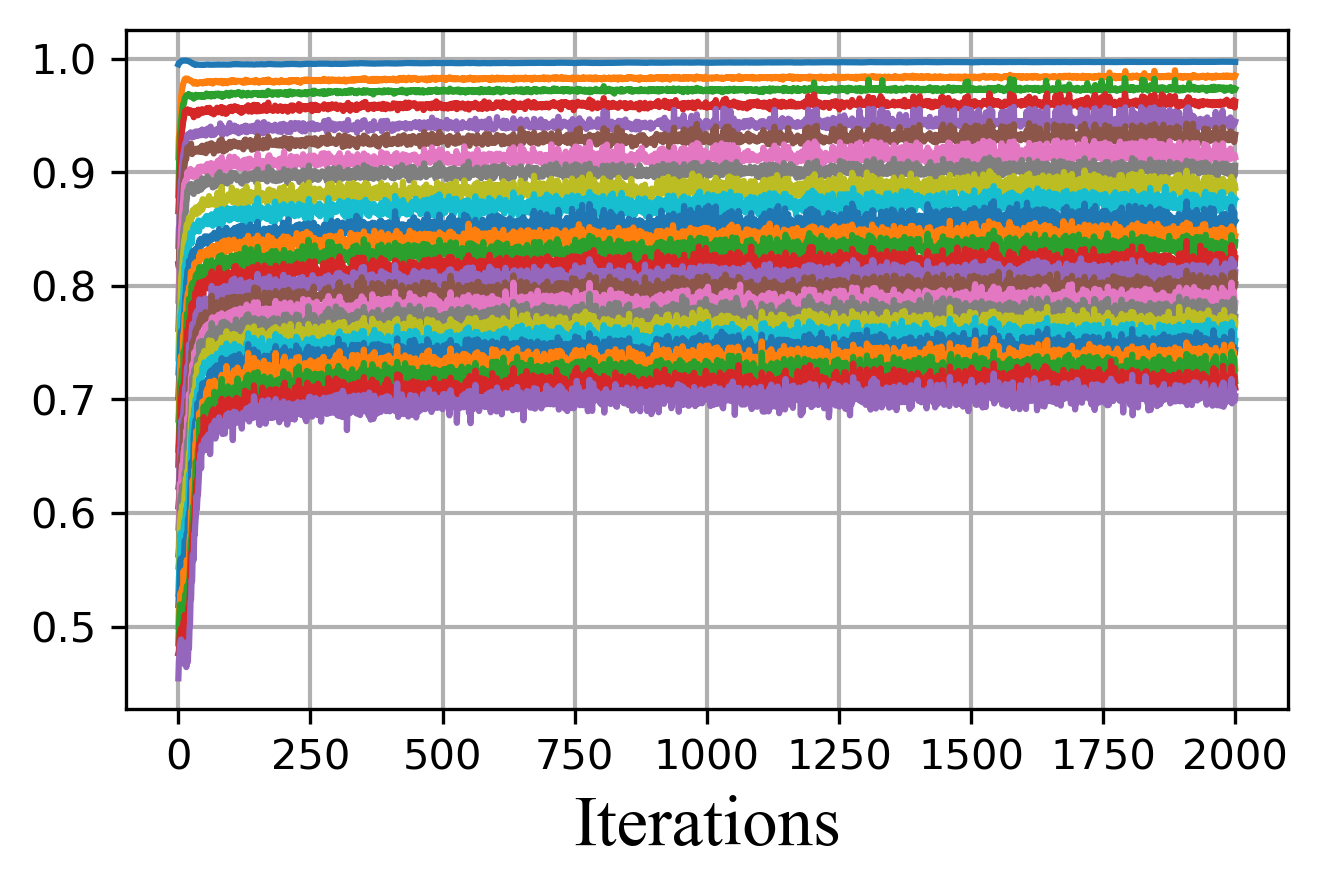}\vspace{-3pt}
\caption{Two-moon with concatenated input noise: the learning curve of the dependence estimator, for the top $50$ singular values after $100000$ autoencoder iterations. A nontrivial spectrum is clearly present, rather than the degenerate solution of the autoencoder in which all singular values converge to $1$.}\vspace{3pt}
\label{22b}
\end{subfigure}\vspace{3pt}
\begin{subfigure}{.23\textwidth}
\centering
\includegraphics[width=1\linewidth]{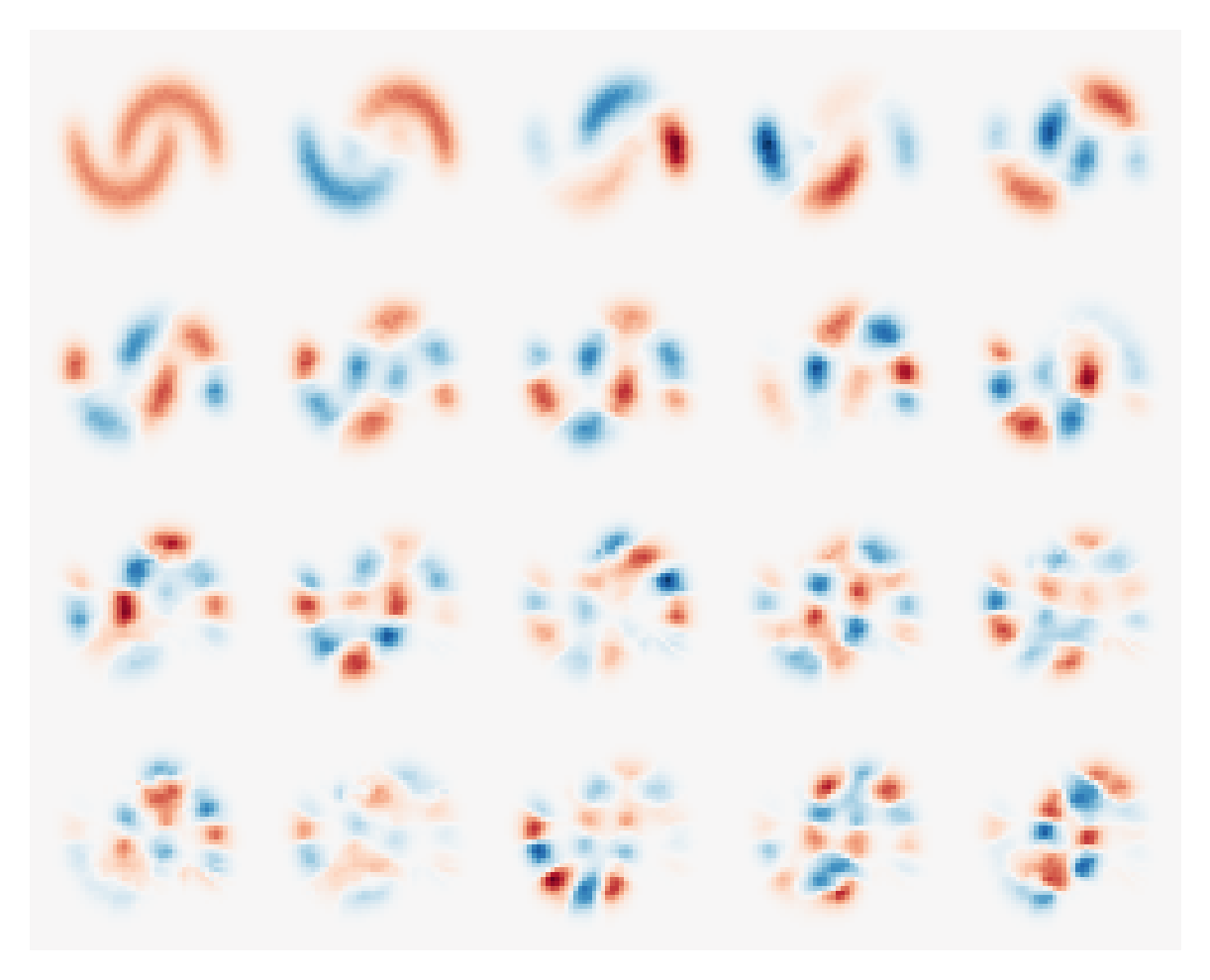}\vspace{-3pt}
\caption{Left singular func. (2D).}
\label{22c}
\end{subfigure}\hfill
\begin{subfigure}{.23\textwidth}
\centering
\includegraphics[width=1\linewidth]{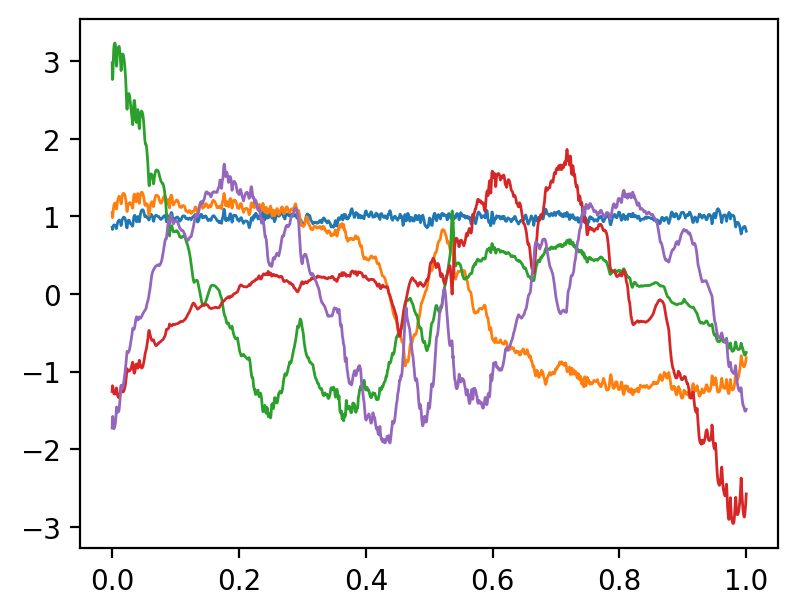}\vspace{-3pt}
\caption{Right singular func. (1D).}
\label{22d}
\end{subfigure}
\caption{Two-moon with concatenated input noise: evolution of the dependence estimate, the singular values, and the singular functions during training. This figure highlights the difference between the noisy-input setting and the fully static setting. In the static case, the estimated dependence remains artificially large (around $1500$) regardless of the training iteration, the singular values collapse to the trivial solution in which they all equal $1$, and the corresponding left and right singular functions are not meaningful.}
\label{figure_ae_iteration}
\end{figure}

Apart from the comparison, we also visualize how the dependence estimate evolves over the course of autoencoder training for the concatenated noise case, shown in Fig.~\ref{figure_ae_iteration}. We stop the training of the autoencoder at multiple checkpoints between $0$ and $100000$ iterations, and at each checkpoint we train the statistical dependence estimator between the input data and the learned features. The autoencoder and the dependence estimator are always trained separately. The learning curves in Fig.~\ref{22a} show that the converged dependence estimate increases as autoencoder training progresses, indicating that the statistical dependence between the data and the features grows during training. When the autoencoder is untrained, the dependence remains at its lower bound, whereas after full training it converges to about $60$.

Fig.~\ref{22b} shows the learning curves of the singular values for the estimator networks after training the autoencoder for $100000$ iterations. The purpose of this plot is to show that the learned density ratio is associated with a nontrivial spectrum. By contrast, in the fully static setting without concatenated input noise, all singular values quickly converge to $1$ rather than forming a meaningful spectrum. Finally, Fig.~\ref{22c} and~\ref{22d} visualize the learned left and right singular functions, showing that they are indeed meaningful in this concatenated-input-noise setting.\vspace{10pt}

We further visualize the learned density-ratio metric in a more direct way. Comparing the learned metric in the static setting (Fig.~\ref{figure20b}) with that in the stochastic setting (Fig.~\ref{21bbbb}), we have shown that the stochastic setting produces a metric that is much less concentrated along the diagonal. So to better understand this behavior, we visualize the metric in the original 2D space.

Note that the density ratio $\frac{p(X,Y)}{p(X)p(Y)}$ is defined for a pair of $X$ and $Y$. Since our dataset contains \(10000\) samples, there are \(10000 \times 10000\) possible pairs. For visualization, we fix one sample \(X\) and evaluate its density ratio against every other sample \(Y\) in the dataset. This yields a heatmap over all other samples. \,We\, show four\, representative exemplars from~{these}

\begin{figure}[H]
\centering
\begin{subfigure}{0.2\textwidth}
  \centering
  \includegraphics[width=.9\linewidth]{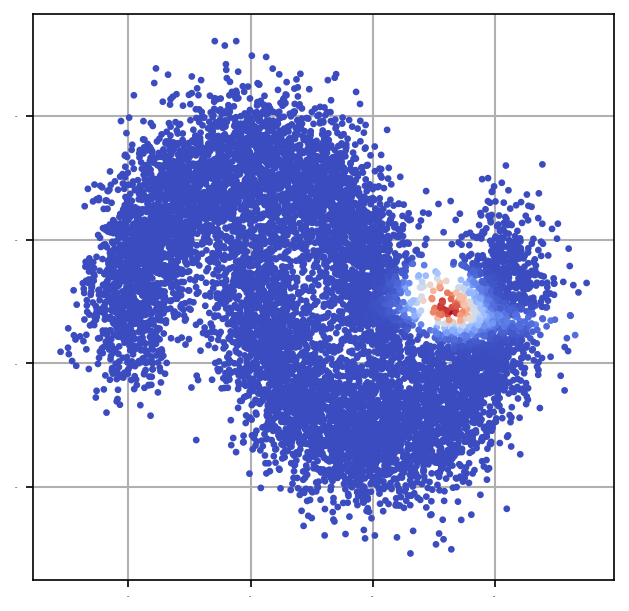}\vspace{-5pt}
  \caption{Sample 1 heatmap}
  \label{img_23a}
\end{subfigure}\hspace{-7pt}
\begin{subfigure}{0.2\textwidth}
  \centering
  \includegraphics[width=.9\linewidth]{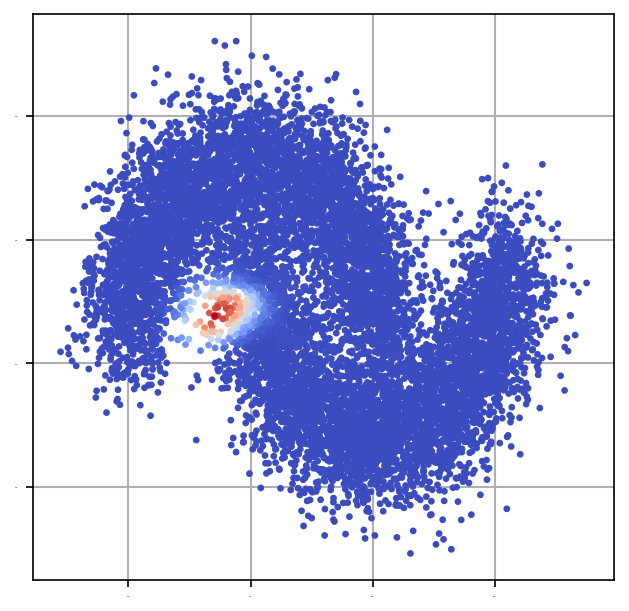}\vspace{-5pt}
  \caption{Sample 2 heatmap}
\end{subfigure}\hspace{-7pt}
\begin{subfigure}{0.2\textwidth}
  \centering
  \includegraphics[width=.9\linewidth]{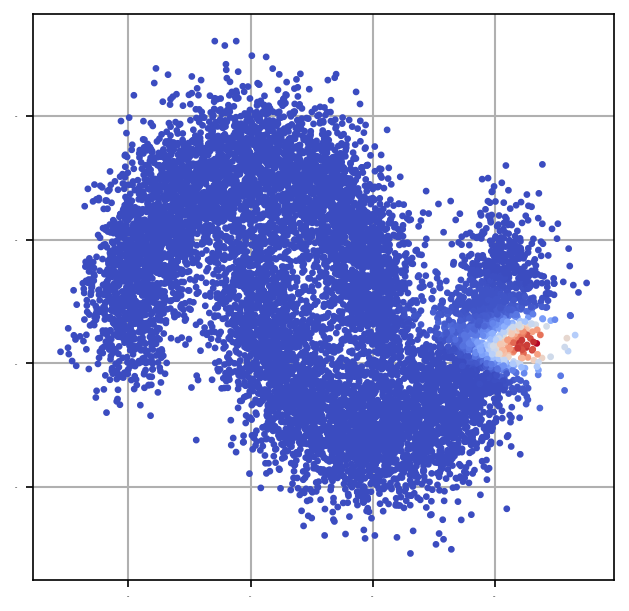}\vspace{-5pt}
  \caption{Sample 3 heatmap}
\end{subfigure}\hspace{-7pt}
\begin{subfigure}{0.2\textwidth}
  \centering
  \includegraphics[width=.9\linewidth]{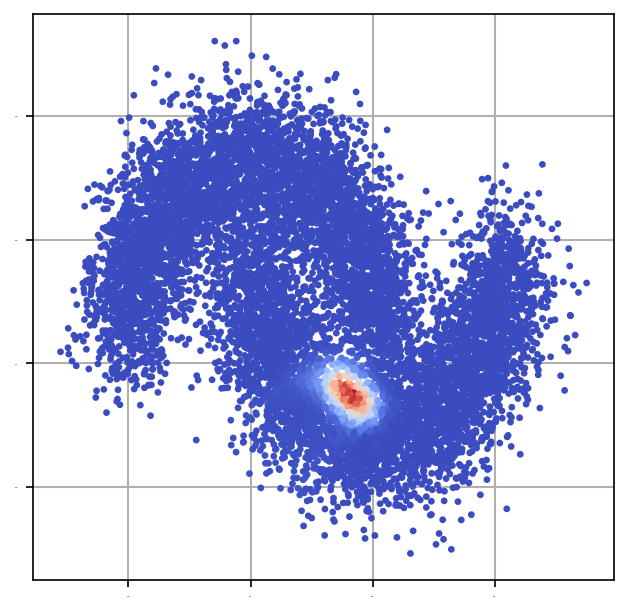}\vspace{-5pt}
  \caption{Sample 4 heatmap}
  \label{img_23d}
\end{subfigure}
\begin{subfigure}{0.24\textwidth}
  \centering
  \includegraphics[width=1\linewidth]{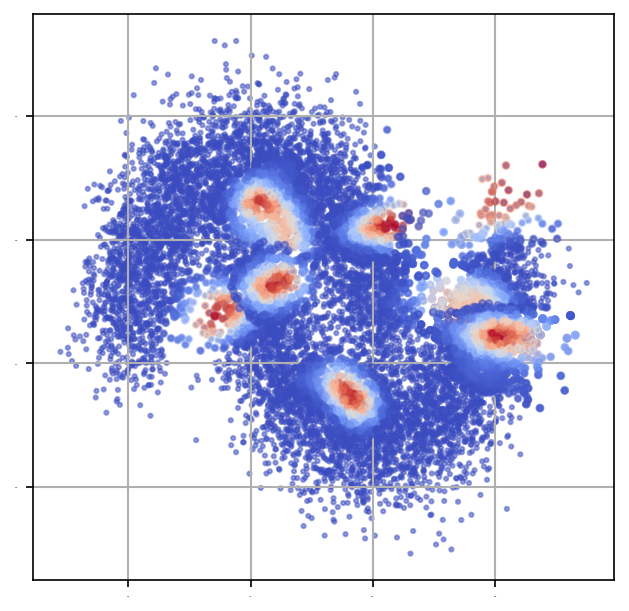}\vspace{-5pt}
  \caption{Visualizing 10 samples}
  \label{img_23e}
\end{subfigure}
\caption{Heatmaps of the learned dependence values obtained by fixing one sample and computing the density ratio between this selected representative sample and all other samples in the dataset. Samples that have positive density ratio values are concentrated around the selected sample.}
\label{figure_dsaoidj}
\end{figure}

\noindent heatmaps in Fig.~\ref{img_23a} to~\ref{img_23d}, and a visualization of ten different fixed samples in Fig.~\ref{img_23e}. 

For each fixed sample, we compute the density ratio values between each representative exemplar with respect to all other \(10000\) samples, thereby producing a heatmap over the entire dataset. The figures show that high-ratio regions are concentrated around the selected sample. In these plots, we visualize $\sum_{k=1}^K f_k(X)g_k(Y)$, where \(X\) is fixed and \(Y\) ranges over all other samples to generate the respective heatmap.

We also visualize how this learned metric changes over the course of autoencoder training. Our earlier theory stated a Gaussian-ball shrinking effect in autoencoder training: as the autoencoder becomes better trained, the reconstruction MSE decreases, which corresponds to a Gaussian ball with a smaller radius. Therefore, as training proceeds, we expect the region of high dependence to shrink. This is particularly natural and easy to show in this concatenated-input-noise setting.

\begin{figure}[H]
\centering
\begin{subfigure}{0.16\textwidth}
  \centering
  \includegraphics[width=1\linewidth]{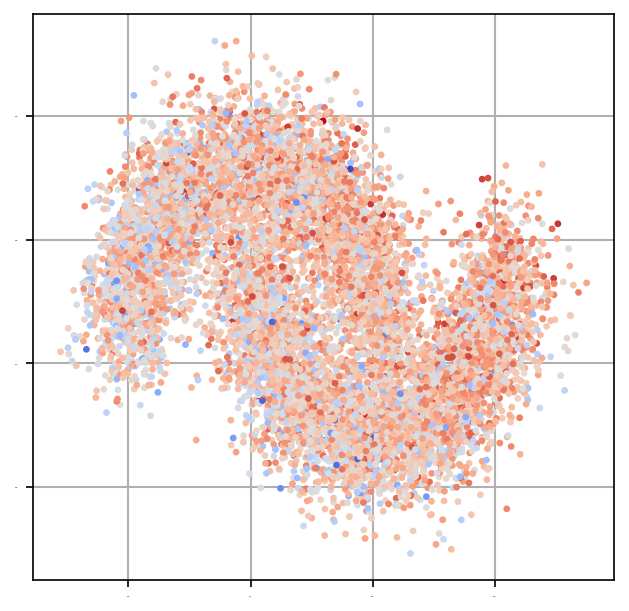}\vspace{-5pt}
  \caption{AE Iter 0}
\end{subfigure}\hspace{-7pt}
\begin{subfigure}{0.16\textwidth}
  \centering
  \includegraphics[width=1\linewidth]{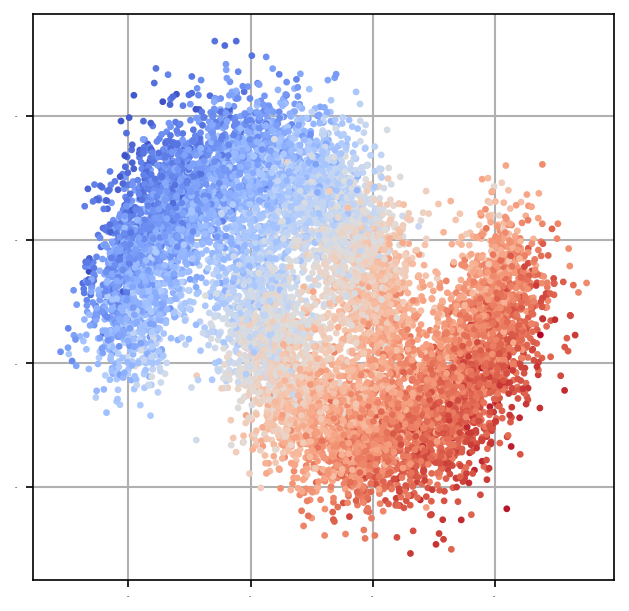}\vspace{-5pt}
  \caption{AE Iter 100}
\end{subfigure}\hspace{-7pt}
\begin{subfigure}{0.16\textwidth}
  \centering
  \includegraphics[width=1\linewidth]{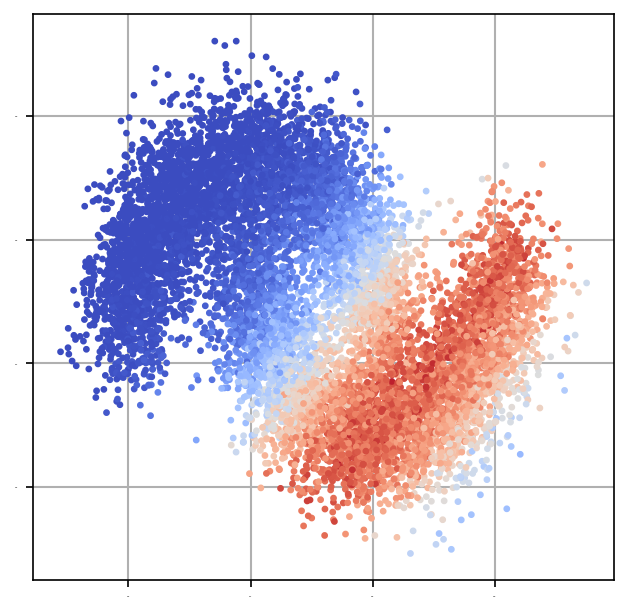}\vspace{-5pt}
  \caption{AE Iter 200}
\end{subfigure}\hspace{-7pt}
\begin{subfigure}{0.16\textwidth}
  \centering
  \includegraphics[width=1\linewidth]{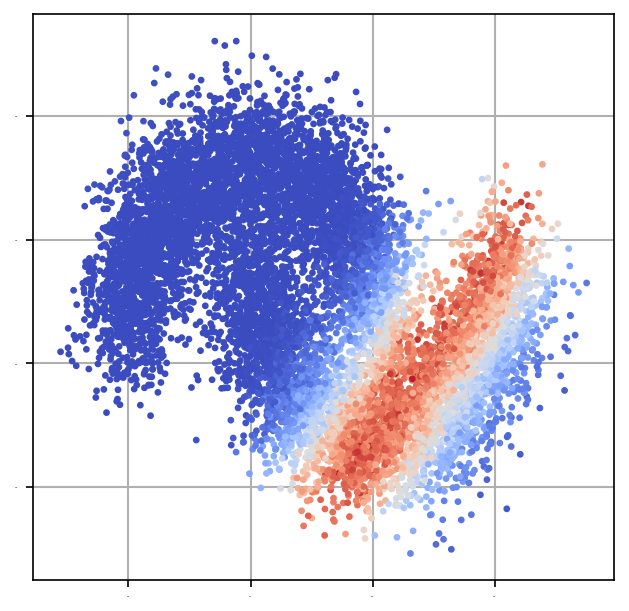}\vspace{-5pt}
  \caption{AE Iter 500}
\end{subfigure}\hspace{-7pt}
\begin{subfigure}{0.16\textwidth}
  \centering
  \includegraphics[width=1\linewidth]{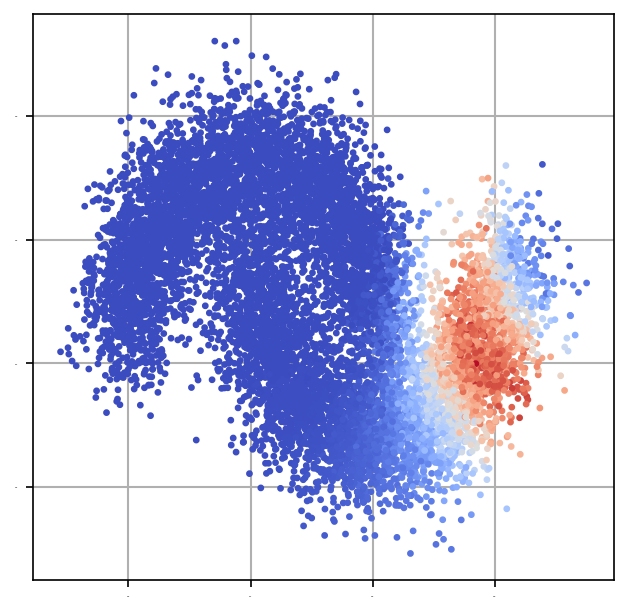}\vspace{-5pt}
  \caption{AE Iter 1000}
\end{subfigure}\hspace{-7pt}
\begin{subfigure}{0.16\textwidth}
  \centering
  \includegraphics[width=1\linewidth]{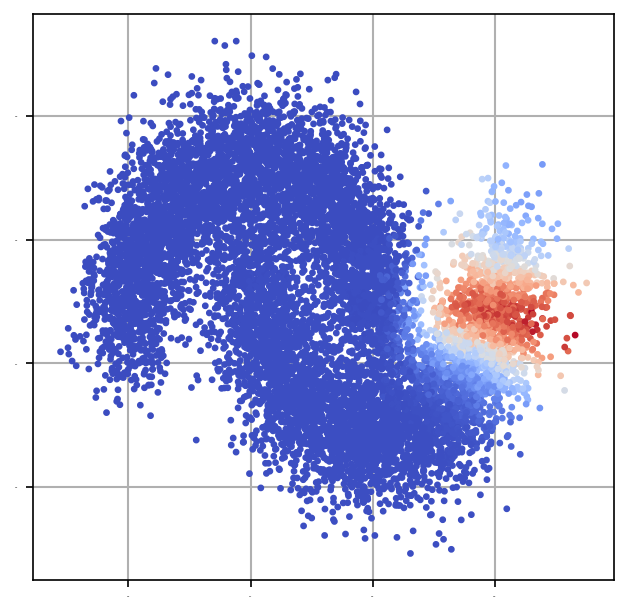}\vspace{-5pt}
  \caption{AE Iter 2000}
\end{subfigure}\hspace{-7pt}
\begin{subfigure}{0.16\textwidth}
  \centering
  \includegraphics[width=1\linewidth]{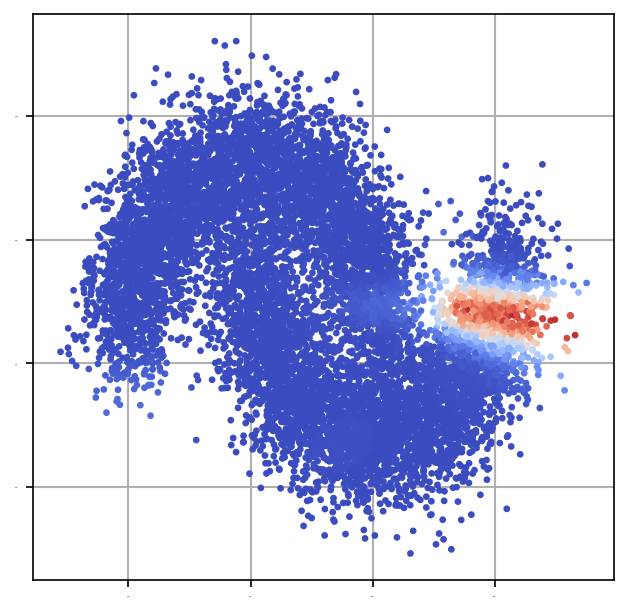}\vspace{-5pt}
  \caption{AE Iter 10000}
\end{subfigure}\hspace{-7pt}
\begin{subfigure}{0.16\textwidth}
  \centering
  \includegraphics[width=1\linewidth]{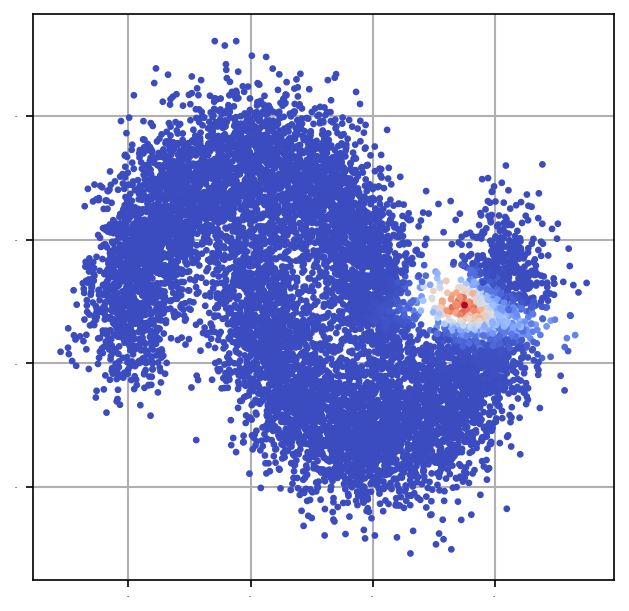}\vspace{-5pt}
  \caption{AE Iter 50000}
\end{subfigure}\hspace{-7pt}
\begin{subfigure}{0.16\textwidth}
  \centering
  \includegraphics[width=1\linewidth]{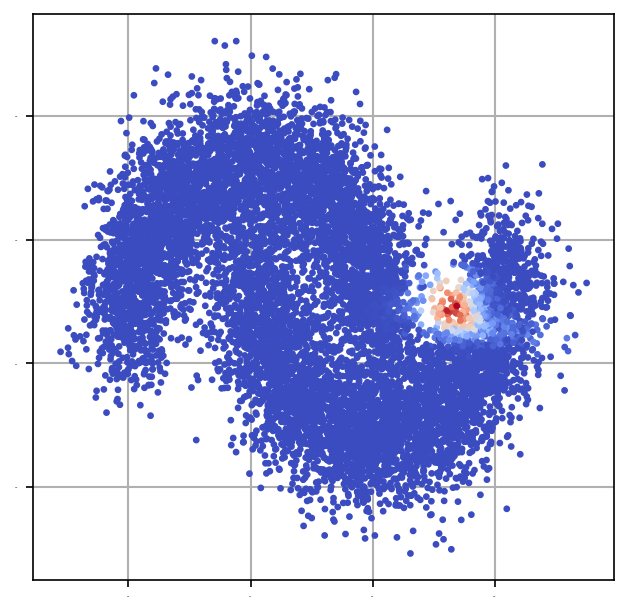}\vspace{-5pt}
  \caption{AE Iter 100000}
\end{subfigure}\hspace{-7pt}
\caption{Evolution of the estimated density ratio between one fixed sample and all other samples across autoencoder training iterations. For each autoencoder checkpoint, a separate dependence estimator is trained. As training proceeds, the high-dependence region with positive density ratio values progressively shrinks, which is consistent with the stated Gaussian-ball shrinking effect. The visualized quantity is \(\sum_{k=1}^{K} f_k(X)g_k(Y)\), where \(X\) is fixed and \(Y\) ranges over all \(10000\) samples.}
\label{figure_4444dasdaslfdkjsdfq}
\end{figure}

To test this, we stop the autoencoder at different training iterations, from \(0\) to \(100000\), train a separate dependence estimator for each checkpoint, fix one sample, and visualize the heatmap of its learned dependence score with all other samples in the dataset. The results are shown in Fig.~\ref{figure_4444dasdaslfdkjsdfq}. They suggest a clear shrinking effect: as the autoencoder becomes more trained, the region containing samples with high dependence with positive density ratio values relative to the fixed sample becomes progressively smaller. This is consistent with the stated Gaussian-ball shrinking behavior.\vspace{5pt}

\subsection{Concatenated input noise on MNIST}

Next, we examine whether the conclusion from the two-moon experiment that concatenated input noise can alleviate overestimation, in addition to the additive feature-noise scheme, also holds for MNIST. To do so, we apply the concatenated-input-noise setting to MNIST. Specifically, before every linear layer in the network, we concatenate the current features at that layer with a 200-dimensional uniform random noise vector, resampled at every training iteration.\vspace{9pt}

To reduce computational cost, we use \(6000\) samples out of the full \(60000\) MNIST training set, keeping one sample out of every ten. This choice is made purely for faster training. At each iteration, we use all \(6000\) samples as a single batch.

\begin{figure}[H]
\centering
\begin{subfigure}{0.23\textwidth}
  \centering
  \includegraphics[width=1\linewidth]{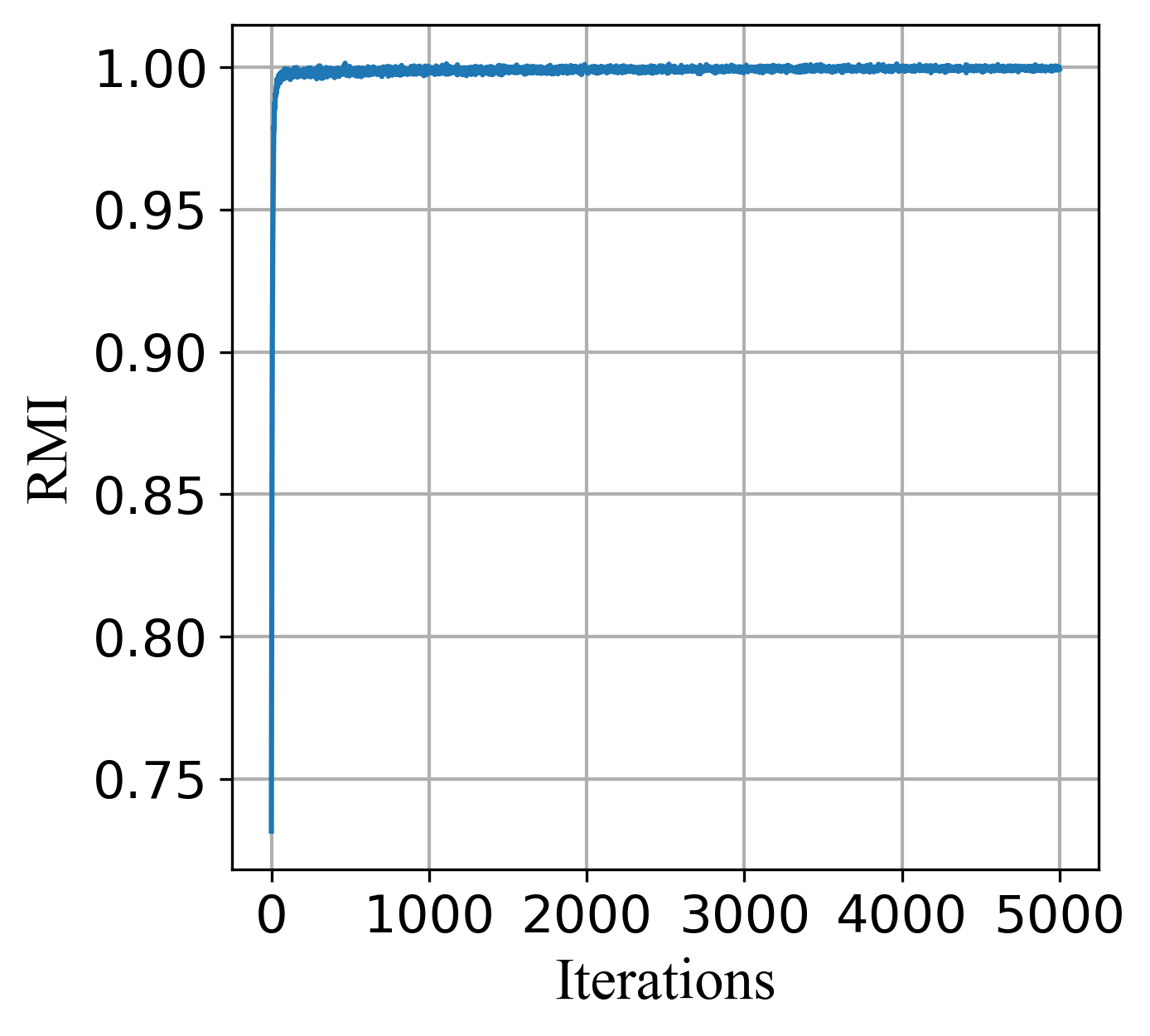}\vspace{-5pt}
  \caption{\textcolor{red}{\footnotesize Noise}: AE Iter 0}
  \label{25a}
\end{subfigure}\hfill
\begin{subfigure}{0.23\textwidth}
  \centering
  \includegraphics[width=1\linewidth]{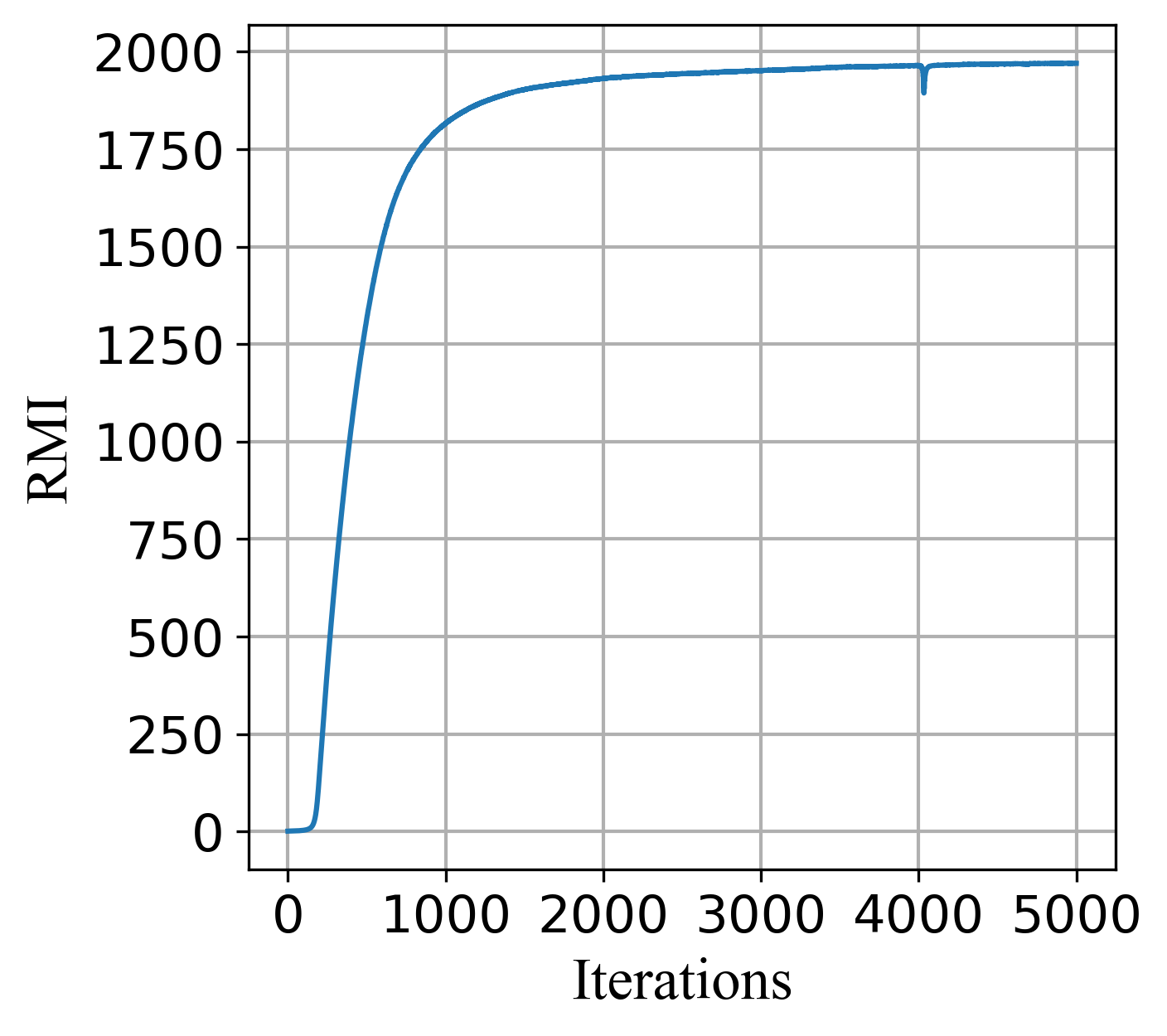}\vspace{-5pt}
  \caption{\textcolor{blue}{Static}: AE iter 0}
  \label{25b}
\end{subfigure}\hspace{-7pt}
\begin{subfigure}{0.23\textwidth}
  \centering
  \includegraphics[width=1\linewidth]{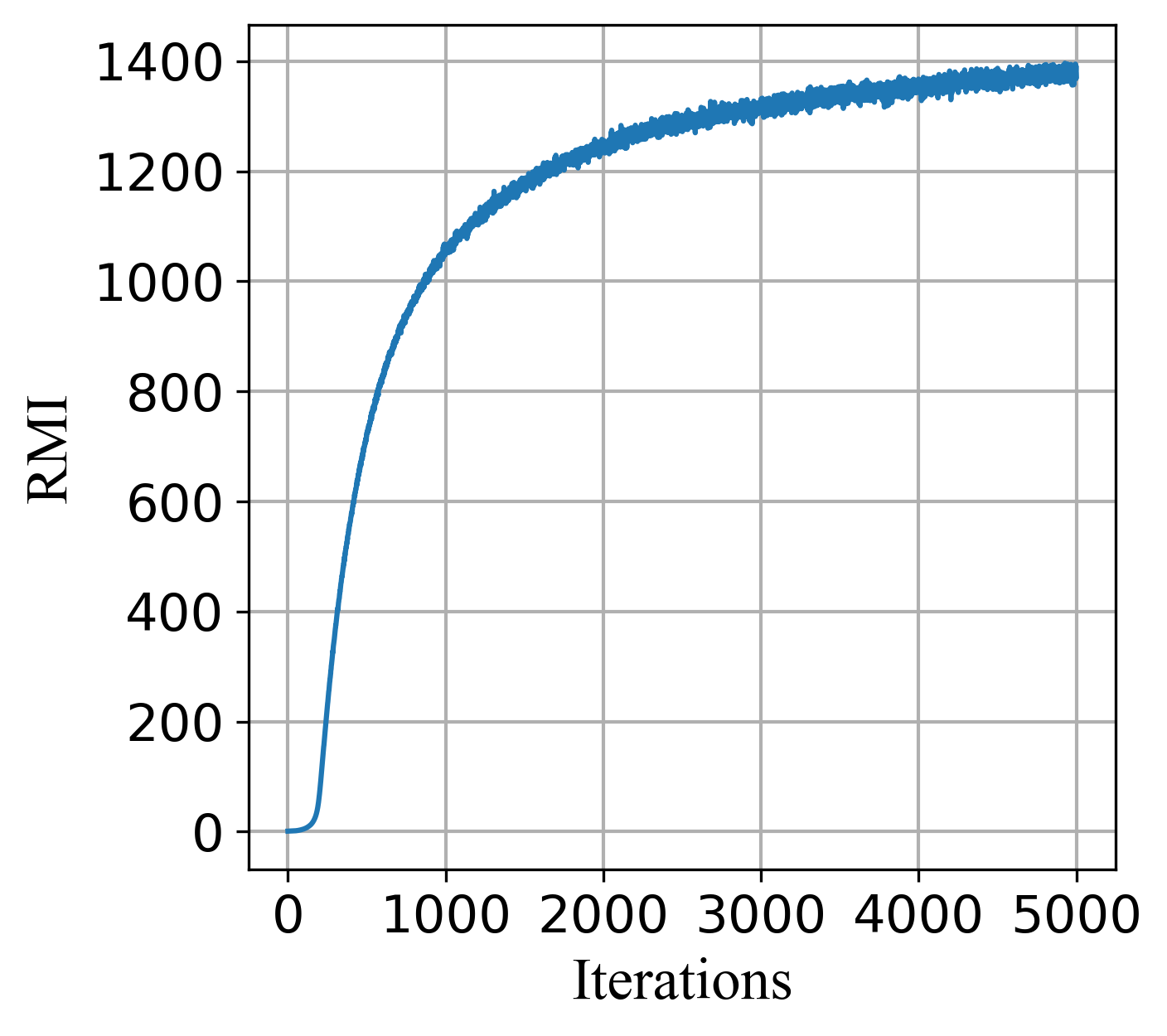}\vspace{-5pt}
  \caption{\textcolor{red}{Noise}: AE iter 1000}
  \label{25c}
\end{subfigure}\hfill
\begin{subfigure}{0.23\textwidth}
  \centering
  \includegraphics[width=1\linewidth]{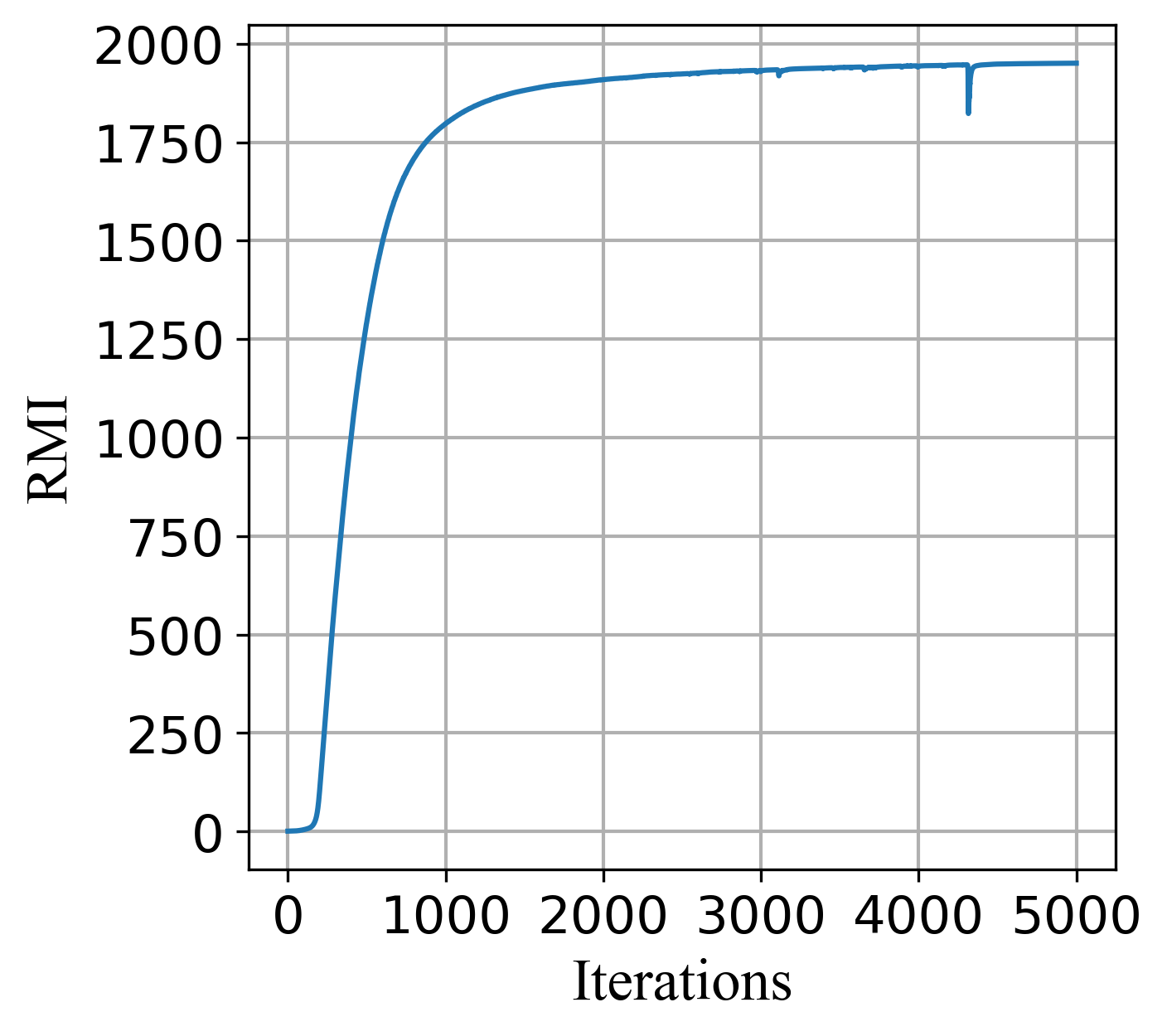}\vspace{-5pt}
  \caption{\textcolor{blue}{Static}: AE iter 1000}
  \label{25d}
\end{subfigure}\hspace{-7pt}
\begin{subfigure}{0.23\textwidth}
  \centering
  \includegraphics[width=1\linewidth]{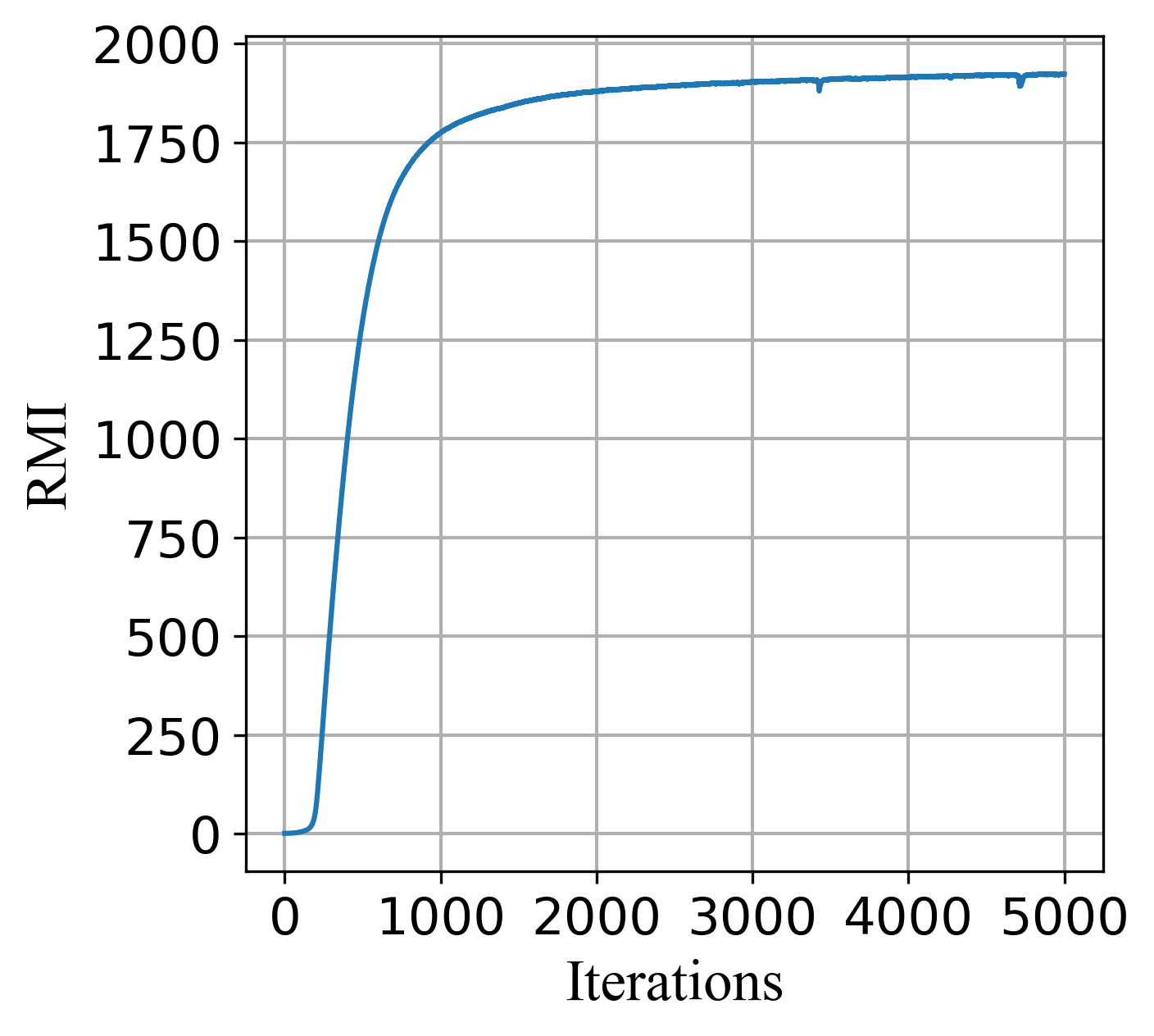}\vspace{-5pt}
  \caption{\textcolor{red}{Noise}: AE iter 10000}
  \label{25e}
\end{subfigure}\hfill
\begin{subfigure}{0.23\textwidth}
  \centering
  \includegraphics[width=1\linewidth]{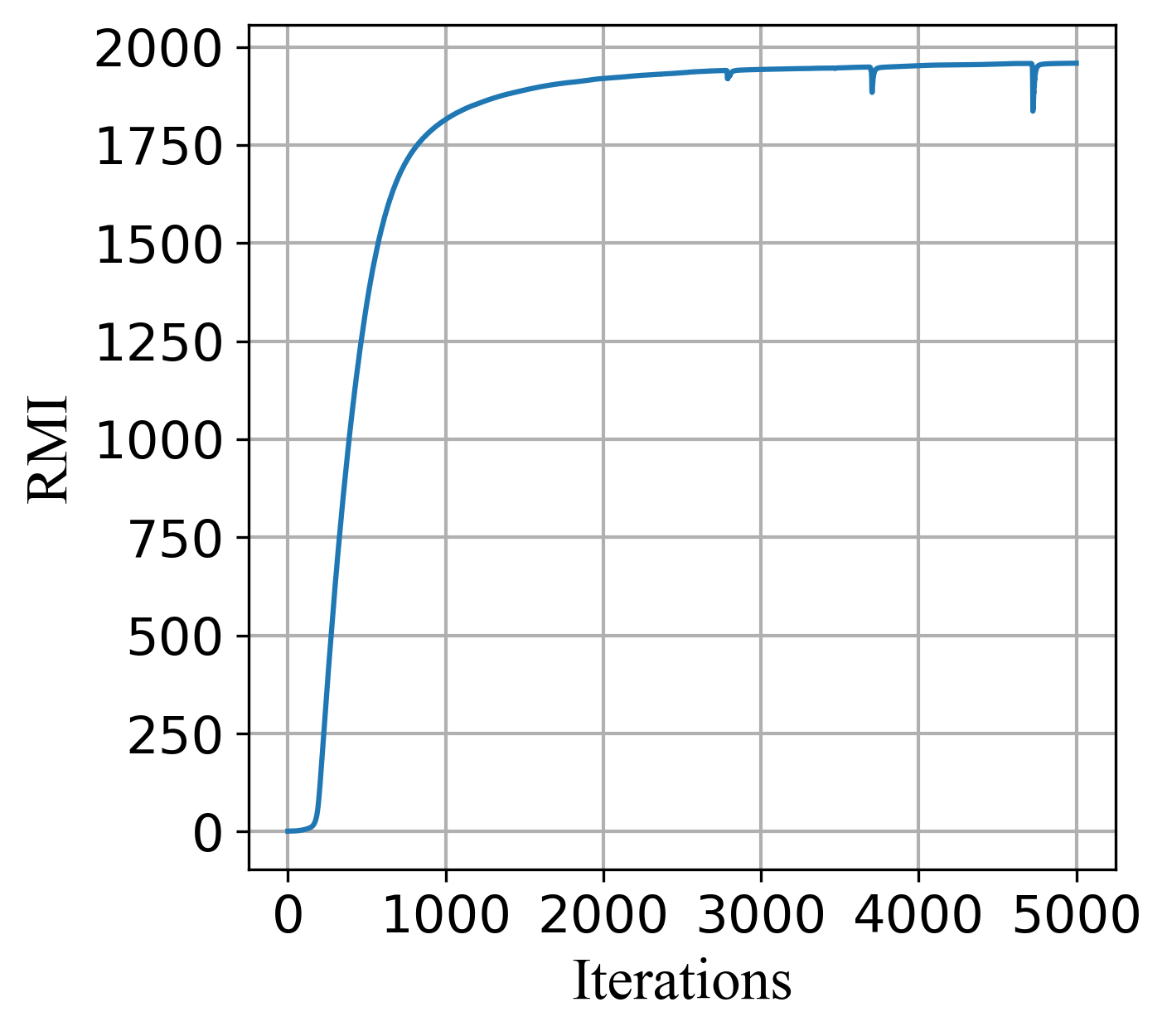}\vspace{-5pt}
  \caption{\textcolor{blue}{Static}: AE iter 10000}
  \label{25f}
\end{subfigure}\hspace{-7pt}
\caption{MNIST: learning curves of the dependence estimator for the concatenated-input-noise setting and the static-network setting, evaluated at different autoencoder checkpoints. When the autoencoder is untrained (iteration 0), concatenated input noise keeps the estimate near its lower bound of \(1\), whereas the static setting converges to a much larger value, close to \(2000\), which matches the output dimension of the estimator. At later checkpoints, concatenated input noise still reduces the estimate during the early stage of autoencoder training (e.g., iteration \(1000\)), but its effect becomes limited by iteration \(10000\), where both settings again converge to large values. In these experiments, the feature dimension is \(2\).\vspace{15pt}}
\label{figure25}
\end{figure}

We first visualize the learning curves for MNIST in Fig.~\ref{figure25}. As in the two-moon experiment, when the autoencoder is untrained, the estimated value remains at \(1\), which is the minimum value of the dependence measure in our setup. This is shown in Fig.~\ref{25a}. Without concatenated input noise, the estimate converges to a very large value, around \(2000\), as shown in Fig.~\ref{25b}. This is consistent with the behavior observed on the two-moon dataset, where we would ideally like the initial value to match the lower bound corresponding to an untrained autoencoder.

When concatenated input noise is used, the estimate converges to approximately \(1400\) when the autoencoder is taken at iteration \(1000\) (Fig.~\ref{25c}, and to approximately \(1900\) when the autoencoder is taken at iteration \(10000\) (Fig.~\ref{25e}). 

When concatenated input noise is not used, the estimate converges to about \(1900\) for both these checkpoints (Fig.~\ref{25d} and Fig.~\ref{25f}). 

These results indicate that, on MNIST, concatenated input noise is helpful in the early stage of training and can reduce overestimation to some extent. However, at later stages of training, the estimate still converges to a very large value, around \(1900\). Therefore, while the effect of concatenated input noise is useful, it is limited on MNIST and is not by itself sufficient to guarantee that the estimator is unbiased.\vspace{9pt}

\begin{figure}
\centering
\hspace{-20pt}\begin{subfigure}{0.45\textwidth}
  \centering
  \includegraphics[width=.7\linewidth]{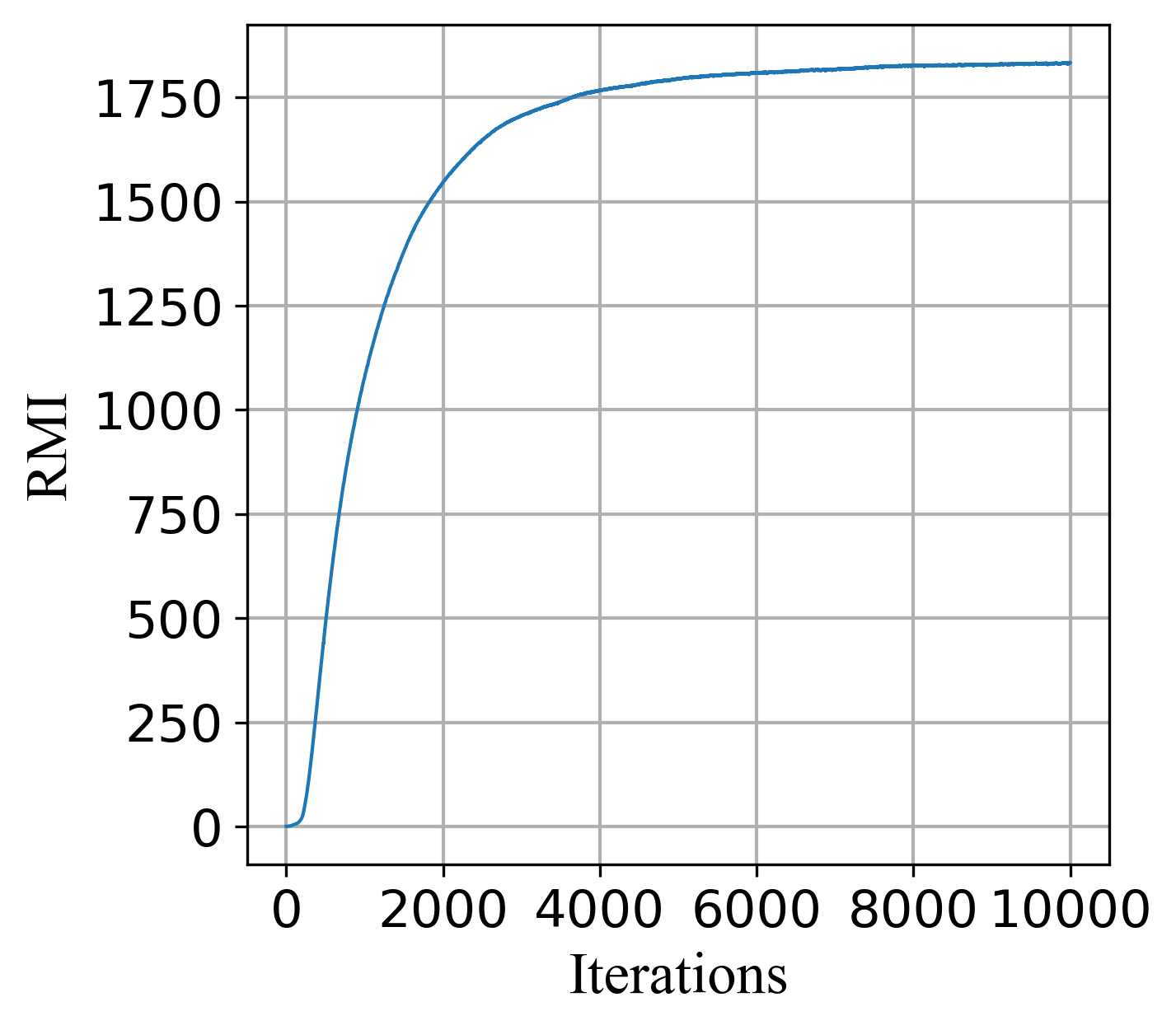}\vspace{-6pt}
  \caption{MNIST with concatenated input noise: learning curve of the dependence estimator under full-batch optimization, using all \(60000\) training samples in each update.}
  \label{learningcurvea}
\end{subfigure}
\begin{subfigure}{0.45\textwidth}
  \centering
  \includegraphics[width=.8\linewidth]{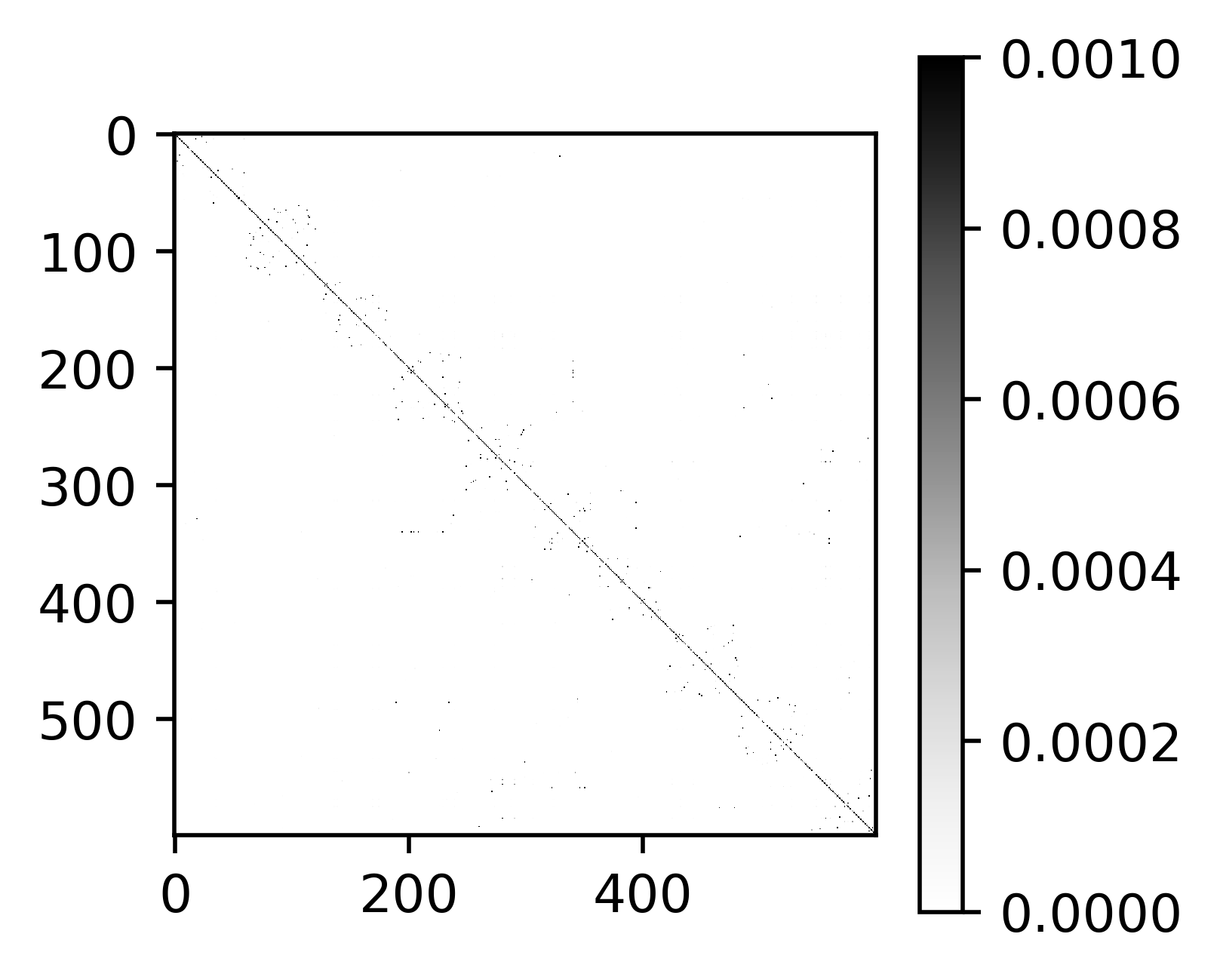}\vspace{-5pt}
  \caption{Estimated density-ratio metric on \(600\) selected samples, represented by a $600\times 600$ matrix. The matrix is strongly diagonal and highly sparse.}
  \label{diagonal_matrix_whole}
\end{subfigure}\vspace{5pt}
\begin{subfigure}{0.45\textwidth}
  \centering
  \includegraphics[width=.85\linewidth]{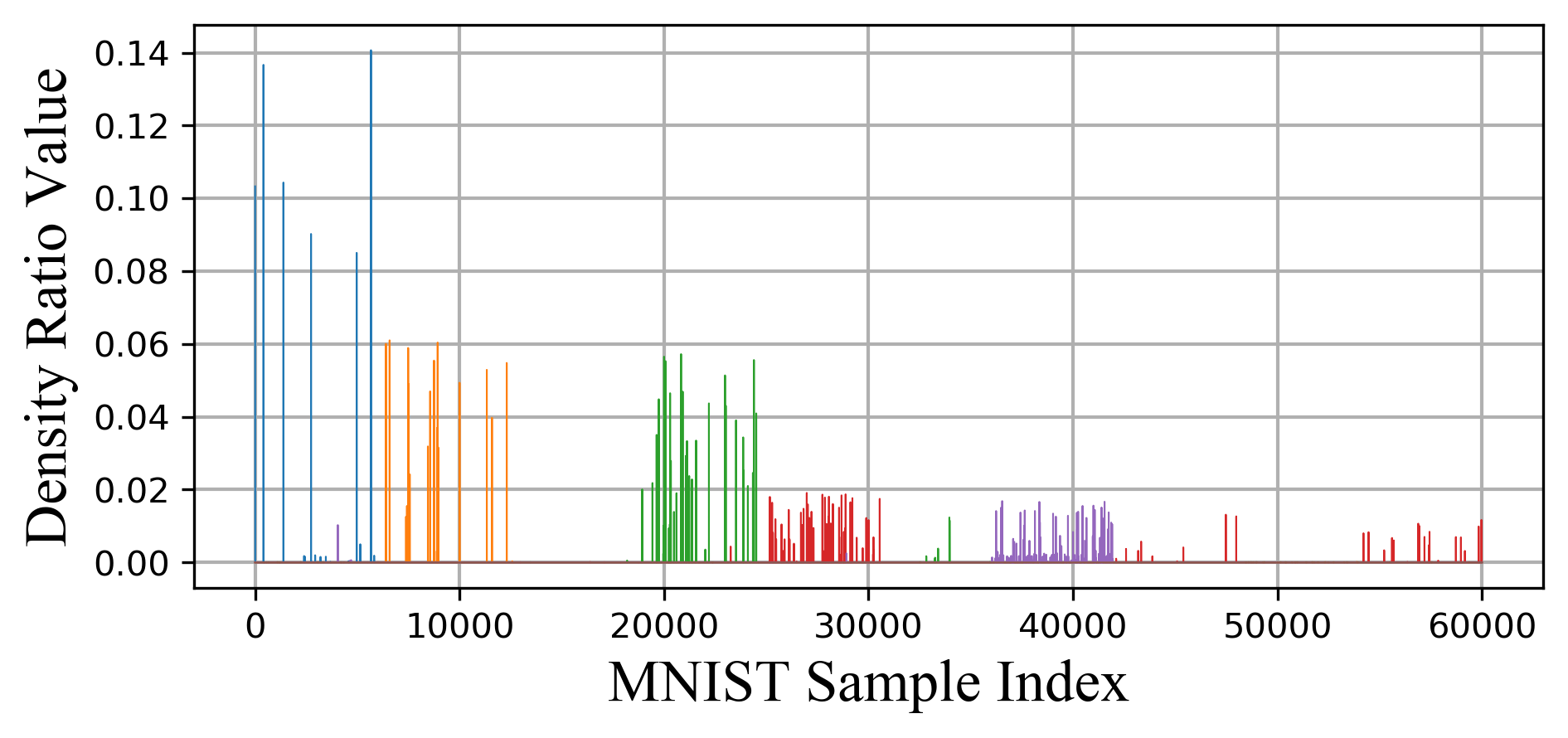}
\caption{Estimated density-ratio responses for $6$ query samples against all \(60000\) training samples, essentially $6$ different rows of the matrix in the previous Fig.~\ref{diagonal_matrix_whole}. The \(x\)-axis is ordered by class labels from \(0\) to \(9\). Positive responses are very sparse and occur predominantly within the same class as the query sample.}
  \label{learningcurvec}
\end{subfigure}
\caption{MNIST results under full-batch optimization: the learning curve, the learned density-ratio matrix, and per-sample responses.}
\label{results_deterministic}
\end{figure}

We next visualize the learned metric, i.e., the estimated density-ratio score. After training the autoencoder, we apply the dependence estimator corresponding to the learning curve in Fig.~\ref{learningcurvea}. We then evaluate the learned score on \(600\) selected samples from the training set, yielding a \(600 \times 600\) matrix. In this experiment, all \(60000\) MNIST training samples are used in a full-batch setting.

As expected, the learned metric is highly concentrated near the diagonal, as shown in Fig.~\ref{diagonal_matrix_whole}. Because this matrix is difficult to inspect visually, we provide an alternative view in Fig.~\ref{learningcurvec}. There, we fix six representative samples and plot their learned responses against all \(60000\) training samples. This should be distinguished from Fig.~\ref{figure20c} and Fig.~\ref{figure21c} where the plotted quantities are the outputs of the estimator networks themselves. Here, instead, we plot the learned density ratio value between one sample and all other samples. Each color corresponds to a different representative sample.

The resulting density ratio responses are highly sparse: each sample has positive responses with only a relatively small subset of the dataset, and these samples mostly belong to the same class as the query sample. This may indicate that the learned metric is overly concentrated and diagonal, although it may also suggest that each point in feature space only requires a small local neighborhood of supporting samples, in a way similar to \(k\)-nearest neighbors.

\begin{figure}[H]
\vspace{40pt}
\centering
\hspace{-15pt}\begin{subfigure}{0.45\textwidth}
  \centering
  \includegraphics[width=.7\linewidth]{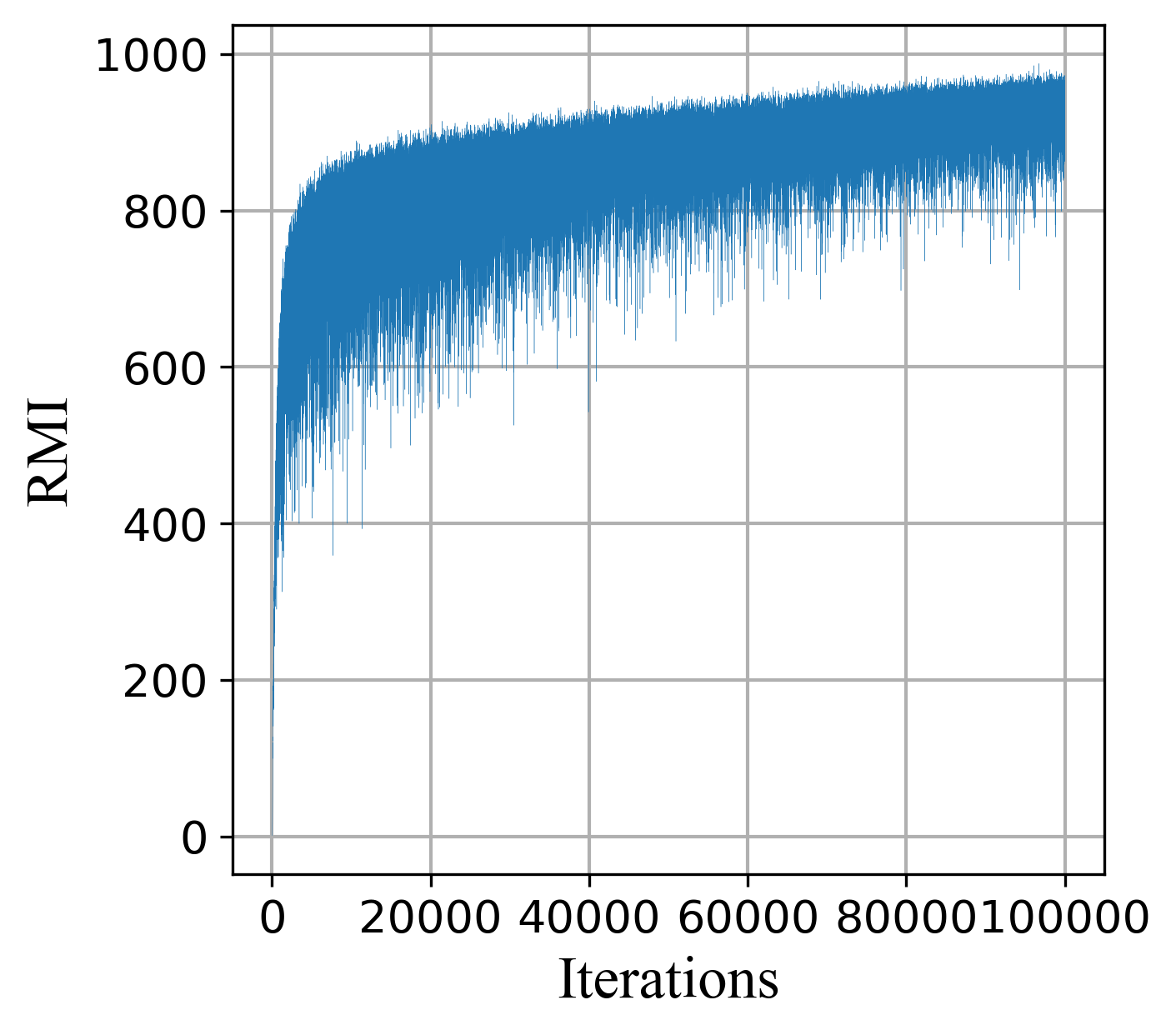}\vspace{-6pt}
  \caption{MNIST with concatenated input noise: learning curve under stochastic minibatch optimization with batch size \(5000\). Compared with the full-batch case in Fig.~\ref{learningcurvea}, the estimated dependence is reduced to around \(1000\).}
  \label{figure27a}
\end{subfigure}
\begin{subfigure}{0.45\textwidth}
  \centering
  \includegraphics[width=.8\linewidth]{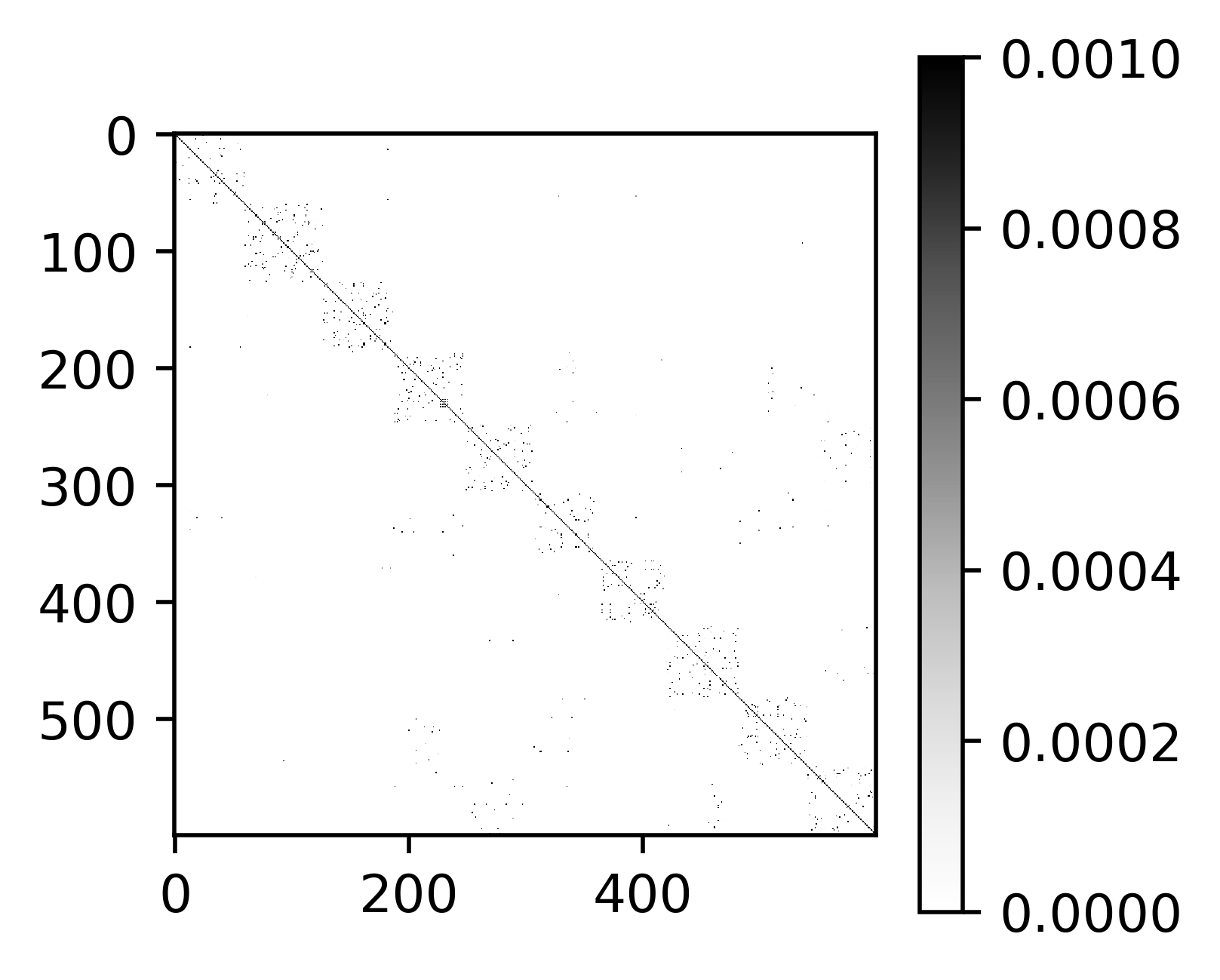}\vspace{-5pt}
  \caption{Estimated density-ratio score matrix under stochastic minibatch optimization. Compared with Fig.~\ref{diagonal_matrix_whole}, the matrix is less diagonal and contains more nonzero structure, although it remains sparse.}
  \label{figure27b}
\end{subfigure}\vspace{5pt}
\begin{subfigure}{0.45\textwidth}
  \centering
  \includegraphics[width=.85\linewidth]{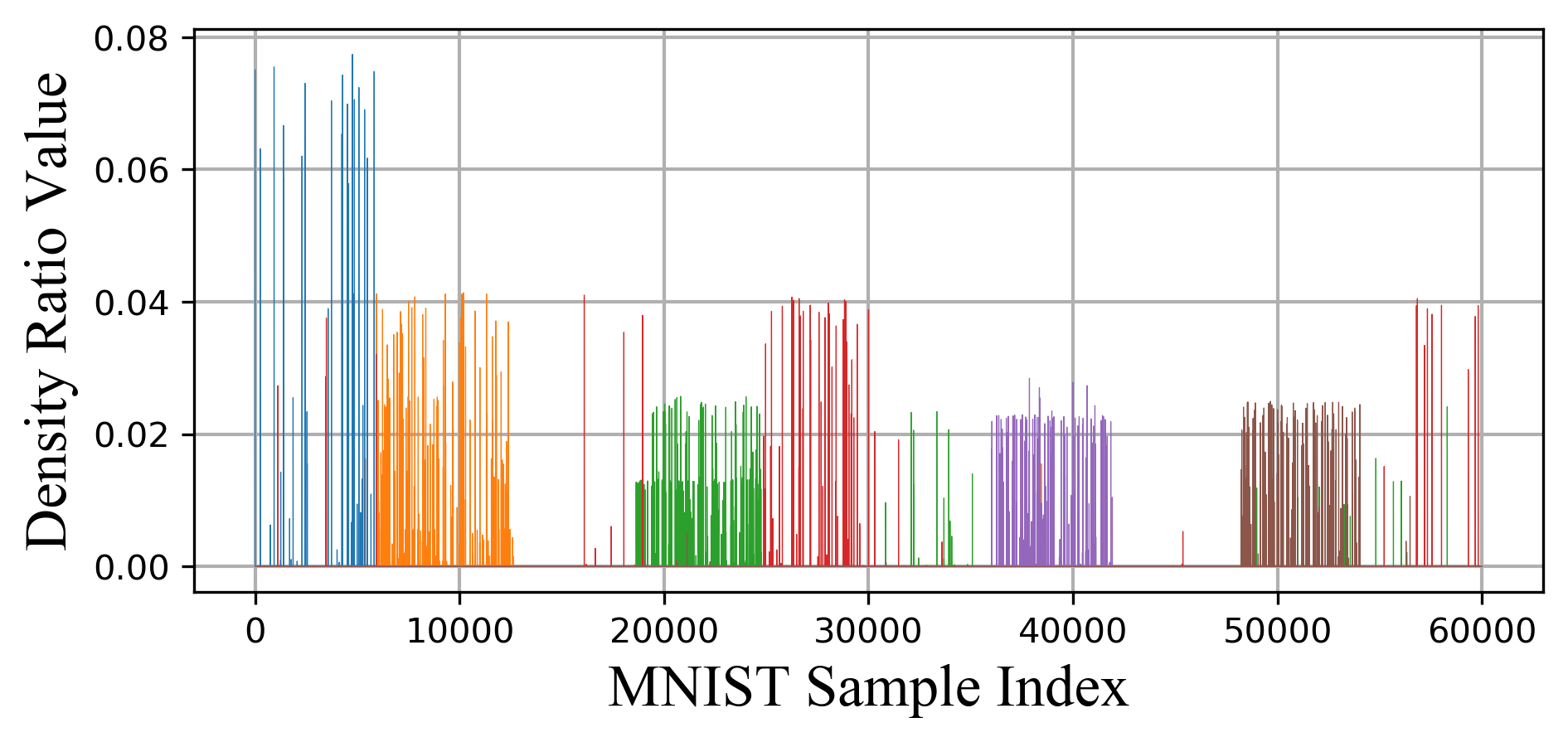}
  \caption{Per-sample density-ratio responses for $6$ query samples corresponding to the matrix in Fig.~\ref{figure27b}.}
  \label{figure27c}
\end{subfigure}
\caption{MNIST results under stochastic minibatch optimization. Compared with the full-batch setting in Fig.~\ref{results_deterministic}, the estimated dependence is reduced and the learned metric becomes much less sparse, although positive responses are still concentrated on relatively few samples.}
\label{figure27}
\end{figure}

We also observe that the stochastic minibatch optimization

\noindent may alleviate the overestimation problem. For example, in the 2D independent uniform case with \(10000\) samples, full-batch optimization still produces substantial overestimation. When we randomly sample minibatches of \(500\) points at each iteration and estimate the dependence on the changing minibatch, the estimated value stays much closer to \(1\). Motivated by this finding, we apply the same scheme to MNIST, with results shown in Fig.~\ref{figure27}. In this experiment, both the autoencoder and the dependence estimator are trained with a batch size of \(5000\), while the total number of samples remains \(60000\).

\newpage

\newpage

\newpage 

Compared with the full-batch learning curve in Fig.~\ref{learningcurvea}, the estimated dependence in Fig.~\ref{figure27a} is reduced to around \(1000\), approximately half of the value obtained in the full-batch case. The learned metric is also less diagonal and less sparse (Fig.~\ref{figure27b} and Fig.~\ref{figure27c}), although the positive responses are still concentrated on only a small subset of samples.\vspace{9pt}

\begin{figure}[H]
\centering
\begin{subfigure}{0.23\textwidth}
  \centering
  \includegraphics[width=1\linewidth]{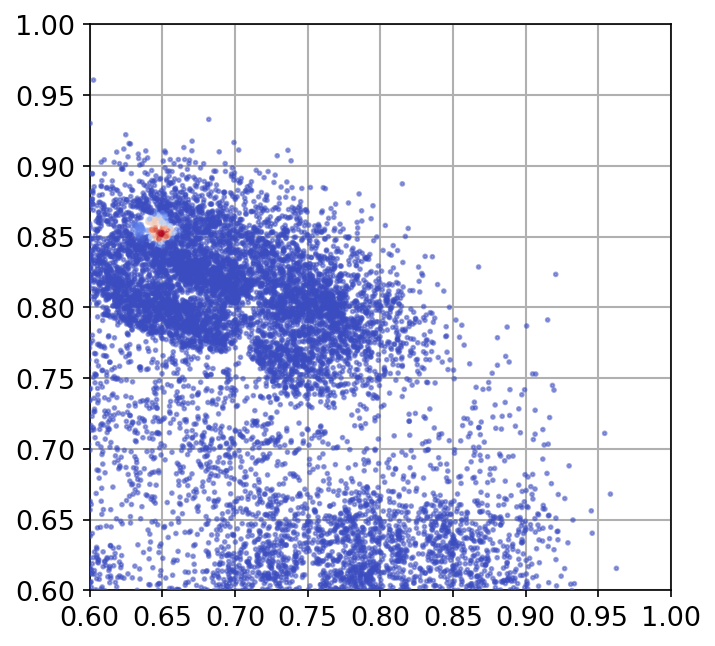}
  \caption{Sample 1 heatmap}
\end{subfigure}\hspace{-7pt}
\begin{subfigure}{0.23\textwidth}
  \centering
  \includegraphics[width=1\linewidth]{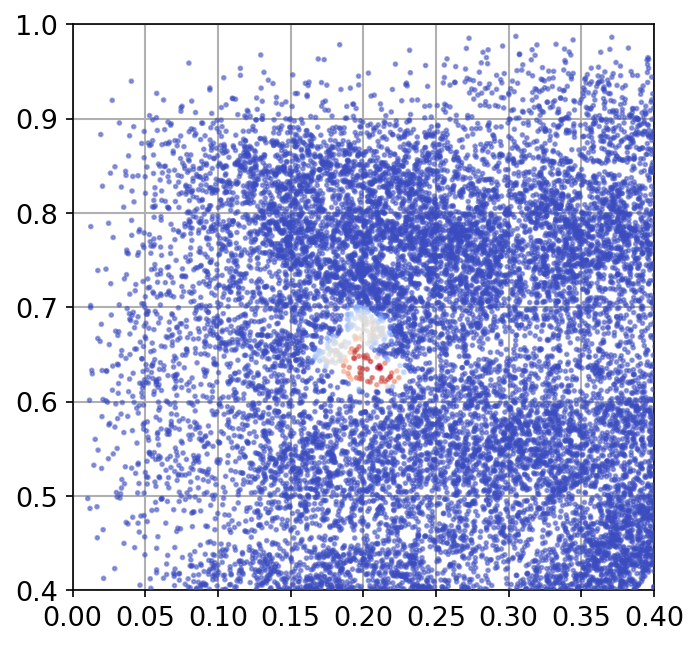}
  \caption{Sample 2 heatmap}
\end{subfigure}\hspace{-7pt}
\begin{subfigure}{0.23\textwidth}
  \centering
  \includegraphics[width=1\linewidth]{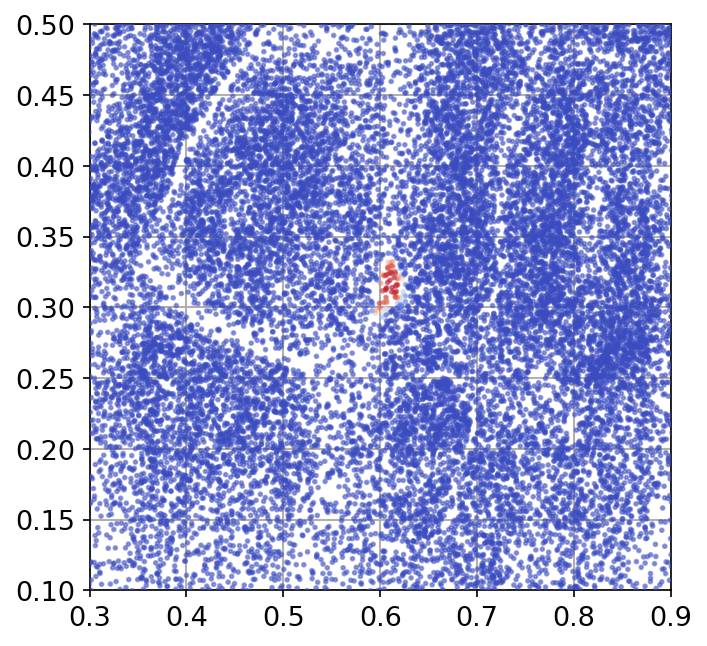}
  \caption{Sample 3 heatmap}
\end{subfigure}\hspace{-7pt}
\begin{subfigure}{0.23\textwidth}
  \centering
  \includegraphics[width=1\linewidth]{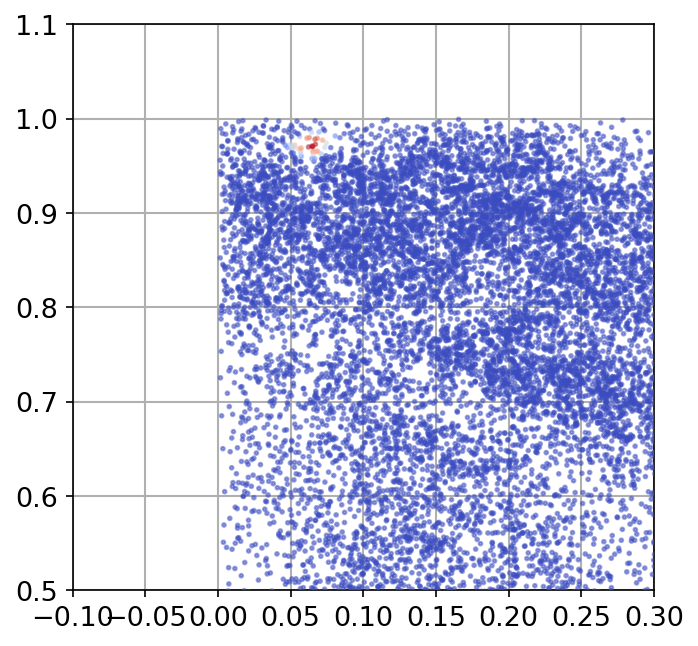}
  \caption{Sample 4 heatmap}
\end{subfigure}\hspace{-7pt}
\caption{Heatmap visualization of the learned density-ratio metric on MNIST in the 2D latent space. Each panel fixes one sample and shows its response to the remaining samples. The positive-response region is small and localized, reflecting the sparsity of the learned density-ratio metric. These results correspond to the stochastic minibatch setting in Fig.~\ref{figure27b}.}
\label{figure_activation}
\end{figure}

Similar to the toy datasets, we visualize the learned density-ratio metric on a sample-by-sample basis. In the toy example, Fig.~\ref{21bbbb} shows the full matrix representation of the density ratio, and Fig.~\ref{figure_dsaoidj} shows, for each fixed sample, which only surrounding points have positive responses. For MNIST, direct visualization in the original \(784\)-dimensional input space is impractical. Thus we first project the samples into the 2D latent space learned by the encoder, and then visualize the learned metric as heatmaps in the 2D space, as shown in Fig.~\ref{figure_activation}.

Each panel fixes one sample and shows its learned response against the remaining samples. To clearly display the positive-response region, the plots are zoomed in around the area of interest. Since the learned metric for MNIST is much sparser than in the toy examples, these positive-response regions are quite small. Based on the toy-data experiments, we would also expect these regions to shrink as autoencoder training progresses. The figures shown here correspond to the stochastic minibatch setting.

\newpage

\begin{figure}[H]
\centering
\begin{subfigure}{0.41\textwidth}
  \centering
  \includegraphics[width=.6\linewidth]{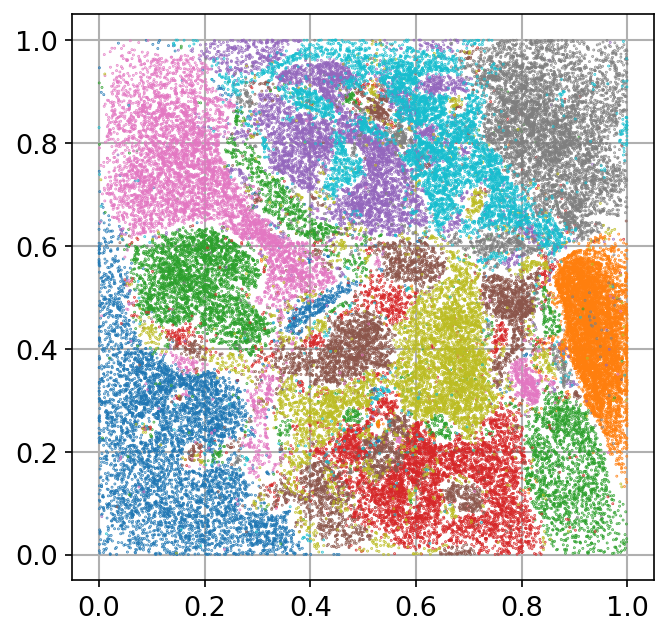}\vspace{-5pt}
  \caption{2D feature projection, colored by class label.}
  \label{figure29aaa}
\end{subfigure}\\ \vspace{7pt}
\begin{subfigure}{0.24\textwidth}
  \centering
  \includegraphics[width=1\linewidth]{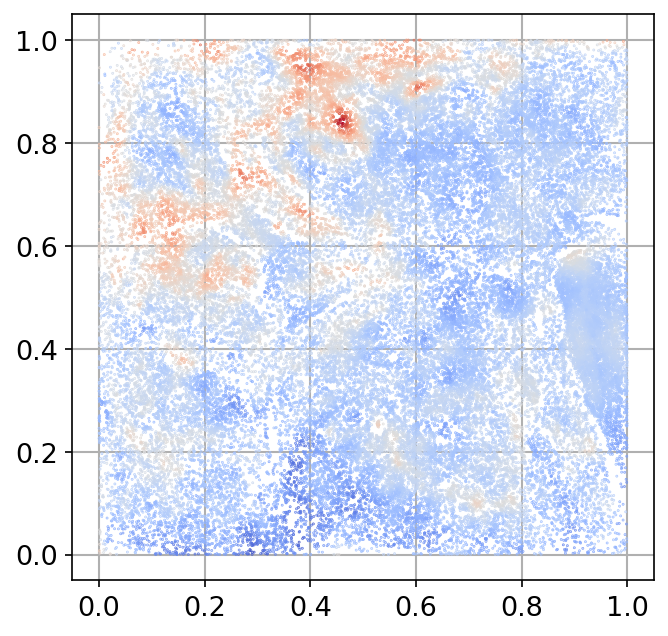}\vspace{-5pt}
  \caption{Sample 1 heatmap}
\end{subfigure}\hspace{-3pt} 
\begin{subfigure}{0.24\textwidth}
  \centering
  \includegraphics[width=1\linewidth]{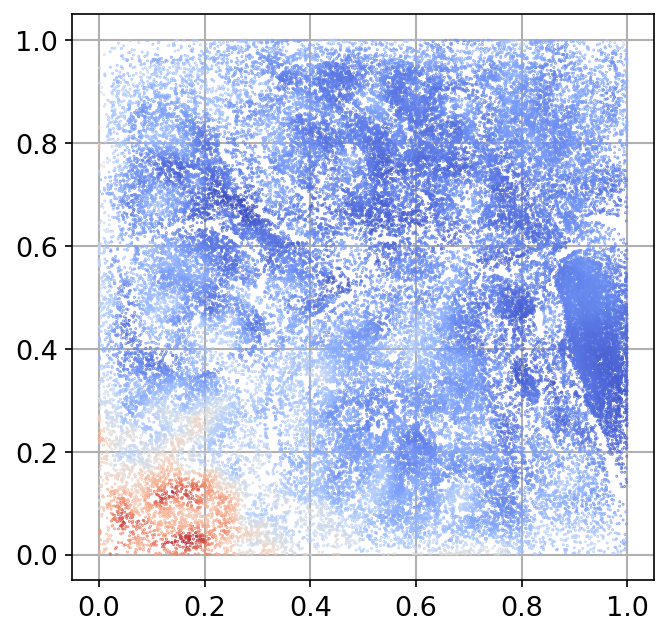}\vspace{-5pt}
  \caption{Sample 2 heatmap}
\end{subfigure}\vspace{5pt}
\begin{subfigure}{0.24\textwidth}
  \centering
  \includegraphics[width=1\linewidth]{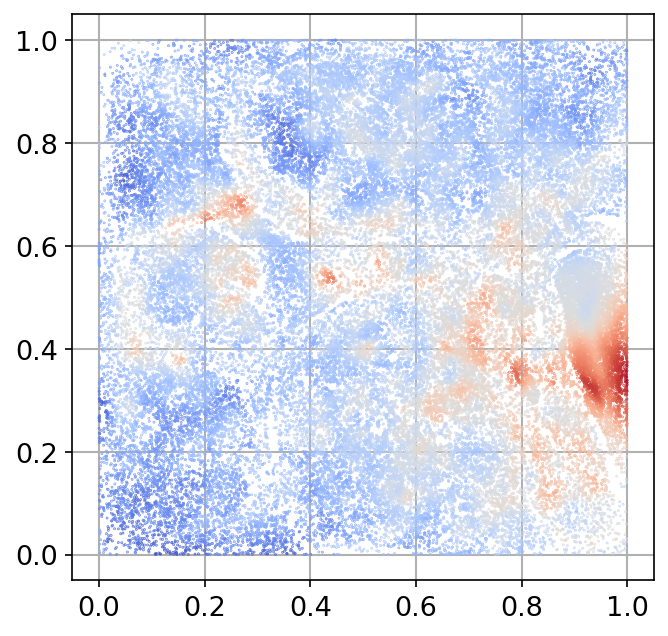}\vspace{-5pt}
  \caption{Sample 3 heatmap}
\end{subfigure}\hspace{-3pt}
\begin{subfigure}{0.24\textwidth}
  \centering
  \includegraphics[width=1\linewidth]{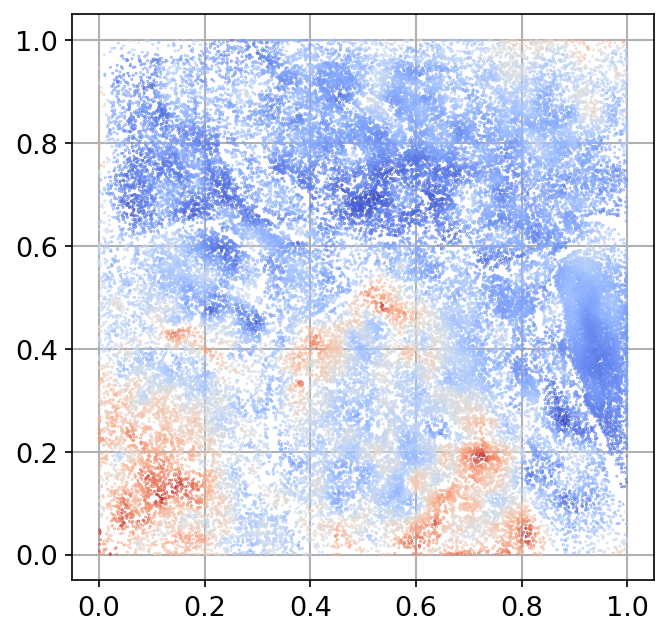}\vspace{-5pt}
  \caption{Sample 4 heatmap}
\end{subfigure}\vspace{1pt}
\caption{Heatmaps generated in the same way as in Fig.~\ref{figure_activation}, but for the setting in Section~\ref{section_emperical_estimation} where only the encoder, without the decoder, is trained using the NMF-like cost with additive Gaussian noise on the input. In this case, more samples receive positive responses relative to the query point, even though the 2D projected features in panel (a) remain visually similar to those obtained in the autoencoder-based setting.}
\label{FIGURE29}
\end{figure}

\begin{figure}[H]
\centering
\hspace{-20pt}
\begin{subfigure}{0.35\textwidth}
  \centering
  \includegraphics[width=.9\linewidth]{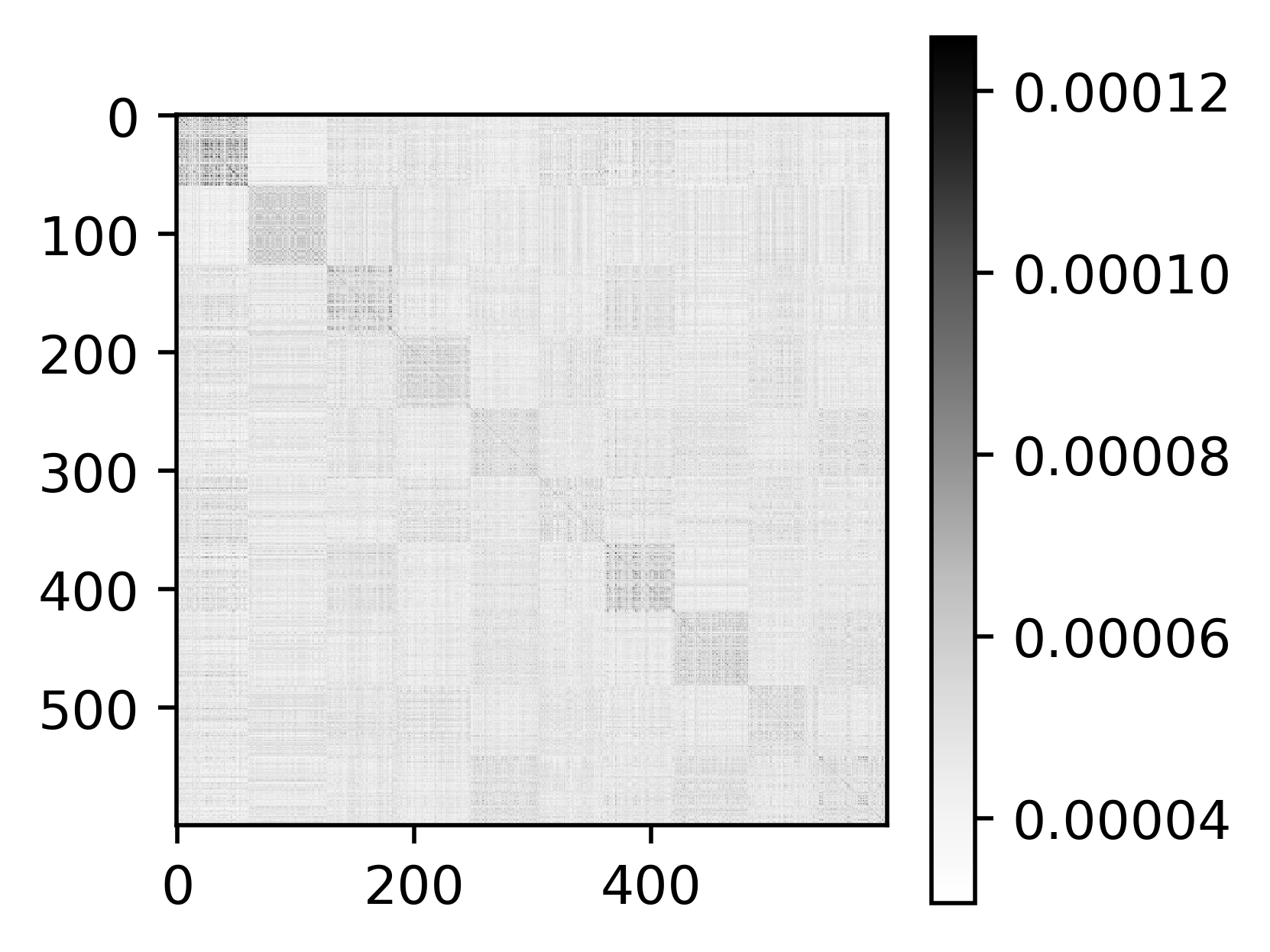}\vspace{-5pt}
  \caption{Estimated density-ratio metric, represented by a $600\times 600$ matrix.}
  \label{figure30a}
\end{subfigure}\vspace{10pt}
\begin{subfigure}{0.44\textwidth}
  \centering
  \includegraphics[width=.9\linewidth]{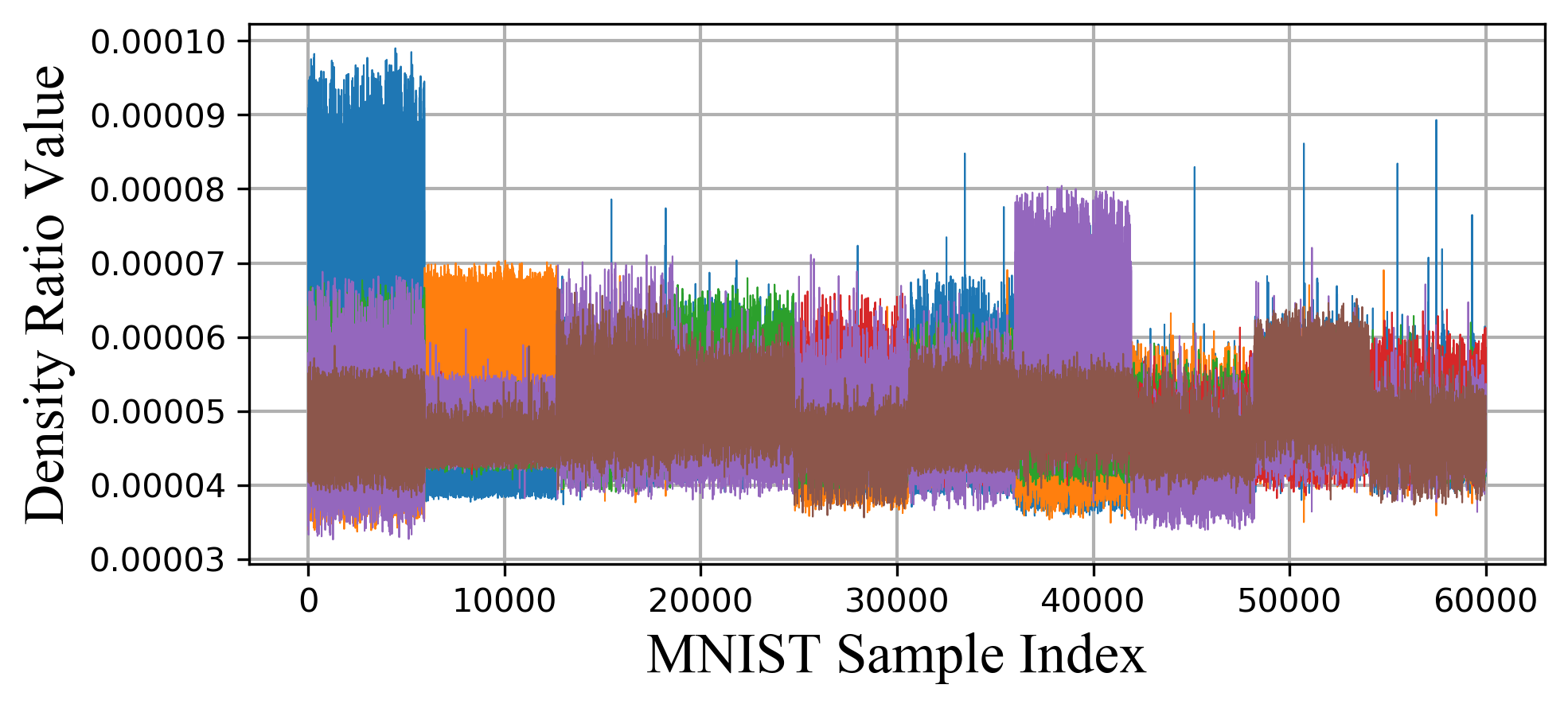}
    \caption{Per-sample responses corresponding to Fig.~\ref{figure30a}.}
  \label{figure30b}
\end{subfigure}\vspace{5pt}
\caption{Further visualizing the case of the NMF-like scalar cost with additive Gaussian noise on the input for optimizing the encoder. Compared with Fig.~\ref{results_deterministic} and Fig.~\ref{figure27}, which correspond to standard autoencoder training, the learned metric here is substantially less sparse, even though the projected features shown in Fig.~\ref{figure29aaa} appear visually similar to training an end-to-end autoencoder.}
\label{figure30}
\end{figure}

Recall from Section~\ref{section_emperical_estimation} that, in the dependence-maximization setting with additive Gaussian noise on the input and the NMF-like ratio objective, the learned metric becomes much less sparse and the number of significant singular values is substantially reduced. Visualizing the metric in that setting reveals a noticeably different behavior, shown in Fig.~\ref{FIGURE29} and Fig.~\ref{figure30}, which could be more meaningful. \vspace{12pt}

\newpage


\subsection{Techniques for more stable estimation}

We need to clarify that, to produce the heatmaps in Figures~\ref{figure_dsaoidj}, \ref{figure_4444dasdaslfdkjsdfq}, \ref{figure_activation}, and \ref{FIGURE29}, we need to perform test-time averaging over multiple forward runs. Because we make the encoder stochastic due to additive noise in the features or concatenated noise in the inputs, repeated forward passes of the same test sample inputs will produce different feature outputs. Therefore, for each run, we compute the metric distance and then average these matrices over multiple runs. The resulting averaged matrix is used for visualization. This procedure reduces the variance introduced by noise and produces smoother and more stable heatmaps.\vspace{10pt}

In this experiment, we also made the following efforts to make the estimation more stable. Consider the cost $c  =\frac{\left(\mathbb{E}\left[ \sum_{k=1}^K f_k(X)g_k(Y) \right]\right)^2}{\sum_{i,j=1}^K (\mathbf{R}_F \odot \mathbf{R}_G)_{i,j}}$. We find it to be much more stable to maximize the logarithm of this cost, in particular, maximizing
\begin{equation}
\resizebox{1\linewidth}{!}{
$\begin{aligned}
2 \cdot \log \Big(\mathbb{E}\big[ \sum_{k=1}^K f_k(X)g_k(Y) \big] + \epsilon \Big)
- \log \Big( \sum_{i,j=1}^K (\mathbf{R}_F \odot \mathbf{R}_G)_{i,j} + \epsilon \Big).
\end{aligned}$}
\label{maximizing_log_costs}
\end{equation}

\noindent We add a small constant $\epsilon$ not only to the second $\log$ term that corresponds to the denominator, but also to the first $\log$ term that corresponds to the numerator. We find this form to be more stable in practice. The constant $\epsilon$ can be chosen to be $10^{-9}$. Second, we find that concatenated input noise is helpful not only for training the autoencoder, but also for training the statistical dependence estimator here. We concatenate the inputs to the two estimator networks, $X$ and $Y$, with 10 dimensions of uniform noise. We find that this greatly stabilizes training and avoids the case in which the network suddenly takes a wrong step and collapses, possibly because the value inside the $\log$ becomes too small. The noises are resampled at each iteration, same as before.\vspace{15pt}

We can illustrate this on the MNIST example. We train a regular autoencoder that projects the samples from the original sample space onto a 1D feature space. We stop the training of the autoencoder at checkpoints $0, 100, 200, \cdots, 1000$. At each checkpoint, we then train the dependence estimator networks with the concatenated input noise introduced above, and maximize the cost described in Eq.~\eqref{maximizing_log_costs}, i.e., the logarithmic form of the two terms with a small constant added for numerical stability.

\begin{figure}[H]
  \centering
\begin{subfigure}{.45\textwidth}
\includegraphics[width=\linewidth]{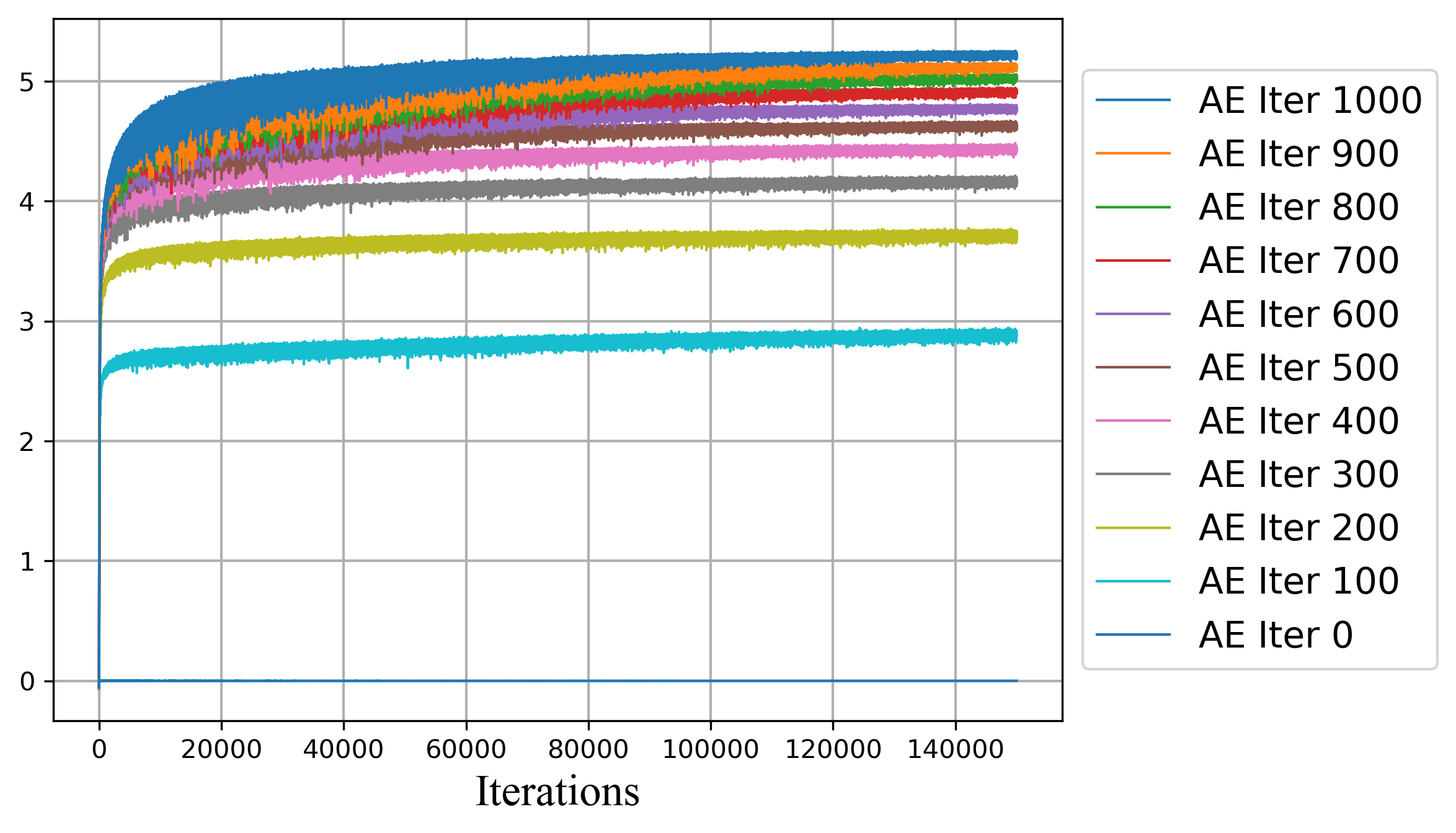}\vspace{-5pt}
\caption*{}
\end{subfigure}\vspace{-15pt}
\caption{Learning curves of the exact objective values described in Eq.~\eqref{maximizing_log_costs}, i.e., the logarithm of the original cost with small constants added inside the $\log$ terms for numerical stability, evaluated at multiple checkpoints during autoencoder training. The training of the dependence estimator is stable and fast. As the autoencoder becomes better trained, the estimated dependence value increases, which is consistent with our assumption.}
\label{figure31}
\end{figure}

Fig.~\ref{figure31} shows the learning curves for the exact objective values described in Eq.~\eqref{maximizing_log_costs}. We observe that the curves are stable, smooth, and converge to steady values. They clearly indicate that the converged statistical dependence estimate increases as the training of the autoencoder progresses, which matches our assumptions.

\vspace{15pt}

We also find that this stable implementation of our dependence estimator can be used to quantify the statistical dependence between any pair of internal layers in a neural network. We use a standard feedforward neural network as the decoder. The encoder has eight layers, and each encoder layer consists of a linear layer, a batch-normalization layer, and a ReLU activation, except for the final layer, which uses a Sigmoid activation. Before each linear layer, we concatenate the current features with a 200-dimensional vector of independent Gaussian noise.

\begin{figure}[H]
  \centering
  \begin{subfigure}[t]{.46\textwidth}
    \centering
    \includegraphics[width=.7\linewidth]{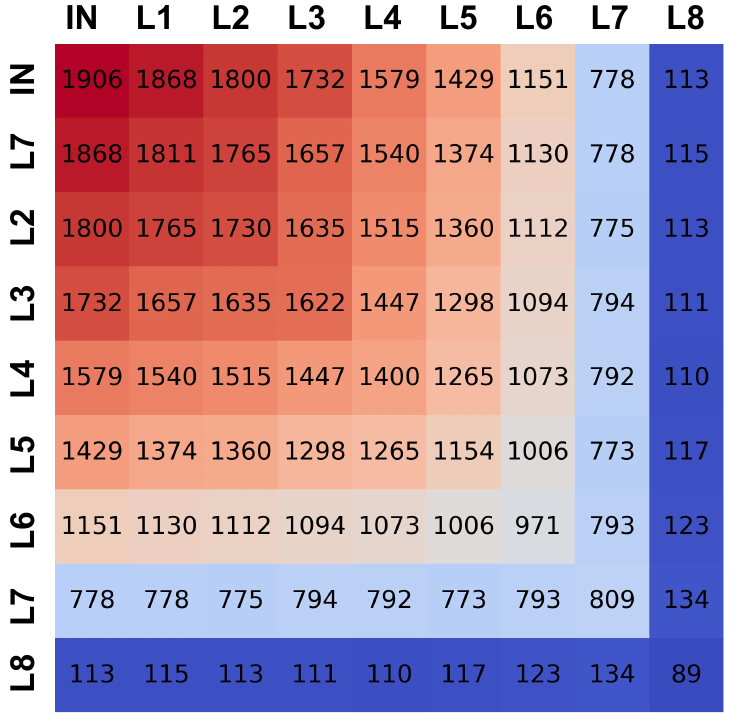}
    \caption{Estimated pairwise dependence matrix for the input and encoder representations using $6000$ MNIST samples. The estimator output dimension is $2000$, so the estimate is upper-bounded by $2000$, since there are at most $2000$ singular values and each is bounded by $1$. The largest value is $1906$ at $(\textbf{IN}, \textbf{IN})$.}
    \label{figure32a}
  \end{subfigure}\vspace{5pt}
  \begin{subfigure}[t]{.46\textwidth}
    \centering
    \includegraphics[width=.7\linewidth]{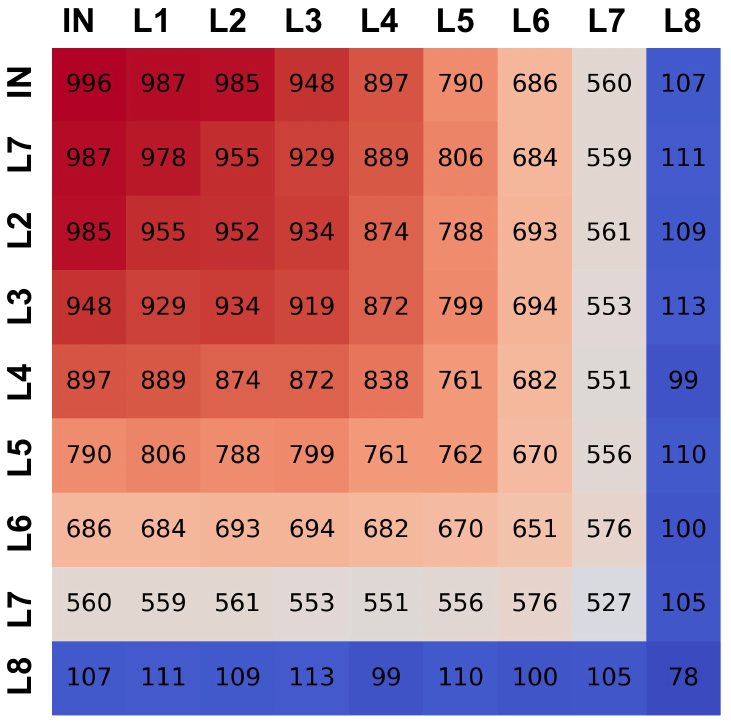}
    \caption{Estimated pairwise dependence matrix for the encoder using $1000$ MNIST samples, ensuring that the estimator is nearly unbiased. Since the estimator output dimension ($2000$) exceeds the sample size, all estimates are upper-bounded by $1000$. The qualitative dependence pattern remains largely consistent with panel~(a).}
    \label{figure32b}
  \end{subfigure}
  \caption{Layerwise statistical dependence in the trained encoder. \textbf{IN} denotes the input, \textbf{L1}--\textbf{L7} denote the outputs of the seven ReLU layers, and \textbf{L8} denotes the output of the final Sigmoid layer. Each matrix entry represents the estimated dependence between a pair of two layer representations, while diagonal entries represent self-dependence. Panel (a) that uses $6000$ samples could be biased, whereas panel (b) that uses $1000$ samples can be considered unbiased because the sample size $(1000)$ does not exceed the estimator output dimension $(2000)$, i.e., the total number of singular values. The concatenated noise in the encoder is important for producing a meaningful pattern, and its dimensionality affects the values in this matrix.}
  \label{figure_all}
\end{figure}

To analyze the encoder, we record the input samples, the outputs of the $7$ ReLU activations, and the output of the final Sigmoid activation, yielding $9$ variables in total. We then apply our proposed pairwise dependence estimator to every pair of variables, maximize the NMF-like objective, and obtain a $9 \times 9$ matrix of dependence estimates. The resulting heatmaps are shown in Figure~\ref{figure_all}.

In the heatmap, \textbf{IN} denotes the input samples, $\textbf{L1}, \cdots, \textbf{L7}$ denote the outputs of the $7$ ReLU layers, and \textbf{L8} denotes the output of the final Sigmoid layer. Each entry in the matrix represents the estimated statistical dependence between the corresponding pair of layers. Diagonal entries therefore represent self-dependence; that is, the two networks in the dependence estimator receive the same output from that particular layer.\vspace{10pt}

Several patterns can be observed in Figure~\ref{figure32a}. This experiment uses 6000 samples from a subset of MNIST, and the estimator network has an output dimension of 2000. Hence, the theoretical upper bound on the estimated dependence is 2000. The largest value is 1906, attained at $(\textbf{IN}, \textbf{IN})$. In addition, for each row, the largest off-diagonal value occurs in the first column, indicating that every hidden layer remains most strongly dependent on the input.

A third notable pattern concerns the entries around the diagonal. For a given layer, the entries to the left of the diagonal are only slightly smaller than the diagonal entry itself, whereas the entries to the right often decrease sharply. For example, the value at $(\textbf{L7}, \textbf{L7})$ is 809, and the values to its left are all around 780, whereas the value at $(\textbf{L7}, \textbf{L8})$ drops sharply to 134. This explains why the upper-left blocks of the heatmap have very similar colors.

The statistical dependence involving the output is much smaller than that of the other entries because the output dimension of the encoder network is $1$.

The dimensionality of the concatenated Gaussian noise is an important hyperparameter. We find that it has little effect on the training MSE of the autoencoder, but it can substantially reduce the magnitude of the dependence estimates and improve their stability.\vspace{10pt}

Figure~\ref{figure32b} shows the same analysis using only $1000$ MNIST samples instead of $6000$ samples, while keeping the estimator output dimension fixed at $2000$. In this setting, because the sample size is $1000$ and the estimator output dimension exceeds the sample size, the maximum attainable dependence value is $1000$. The qualitative structure of the heatmap remains largely unchanged. Since all values in the matrix are now bounded by $1000$, we can expect the dependence estimates to be more accurate without a bias.

\end{document}